\documentclass{clv3}

\usepackage{hyperref}
\usepackage{xcolor}
\definecolor{darkblue}{rgb}{0, 0, 0.5}
\hypersetup{colorlinks=true,citecolor=darkblue, linkcolor=darkblue, urlcolor=darkblue}

\usepackage{color}
\usepackage{colortbl}

\usepackage{microtype}

\usepackage{comment}
\usepackage{rotating}

\usepackage{multirow}
\usepackage{multicol}
\usepackage{graphicx}
\usepackage{booktabs}
\usepackage{makecell}
\usepackage{enumitem}
\usepackage{setspace}
\usepackage{longtable}
\usepackage{pdflscape}
\usepackage{tabu}
\usepackage{soul}
\usepackage{amsmath}

\usepackage{caption}
\usepackage{subcaption}

\colorlet{Mycolor1}{green!10!orange!90!}

\graphicspath{ {images/} }

\historydates{Submission received: 19 August 2020, Revised version received: 15 February 2021, Accepted for publication: 4 March 2021.}

\issue{0}{1}{2020}

\dochead{}

\runningtitle{Analysis and Evaluation of Language Models for WSD}

\runningauthor{Loureiro and Rezaee et al.}

\begin{document}


\title{Analysis and Evaluation of Language Models for Word Sense Disambiguation}

\author{Daniel Loureiro\thanks{Authors marked with * contributed equally}}
\affil{LIAAD - INESC TEC \\ Department of Computer Science - FCUP \\ University of Porto, Portugal\\
\texttt{dloureiro@fc.up.pt}}

\author{Kiamehr Rezaee$^*$}
\affil{Department of Computer Engineering, \\Iran University of Science and Technology, Tehran, Iran\\
\texttt{k$\_$rezaee@comp.iust.ac.ir}} 

\author{Mohammad Taher Pilehvar}
\affil{Tehran Institute for Advanced Studies, \\Tehran, Iran\\
\texttt{mp792@cam.ac.uk}}

\author{Jose Camacho-Collados}
\affil{School of Computer Science and Informatics, Cardiff University, \\ United Kingdom\\
\texttt{camachocolladosj@cardiff.ac.uk}}

\maketitle

\begin{abstract}
Transformer-based language models have taken many fields in NLP by storm. BERT and its derivatives dominate most of the existing evaluation benchmarks, including those for Word Sense Disambiguation (WSD), thanks to their ability in capturing context-sensitive semantic nuances.
However, there is still little knowledge about their capabilities and potential limitations in encoding and recovering word senses.
In this article, we provide an in-depth quantitative and qualitative analysis of the celebrated BERT model with respect to lexical ambiguity. 
One of the main conclusions of our analysis is that BERT can accurately capture high-level sense distinctions, even when a limited number of examples is available for each word sense.
Our analysis also reveals that in some cases language models come close to solving coarse-grained noun disambiguation under ideal conditions in terms of availability of training data and computing resources. 
However, this scenario rarely occurs in real-world settings and, hence, many practical challenges remain even in the coarse-grained setting. 
We also perform an in-depth comparison of the two main language model based WSD strategies, i.e., fine-tuning and feature extraction, finding that the latter approach is more robust with respect to sense bias and it can better exploit limited available training data.
In fact, the simple feature extraction strategy of averaging contextualized embeddings proves robust even using only three training sentences per word sense, with minimal improvements obtained by increasing the size of this training data.
\end{abstract}

\section{Introduction}

In the past decade, word embeddings have undoubtedly been one of the major points of attention in research on lexical semantics.
The introduction of Word2vec \cite{mikolov2013distributed}, as one of the pioneering word {\it embedding} models, generated a massive wave in the field of lexical semantics the impact of which is still being felt today.
However, static word embeddings (such as Word2vec) suffer from the limitation of being {\it fixed} or context insensitive, i.e., the word is associated with the same representation in all contexts, disregarding the fact that different contexts can trigger various meanings of the word, which might be even semantically unrelated.
Sense representations were an attempt at addressing the meaning conflation deficiency of word embeddings \cite{ReisingerMooney:2010,camacho2018word}.
Despite computing distinct representations for different senses of a word, hence addressing this deficiency of word embeddings, sense representations are not directly integrable into downstream NLP models.
The integration usually requires additional steps, including a (non-optimal) disambiguation of the input text, which make sense embeddings fall short of fully addressing the problem. 

The more recent {\it contextualized} embeddings \cite{peters-etal-2018-deep,devlin-etal-2019-bert} are able to simultaneously address both these limitations.
Trained with language modelling objectives, contextualized models can compute {\it dynamic} meaning representations for words in context that highly correlate with humans' word sense knowledge \cite{nair2020contextualized}. Moreover, contextualized embeddings provide a seamless integration into various NLP models, with minimal changes involved.
Even better, given the extent of semantic and syntactic knowledge they capture, contextualized models get close to the one system for all tasks setting.
Surprisingly, fine-tuning the same model on various target tasks often results in comparable or even higher performance when compared to sophisticated state-of-the-art task-specific models \cite{peters2019tune}. 
This has been shown for a wide range of NLP applications and tasks, including WSD, for which they have provided a significant performance boost, especially after the introduction of Transformer-based language models like BERT \cite{loureiro-jorge-2019-language,vial-etal-2019-sense,wiedemann2019does}.

Despite their massive success, there has been limited work on the analysis of recent language models and on explaining the reasons behind their effectiveness in lexical semantics.
Most analytical studies focus on syntax \cite{hewitt-manning:2019,saphra-lopez:2019} or explore the behaviour of self-attention heads \cite{clark-etal-2019-bert} or layers \cite{tenney-etal-2019-bert}, but there has been little work on investigating the potential of language models and their limitations in capturing other linguistic aspects, such as lexical ambiguity. 
Moreover, the currently-popular language understanding evaluation benchmarks, e.g., GLUE \cite{wang-etal-2018-glue} and SuperGLUE \cite{wang2019superglue}, mostly involve sentence-level representation which does not shed much light on the semantic properties of these models for individual words.\footnote{WiC \cite{pilehvar2019-wic} is the only SuperGLUE task where systems need to model the semantics of words in context (extended to several more languages in XL-WiC \cite{raganato2020xl}). In the appendix we provide results for this task.} 
To our knowledge, there has so far been no in-depth analysis of the abilities of contextualized models in capturing the ambiguity property of words.

In this article, we carry out a comprehensive analysis to investigate how pre-trained language models capture lexical ambiguity in the English language.
Specifically, we scrutinize the two major language model-based WSD strategies (i.e., feature extraction and fine-tuning) under various disambiguation scenarios and experimental configurations.
The main contributions of this work can be summarized as follows: (1) we provide an extensive quantitative evaluation of pre-trained language models in standard WSD benchmarks; (2) we develop a new dataset, CoarseWSD-20, which is particularly suited for the qualitative analysis of WSD systems; and (3) with the help of this dataset, we perform an in-depth qualitative analysis and test the limitations of BERT on coarse-grained WSD.
Data and code to reproduce all our experiments is available at \url{https://github.com/danlou/bert-disambiguation}.

The remainder of the article is organized as follows.
In Section \ref{sec:related_work}, we delineate the literature on probing pre-trained language models and on analyzing the potential of representation models in capturing lexical ambiguity.
We also describe in the same section the existing benchmarks for evaluating Word Sense Disambiguation.
Section \ref{wsd:overview} presents an overview of Word Sense Disambiguation and its conventional paradigms.
We then describe in the same section the two major approaches to utilizing language models for WSD, i.e., nearest neighbours feature extraction and fine-tuning.
We also provide a quantitative comparison of some of the most prominent WSD approaches in each paradigm in various disambiguation scenarios, including fine- and coarse-grained settings.
This quantitative analysis is followed by an analysis of models' performance per word categories (parts of speech) and for various layer-wise representations (in the case of language model based techniques).
Section \ref{sec:coarsewsd-20} introduces CoarseWSD-20, the WSD dataset we have constructed to facilitate our in-depth qualitative analysis.
In Section \ref{sec:evaluation} we evaluate the two major BERT-based WSD strategies on the benchmark.
To highlight the improvement attributable to contextualized embeddings, we also provide results of a linear classifier based on pre-trained FastText static word embeddings.
Based on these experiments, we carry out an analysis on the impact of fine-tuning and also compare the two strategies with respect to robustness across domains and bias towards the most frequent sense. 
Section \ref{sec:analysis} reports our observations upon further scrutinizing the two strategies on a wide variety of settings such as few-shot learning and different training distributions. 
Section \ref{sec:discussion} summarizes the main results from the previous sections and discusses the main takeaways. 
Finally, Section \ref{sec:conclusions} presents the concluding remarks and potential areas for future work.

\section{Related Work} 
\label{sec:related_work}

Recently, there have been several attempts at analyzing pre-trained language models. 
In Section \ref{related-analysis} we provide a general overview of the relevant works, while Section \ref{related-ambiguity} covers those related to lexical ambiguity.
Finally, in Section \ref{rel-datasets} we outline existing evaluation benchmarks for WSD, including CoarseWSD-20, which is the disambiguation dataset we have constructed for our qualitative analysis.

\subsection{Analysis of pre-trained language models}
\label{related-analysis}

Despite their young age, pre-trained language models, in particular those based on Transformers, have now dominated the evaluation benchmarks for most NLP tasks \cite{devlin-etal-2019-bert,liu2019roberta}. 
However, there has been limited work on understanding behind the scenes of these models. 

Various studies have shown that fulfilling the language modeling objective inherently forces the model to capture various linguistic phenomena.
A relatively highly-studied phenomenon is syntax, which is investigated both for earlier LSTM-based models \cite{linzen-etal-2016-assessing,kuncoro-etal-2018-lstms} as well as for the more recent Transformer-based ones \cite{goldberg2019assessing,hewitt-manning:2019,saphra-lopez:2019,jawahar-etal-2019-bert,van-schijndel-etal-2019-quantity,Tenney-et-al-2019}.
A recent work in this context is the \textit{probe} proposed by \citet{hewitt-manning:2019} which enabled them to show that Transformer-based models encode human-like parse trees to a very good extent. In terms of semantics, fewer studies exist, including the probing study of \citet{Ettinger2019WhatBI} on semantic roles, and that of \citet{tenney-etal-2019-bert} which also investigates entity types and relations.
The closest analysis to ours is that of \citet{peters-etal:2018-dissecting}, which provides a deep analysis of contextualized word embeddings, both from the representation point of view and per architectural choices.
In the same spirit, \citet{conneau-etal-2018-cram} proposed a number of linguistic probing tasks to analyze sentence embedding models.
Perhaps more related to the topic of this paper, \citet{shwartz2019still} showed how contextualized embeddings are able to capture non literal usages of words in the context of lexical composition. 
For a complete overview of existing probe and analysis methods, the survey of \citet{belinkov-glass-2019-analysis} provides a synthesis of analysis studies on neural network methods. The more recent survey of \citet{rogers2020primer} is a similar synthesis but targeted at BERT and its derivatives.

Despite all this analytical work, the investigation of neural language models from the perspective of ambiguity (and, in particular, lexical ambiguity) has been surprisingly neglected. 
In the following we discuss studies that aimed at shedding some light on this important linguistic phenomenon.

\subsection{Lexical ambiguity and language models}
\label{related-ambiguity}

Given its importance, lexical ambiguity has for long been an area of investigation in vector space model representations \cite{wordspace:93,ReisingerMooney:2010,camacho2018word}.
In a recent study on word embeddings, \citet{yaghoobzadeh-etal-2019-probing} showed that Word2vec \cite{Mikolovetal:2013} can effectively capture different coarse-grained senses if they are all frequent enough and evenly distributed. 
In this work we try to extend this conclusion to language model based representation and to the more realistic scenario of disambiguating words in context, rather than probing them in isolation for if they capture specific senses (as was the case in that work).

Most of the works analyzing language models and lexical ambiguity have opted for lexical substitution as their experimental benchmark. 
\citet{amrami-goldberg-2018-word} showed that an LSTM language model can be effectively applied to the task of word sense induction. 
In particular, they analyzed how the predictions of an LSTM for a word in context provided a useful way to retrieve substitutes, proving that this information is indeed captured in the language model. 
From a more analytical point of view, \citet{aina-etal-2019-putting} proposed a probe task based on lexical substitution to understand the internal representations of an LSTM language model for predicting words in context. Similarly, \citet{soler2019comparison} provided an analysis of LSTM-based contextualized embeddings in distinguishing between usages of words in context. 
As for Transformer-based models, \citet{zhou-etal-2019-bert} proposed a model based on BERT to achieve state-of-the-art results in lexical substitution, showing that BERT is particularly suited to find senses of a word in context.
While lexical substitution has been shown to be an interesting proxy for WSD, we provide a direct and in-depth analysis of the explicit capabilities of recent language models in encoding lexical ambiguity, both quantitatively and qualitatively.

Another related work to ours is the analysis of \citet{reif2019visualizing} on quantifying the geometry of BERT.
The authors observed that, generally, when contextualized BERT embeddings for ambiguous words are visualized, clear clusters for different senses are identifiable.
They also devised an experiment to highlight a potential failure with BERT (or presumably other attention-based models): it does not necessarily respect semantic boundaries when attending to neighboring tokens.
In our qualitative analysis in Section \ref{sec:bias} we further explore this. Additionally, \citet{reif2019visualizing} presents evidence supporting the specialization of representations from intermediate layers of BERT for sense representation, which we further confirm with layer-wise WSD evaluation in Section \ref{sec:layer}. 
Despite these interesting observations, the paper mostly focuses on the syntactic properties of BERT, similarly to most other studies in the domain (see Section \ref{related-analysis}).

Finally, a few works have attempted to induce semantic priors coming from knowledge resources like WordNet to improve the generalization of pre-trained language models like BERT \cite{levine2019sensebert,peters-etal-2019-knowledge}. Other works have investigated BERT's emergent semantic space using clustering analyses \cite{yenicelik-etal-2020-bert,chronis-erk-2020-bishop}, seeking to characterize how distinct sense-specific representations occupy this space.

Our work differs in that we are trying to understand to what extent pre-trained language models already encode this semantic knowledge and, in particular, what are their implicit practical disambiguation capabilities.

\subsection{Evaluation benchmarks}
\label{rel-datasets}

The most common evaluation benchmarks for WSD are based on fine-grained resources, with WordNet \cite{Fellbaum:98} being the de-facto sense inventory. For example, the unified all-words WSD benchmark of \citet{raganato-etal-2017-word} is composed of five datasets from Senseval/SemEval tasks, i.e., Senseval-2 \cite[SE02]{edmonds-cotton-2001-senseval}, Senseval-3 \cite[SE03]{Mihalceaetal:04b}, SemEval-2007 \cite[SE07]{Semeval:07}, SemEval-2013 \cite[SE13]{Naviglietal:13}, and SemEval-2015 \cite[SE15]{moro2015semeval}. \citet{vial-etal-2018-ufsac} extended this framework with other manually and automatically constructed datasets.\footnote{\citet{pasini-camacho-collados-2020-short} provide an overview of existing sense-annotated corpora for WordNet and other resources.} All these datasets are WordNet-specific and mostly use SemCor \cite{Milleretal:93} as their training set. 
Despite being the largest WordNet-based sense-annotated dataset, SemCor does not cover many senses occurring in the test sets, besides providing a limited number of examples per sense.
While scarcity in the training data is certainly a realistic setting, in this paper we are interested in analyzing the limits of language models with and without training data, also for senses not included in WordNet, and run a qualitative analysis. 

To this end, in addition to running evaluation in standard benchmarks, for this paper we constructed a coarse-grained word sense disambiguation dataset, called CoarseWSD-20. CoarseWSD-20 includes a selection of twenty ambiguous words of different nature (see Section \ref{sec:coarsewsd-20} for more details on CoarseWSD-20) where we run a qualitative analysis on various aspects of sense-specific information encoded in language models. 
Perhaps the closest datasets to CoarseWSD-20 are those of Lexical Sample WSD \cite{edmonds-cotton-2001-senseval,Mihalceaetal:04b,Pradhanetal:07a}.
These datasets usually target dozens of ambiguous words and list specific examples for their different senses. 
However, these examples are usually fine-grained, limited in number\footnote{For instance, the dataset of \citet{Pradhanetal:07a}, which is the most recent and the largest among the three mentioned lexical sample datasets, provides an average of 320/50 training/test instances for each of the 35 nouns in the dataset. In contrast, CoarseWSD-20 includes considerably larger datasets for all words (1,160 and 510 sentences on average for each word in the training and test sets, respectively).} and are limited to concepts (i.e., no entities such as \textit{Java} are included).
The CoarseWSD-20 dataset is similar in spirit, but has larger training sets extracted from Wikipedia.
Constructing the dataset based on the sense inventory of Wikipedia brings the additional advantage of having both entities and concepts as targets, and a direct mapping to Wikipedia pages, which is the most common resource for entity linking \cite{ling2015design,usbeck2015gerbil}, along with similar inter-connected resources such as DBpedia.

Another related dataset to CoarseWSD-20 is WIKI-PSE \cite{yaghoobzadeh-etal-2019-probing}.
Similarly to ours, WIKI-PSE is constructed based on Wikipedia, but with a different purpose. 
WIKI-PSE clusters all Wikipedia concepts and entities into eight general ``semantic classes''. 
This is an extreme coarsening of the sense inventory that may not fully reflect the variety of human-interpretable senses that a word has. 
Instead, for CoarseWSD-20, sense coarsening is performed at the word level which preserves sense-specific information. 
For example, the word \textit{bank} in WIKI-PSE is mainly identified as a \textit{location} only, conflating the financial institution and river meanings of the word.
Whereas CoarseWSD-20 distinguishes between the two senses of \textit{bank}.
Moreover, our dataset is additionally post-processed in a semi-automatic manner (an automatic pre-processing, followed by a manual check for problematic cases), which helps remove errors from the Wikipedia dump.

\section{Word Sense Disambiguation: An Overview}
\label{wsd:overview}

Our analysis is focused on the task of Word Sense Disambiguation (WSD). 
WSD is a core module of human cognition and a long-standing task in Natural Language Processing. 
Formally, given a word in context, the task of WSD consists of selecting the intended meaning (sense) from a pre-defined set of senses for that word defined by a sense inventory \cite{navigli:09}. 
For example consider the word \textit{star} in the following context:
\begin{description}
  \item[$\bullet$] Sirius is the brightest \textit{star} in Earth's night.
\end{description}
The task of a WSD system is to identify that the usage of \textit{star} in this context refers to its astronomical meaning (as opposed to celebrity or star shape, among others).
The context could be a document, a sentence, or any other information-carrying piece of text that can provide a hint on the intended semantic usage\footnote{For this analysis we focus on sentence-level WSD, since it is the most standard practice in the literature.}, probably as small as a word, e.g., ``\textit{dwarf} star''.\footnote{A \textit{dwarf star} is a relatively small star with low luminosity, such as the Sun.}

WSD is described as an AI-hard\footnote{By analogy to NP-completeness, the most difficult problems are referred to as AI-complete, implying that solving them is equivalent to solving the central artificial intelligence problem.}  problem \cite{Mallery:88}.
In a comprehensive survey of WSD, \citet{navigli:09} discusses some of the reasons behind its difficulty, including heavy reliance on knowledge, difficulty in distinguishing fine-grained sense distinctions, and lack of application to real-world tasks.
On WordNet-style sense inventories, the human level performance (which is usually quoted as glass ceiling) is estimated to be 80\% in the fine-grained setting \cite{Galeetal:92c} and 90\% for the coarse-grained one \cite{Palmeretal:07}.
This performance gap can be mainly attributed to the fine-grained semantic distinctions in WordNet which are sometimes even difficult for humans to distinguish.
For instance, the noun \textit{star} has 8 senses in WordNet 3.1, two of which refer to the astronomical sense (celestial body) with the minor semantic difference of if the star is visible from Earth at night.
In fact, it is argued that sense distinctions in WordNet are too fine-grained for many NLP applications \cite{Hovyetal:13}.
CoarseWSD-20 addresses this issue by devising sense distinction that are easily interpretable by humans, essentially pushing the human performance on the task. 

Similarly to many other tasks in Natural Language Processing, WSD has gone under significant change after the introduction of Transformer-based language models, which are now dominating most WSD benchmarks.
In the following we first present a background on existing sense inventories, with a focus on WordNet (Section \ref{inventories}), and then describe the state of the art in both the conventional paradigm (Section \ref{paradigms}) and the more recent paradigm based on (Transformer-based) language models (Section \ref{languagemodelsWSD}).
We then carry out a quantitative evaluation of some of the most prominent WSD approaches in each paradigm in various disambiguation scenarios, including fine- and coarse-grained settings (Section \ref{evaluation}).
This quantitative analysis is followed by an analysis of layer-wise representations (Section \ref{sec:layer}) and performance per word categories (parts of speech, Section \ref{sec:pos}).

\subsection{Sense inventories}
\label{inventories}

Given that WSD is usually tied with sense inventories, we briefly describe existing sense inventories that are also used in our experiments. The main sense inventory for WSD research in English is the Princeton \textbf{WordNet} \cite{Fellbaum:98}. 
The basic constituents of this expert-made lexical resource are \textit{synsets}, which are sets of synonymous words that represent unique concepts.
A word can belong to multiple synsets denoting to its different meanings. Version 3.0 of the resource, which is used in our experiments, covers 147,306 words and 117,659 synsets.\footnote{There are several other variants of WordNet available, either the newer v3.1, which is slightly different from the former version, or other non-Princeton versions that improve coverage, such as WordNet 2020 \cite{mccrae-etal-2020-english} or CROWN \cite{jurgens-pilehvar-2015-reserating}. We opted for v3.0 given that it is the widely-used inventory according to which most existing benchmarks are annotated.} WordNet is also available for languages other than English through the Open Multilingual WordNet project \cite{bond2013linking} and related efforts.

Other common sense inventories are Wikipedia and BabelNet. The former is generally used for Entity Linking or \textit{Wikification} \cite{mihalcea2007wikify}, in which the Wikipedia pages are considered as concept or entities to be linked in context. On the other hand, BabelNet \cite{NavigliPonzetto:12aij} is a merger of WordNet, Wikipedia and several other lexical resources, such as Wiktionary and OmegaWiki. One of the key features of this resource is its multilinguality, highlighted by the 500 languages covered in its most recent release (version 5.0).

\subsection{WSD paradigms}
\label{paradigms}

WSD approaches are traditionally categorized as {\it knowledge-based} and {\it supervised}. 
The latter makes use of sense-annotated data for its training whereas the former exploits sense inventories, such as WordNet, for the encoded knowledge, such as sense glosses \cite{Lesk:86,BanerjeePedersen:03,basile-caputo-semeraro:2014:Coling}, semantic relations \cite{agirre2014random,Moroetal:14tacl} or sense distributions \cite{chaplot2018knowledge}.
Supervised WSD has been shown to clearly outperform the knowledge-based counterparts, even before the introduction of pre-trained language models \cite{raganato-etal-2017-word}. 
Large pre-trained language models have further provided improvements, with BERT-based models currently approaching human-level performance \cite{loureiro-jorge-2019-language,vial-etal-2019-sense,huang-etal-2019-glossbert,bevilacqua-navigli-2020-breaking,blevins-zettlemoyer-2020-moving}.
A third category of WSD techniques, called \textit{hybrid}, has recently attracted more attention.
In this approach, the model benefits from both sense-annotated instances and knowledge encoded in sense inventories.\footnote{Note that knowledge-based WSD systems might benefit from sense frequency information obtained from sense-annotated data, such as SemCor. Given that such models do not incorporate sense-annotated instances, we do not categorize them as hybrid.}
Most of the recent state-of-the-art approaches can be put in this category.

\subsection{Language models for WSD}
\label{languagemodelsWSD}

In the context of Machine Translation (MT), a language model is a statistical model that estimates the probability of a sequence of words in a given language. 
Recently, the scope of LMs has gone far beyond MT and generation tasks.
This is partly due to the introduction of Transformers \cite{vaswani2017attention}, attention-based neural architectures that have proven immense potential in capturing complex and nuanced linguistic knowledge.
In fact, despite the young age, Transformer-based LMs dominate most language understanding benchmarks, such as GLUE \cite{wang-etal-2018-glue} and SuperGLUE \cite{wang2019superglue}.

There are currently two popular varieties of Transformer-based Language Models, differentiated most significantly by their choice of language modelling objective.
There are causal (or left-to-right) models, epitomized by GPT-3 \cite{brown2020language}, where the objective is to predict the next word, given the past sequence of words.
Alternatively, there are masked models, where the objective is to predict a masked (i.e. hidden) word given its surrounding words, traditionally known as the Cloze task \cite{taylor1953cloze}, of which the most prominent example is BERT.
Benchmark results reported in \citet{devlin-etal-2019-bert} and \citet{brown2020language} show that masked LMs are preferred for semantic tasks, whereas causal LMs are more suitable for language generation tasks.
As a potential explanation for the success of BERT-based models, \citet{voita-etal-2019-bottom} present empirical evidence suggesting that the masked LM objective induces models to produce more generalized representations in intermediate layers.

In our experiments, we opted for the BERT \cite{devlin-etal-2019-bert} and ALBERT \cite{lan2019albert} models given their prominence and popularity.
Nonetheless, our empirical analysis could be applied to other pre-trained language models as well (e.g., \citealt{liu2019roberta,raffel2019exploring}).
Our experiments focus on two dominant WSD approaches based on language models: (1) nearest neighbors classifiers based on features extracted from the model (Section \ref{sec:nn}), and (2) fine-tuning of the model for WSD classification (Section \ref{sec:fine-tuning}). 
In the following we describe the two strategies.

\subsubsection{Feature extraction}
\label{sec:nn}

Neural LMs have been utilized for WSD, even before the introduction of Transformers, when LSTMs were the first choice for encoding sequences \cite{melamud-etal-2016-context2vec,yuan-etal-2016-semi,peters-etal-2018-deep}.
In this context, LMs were often used to encode the context of a target word, or in other words, generate a contextual embedding for that word.
Allowing for various sense-inducing contexts to produce different word representations, these contextual embeddings proved more suitable for lexical ambiguity than conventional word embeddings (e.g. Word2vec).

Consequently, \citet{melamud-etal-2016-context2vec,yuan-etal-2016-semi,peters-etal-2018-deep} independently demonstrated that, given sense-annotated corpora (e.g., SemCor), it is possible to compute an embedding for a specific word sense as the average of its contextual embeddings. 
Sense embeddings computed in this manner serve as the basis for a series of WSD systems.
The underlying approach is straightforward: match the contextual embedding of the word to be disambiguated against its corresponding pre-computed sense embeddings.
The matching is usually done using a simple $k$ Nearest Neighbors (often with $k=1$) classifier; hence, we refer to this feature extraction approach as \textbf{1NN} in our experiments.
A simple 1NN approach based on LSTM contextual embeddings proved effective enough to rival the performance of other systems using task-specific training, such as \citet{raganato-etal-2017-neural}, despite using no WSD specific modelling objectives. 
\citet[LMMS]{loureiro-jorge-2019-language} and \citet{wiedemann2019does} independently showed that the same approach using contextual embeddings from BERT could in fact surpass the performance of those task-specific alternatives.  \citet{loureiro-jorge-2019-language} also explored a propagation method using WordNet to produce sense embeddings for senses not present in training data (LMMS$_{1024}$) and a variant which introduced information from glosses into the same embedding space (LMMS$_{2048}$). 
Similar methods have been also introduced for larger lexical resources such as BabelNet, with similar conclusions \citep[SensEmBERT]{scarlinietal:2020}. 

There are other methods based on feature extraction, while not using 1NN for making predictions.
\citet[Sense Compression]{vial-etal-2019-sense} used contextual embeddings from BERT as input for additional Transformer encoder layers with a softmax classifier on top. \citet{blevins-zettlemoyer-2020-moving} also experimented with a baseline using the final states of a BERT model with a linear classifier on top. 
Finally, the solution by \citet{bevilacqua-navigli-2020-breaking} relied on an ensemble of sense embeddings from LMMS and SensEmBERT, along with additional resources, to train a high performance WSD classifier.

\subsubsection{Fine-tuning}
\label{sec:fine-tuning}

Another common approach to benefiting from  contextualized language models in downstream tasks is fine-tuning.
For each target task, it is possible to simply plug in the task-specific inputs and outputs into pre-trained models, such as BERT, and fine-tune all or part of the parameters end-to-end.
This procedure adjusts model's parameters according to the objectives of the target task, e.g., the classification task in WSD. One of the main drawbacks of this type of supervised model is their need for building a model for each word, which is unrealistic in practice for all-words WSD. However, there are several successful WSD approaches in this category that overcome this limitation in different ways. GlossBERT \cite{huang-etal-2019-glossbert} uses sense definitions to fine-tune the language model for the disambiguation task, similarly to a text classification tasks. KnowBERT \cite{peters-etal-2019-knowledge} fine-tunes BERT for entity linking exploiting knowledge bases (WordNet and Wikipedia) as well as sense definitions. BEM \cite{blevins-zettlemoyer-2020-moving} proposes a bi-encoder method which learns to represent sense embeddings leveraging sense definitions while performing the optimization jointly with the underlying BERT model.

\subsection{Evaluation in standard benchmarks}
\label{evaluation}

In our first experiment, we perform a quantitative evaluation on the unified WSD evaluation framework (Section \ref{sec:unified}), which verifies the extent to which a model can distinguish between different senses of a word as defined by WordNet's 
inventory.

\subsubsection{BERT models}
\label{bertmodels}

For this task we employ a Nearest Neighbors strategy (1NN henceforth) that has been shown to be effective with pre-trained language models, both for LSTMs and more recently for BERT (see Section \ref{sec:nn}).
In particular, we used the cased base and large variants of BERT released by \cite{devlin-etal-2019-bert}, as well as the xxlarge (v2) variant of ALBERT \cite{lan2019albert}, via the Transformers framework (v2.5.1) \cite{Wolf2019HuggingFacesTS}.
Following LMMS, we also average sub-word embeddings and represent contextual embeddings as the sum of the corresponding representations from the final four layers. However, here we do not apply LMMS propagation method aimed at fully representing the sense inventory, resorting to the conventional MFS fallback for lemmas unseen during training.

\subsubsection{Comparison systems}
\label{comparisonsystems}

In addition to BERT and ALBERT, we include results for 1NN systems that exploit precomputed sense embeddings, namely Context2vec \cite{melamud-etal-2016-context2vec} and ELMo \cite{peters-etal-2018-deep}.
Moreover, we include results for hybrid systems, i.e., supervised models that also make use of additional knowledge sources (cf. Section \ref{paradigms}), particularly semantic relations and textual definitions in WordNet. 
Besides the models already discussed in Sections \ref{sec:nn} and \ref{sec:fine-tuning}, we also report results from additional hybrid models. \citet[Seq2Seq]{raganato-etal-2017-neural} trained a neural BiLSTM sequence model with losses specific not only to specific senses from SemCor but also part-of-speech tags and WordNet supersenses. EWISE \cite{kumar-etal-2019-zero}, which inspired EWISER \cite{bevilacqua-navigli-2020-breaking}, also employs a BiLSTM to learn contextual representations that can be matched against sense embeddings learned from both sense definitions and semantic relations.

For completeness we also add some of the best linear supervised baselines, namely IMS \cite{ZhongNg:2010} and IMS with embeddings \cite[IMS+emb]{ZhongNg:2010,iacobacci-etal-2016-embeddings}, which are Support Vector Machine (SVM) classifiers based on several manually-curated features. 
Finally, we report results for knowledge-based systems (KB) that mainly rely on WordNet: Lesk\textsubscript{ext}+emb \cite{basile-caputo-semeraro:2014:Coling}, Babelfy \cite{Moroetal:14tacl}, UKB \cite{agirre-etal-2018-risk}, and TM \cite{chaplot2018knowledge}.
More recently, SyntagRank \cite{scozzafava-etal-2020-personalized} showed best KB results by combining WordNet with the SyntagNet \cite{maru-etal-2019-syntagnet} database of syntagmatic relations.
However, as it was discussed in Section \ref{paradigms}, we categorize these as knowledge-based since they do not directly incorporate sense-annotated instances as their source of knowledge.

\subsubsection{Datasets: Unified WSD Benchmark}
\label{sec:unified}

Introduced by \citet{raganato-etal-2017-word} as an attempt to construct a standard evaluation framework for WSD, the unified benchmark comprises five datasets from Senseval/SemEval workshops (see Section \ref{rel-datasets}).\footnote{Dataset downloaded from \url{http://lcl.uniroma1.it/wsdeval/}}
The framework provides 7,253 test instances for 4,363 sense types. In total, around 3,663 word types are covered with an average polysemy of 6.2 and across four parts of speech: nouns, verbs, adjectives, and adverbs.

Note that the datasets are originally designed for the fine-grained WSD setting.
Nonetheless, in addition to the fine-grained setting, we provide results on the coarse-grained versions of the same test sets. 
To this end, we merged those senses that belonged to the same domain according to CSI (Coarse Sense Inventory) domain labels from \citet{lacerra2020csi}.\footnote{CSI domains downloaded from \url{http://lcl.uniroma1.it/csi}} 
With this coarsening, we can provide more meaningful comparisons and draw interpretable conclusions.
Finally, we followed the standard procedure and trained all models on SemCor \cite{Milleretal:93}.

\begin{table*}
\begin{center}
\resizebox{\textwidth}{!}{
\begin{tabular}{lllcccccccccc|cc}
\toprule
\multicolumn{1}{c}{} &
\multicolumn{1}{c}{\multirow{2}{*}{\textbf{Type}}} &
\multicolumn{1}{c}{\multirow{2}{*}{\textbf{System}}}	&
\multicolumn{2}{c}{\bf SE2} &
\multicolumn{2}{c}{\bf  SE3} & 
\multicolumn{2}{c}{\bf SE07} &
\multicolumn{2}{c}{\bf SE13} & 
\multicolumn{2}{c|}{\bf SE15} &  
\multicolumn{2}{c}{\bf ALL} \\
\cmidrule(lr){4-5}
\cmidrule(lr){6-7}
\cmidrule(lr){8-9}
\cmidrule(lr){10-11}
\cmidrule(lr){12-13}
\cmidrule(lr){14-15}
\multicolumn{1}{c}{} &
 &	&
\multicolumn{1}{c}{\bf FN} &
\multicolumn{1}{c}{\bf CS} & 
\multicolumn{1}{c}{\bf FN} &
\multicolumn{1}{c}{\bf CS} &
\multicolumn{1}{c}{\bf FN} &
\multicolumn{1}{c}{\bf CS} &
\multicolumn{1}{c}{\bf FN} &
\multicolumn{1}{c}{\bf CS} &
\multicolumn{1}{c}{\bf FN} &
\multicolumn{1}{c|}{\bf CS} &
\multicolumn{1}{c}{\bf FN} &
\multicolumn{1}{c}{\bf CS} \\
\cmidrule(lr){1-15}
\multicolumn{2}{l}{\multirow{5}{*}{\textbf{KB}}}
& Lesk\textsubscript{ext}+emb 	& 63.0 & 74.9 & 63.7 & 75.5 & 56.7 & 71.6 & 66.2 & 77.4 & 64.6 & 73.9 & 63.7 & 75.3     \\
& & Babelfy$\dagger$  & 67.0 & 78.4 & 63.5 & 77.5 &  51.6 & 68.8  	& 66.4 &  77.0 & 70.3 & 79.1 & 65.5 & 77.3 \\
& & TM  & 69.0 & - &	66.9 & - & 55.6 & - & 65.3 & - & 69.6 & - & 66.9 & -      \\
& & UKB & 68.8 & 81.2 & 66.1 & 78.1 & 53.0 & 70.8 & 68.8 & 79.1 & 70.3 & 77.4 & 67.3 & 78.7        \\
& & SyntagRank & \textbf{71.6} & - &	\textbf{72.0} & - & \textbf{59.3} & - & \textbf{72.2} & - & \textbf{75.8} & - & \textbf{71.7} & -      \\
\midrule
\midrule
\multirow{7}{*}{\begin{sideways}\textbf{Supervised}\end{sideways}}
& \multirow{2}{*}{\textbf{SVM}} 
& IMS & 70.9 & 81.5 & 69.3 & 80.8 & 61.3 & 74.3 & 65.3 & 77.4 & 69.5 & 75.7 & 68.4 & 79.1    \\
& & IMS+emb & 72.2 & 82.8 & 70.4 & 81.5 & 62.6 & 75.8 & 65.9 & 76.9 & 71.5 & 76.7 & 69.6 & 79.8  \\
\cmidrule{2-15}
& \multirow{5}{*}{\textbf{1NN}} & Context2vec & 71.8 & 82.6 & 69.1 & 80.5 & 61.3 & 74.5 & 65.6 & 78.0 & 71.9 & 76.6 & 69.0 & 79.7   \\
& & ELMo & 71.6 & 82.8 & 69.6 & 80.9 & 62.2 & 74.7 & 66.2 & 77.7 &	71.3 & 77.0 & 69.0	&  79.6     \\
& & BERT-Base & 75.5 & 84.9 & 71.5 & 81.4 & 65.1 & 78.9 & 69.8 & 82.1 & 73.4 & 78.1 & 72.2 & 82.0        \\
& & BERT-Large & 76.3 & 84.8 & \textbf{73.2} & \textbf{82.9} & 66.2 & 80.0 & 71.7 & 83.1 & 74.1 & \textbf{79.1} & 73.5 & 82.8      \\
& & ALBERT-XXL & \textbf{76.6} & \textbf{85.6} & 73.1 & 82.6 & \textbf{67.3} & \textbf{80.1} & \textbf{71.8} & \textbf{83.5} & \textbf{74.3} & 78.3 & \textbf{73.7} & \textbf{83.0} \\
\midrule
\midrule
\multicolumn{2}{l}{\multirow{9}{*}{\textbf{Hybrid}}}
& Seq2Seq \scriptsize{Att+Lex+PoS} & 70.1 &  -  & 68.5	& - &  63.1*  &  - &  66.5  &  - &  69.2 &  -  & 68.6* & - \\
& & Sense Compr. \scriptsize{Ens.} & 79.7 &  -  & 77.8	& - &  73.4  &  - &  78.7  &  - &  82.6 &  -  & \textbf{79.0} & - \\
& & LMMS \scriptsize{1024} & 75.4 &  -  & 74.0 & - & 66.4 & - & 72.7 & - & 75.3 & - & 73.8 & - \\
& & LMMS \scriptsize{2048} & 76.3 &  84.5  & 75.6 & 85.1 & 68.1 & 81.3 & 75.1 & 86.4 & 77.0 & 80.8 & 75.4 & 84.4 \\
& & EWISE & 73.8 &  -  &  71.1 	& - &  ~~67.3*  &  - &  69.4  &  - &  74.5 &  -  &  ~~71.8* & - \\
& & KnowBert$\dagger$ \scriptsize{WN+WK}  & 76.4 &  85.6  &  76.0 	& 85.1 &  71.4  &  82.6 &  73.1  &  83.8 &  75.4 &  80.2  &  75.1 & 84.1 \\
& & GlossBERT & 77.7 &  -  &  75.2	& - &  72.5*  &  - &  76.1  &  - &  80.4 &  -  & 77.0* & - \\
& & BEM & 79.4 &  -  &  77.4	& - &  74.5*  &  - &  79.7  &  - &  \textbf{81.7} &  -  & 79.0* & - \\
& & EWISER$\dagger$ & \textbf{80.8} &  -  & \textbf{79.0}	& - &  \textbf{75.2}  &  - &  \textbf{80.7}  &  - &  81.8* &  -  & 80.1* & - \\
\midrule
\midrule
& - & \textit{MFS Baseline} & 65.6 & 77.4 & 66.0 & 77.8 & 54.5 & 70.6	& 63.8 & 74.8 & 67.1 & 75.3 & 64.8 & 76.2\\
\bottomrule
\end{tabular}
}
\end{center}
\caption{\label{tab:evWSD} F-Measure performance on the unified WSD evaluation framework \cite{raganato-etal-2017-word} for three classes of WSD models, i.e., knowledge-based (KB), supervised, and hybrid, and for two sense specification settings, i.e., fine-grained (FN) and coarse-grained (CS). Results marked with * make use of SE07/SE15 as development set. Systems marked with $\dagger$ rely on external resources other than WordNet. The results from complete rows were computed by ourselves given the system outputs, while those from incomplete rows were taken from the original papers.}
\end{table*}

\subsubsection{Results} Table \ref{tab:evWSD} shows the results\footnote{SensEmBERT not included because it is only applicable to the noun portions of these test sets.} of all comparison systems on the unified word sense disambiguation framework, both for fine-grained (FN) and coarse-grained (CS) versions.
The LMMS$_{2048}$ hybrid model, which is based on the 1NN BERT classifier is the best-performer based solely on feature extraction. The latest fine-tuning hybrid solutions, particularly BEM and EWISER, show overall best performance, making the case for leveraging glosses and semantic relations to optimize pre-trained weights for the WSD task.
Generally, all BERT-based models achieve fine-grained results which are in the same ballpark as human average inter-annotator agreements for fine-grained WSD, which ranges from 64\% and 80\% in the three earlier datasets of this benchmark \cite{navigli:09}. 
In the more interpretable coarse-grained setting, LMMS achieves a score of 84.4\%, similar to the other BERT-based models which surpass 80\%. The remaining supervised models perform roughly equal, marginally below 80\% and clearly underperformed by BERT-based models. 

\begin{figure}
\centering
\includegraphics[width=1.0\linewidth]{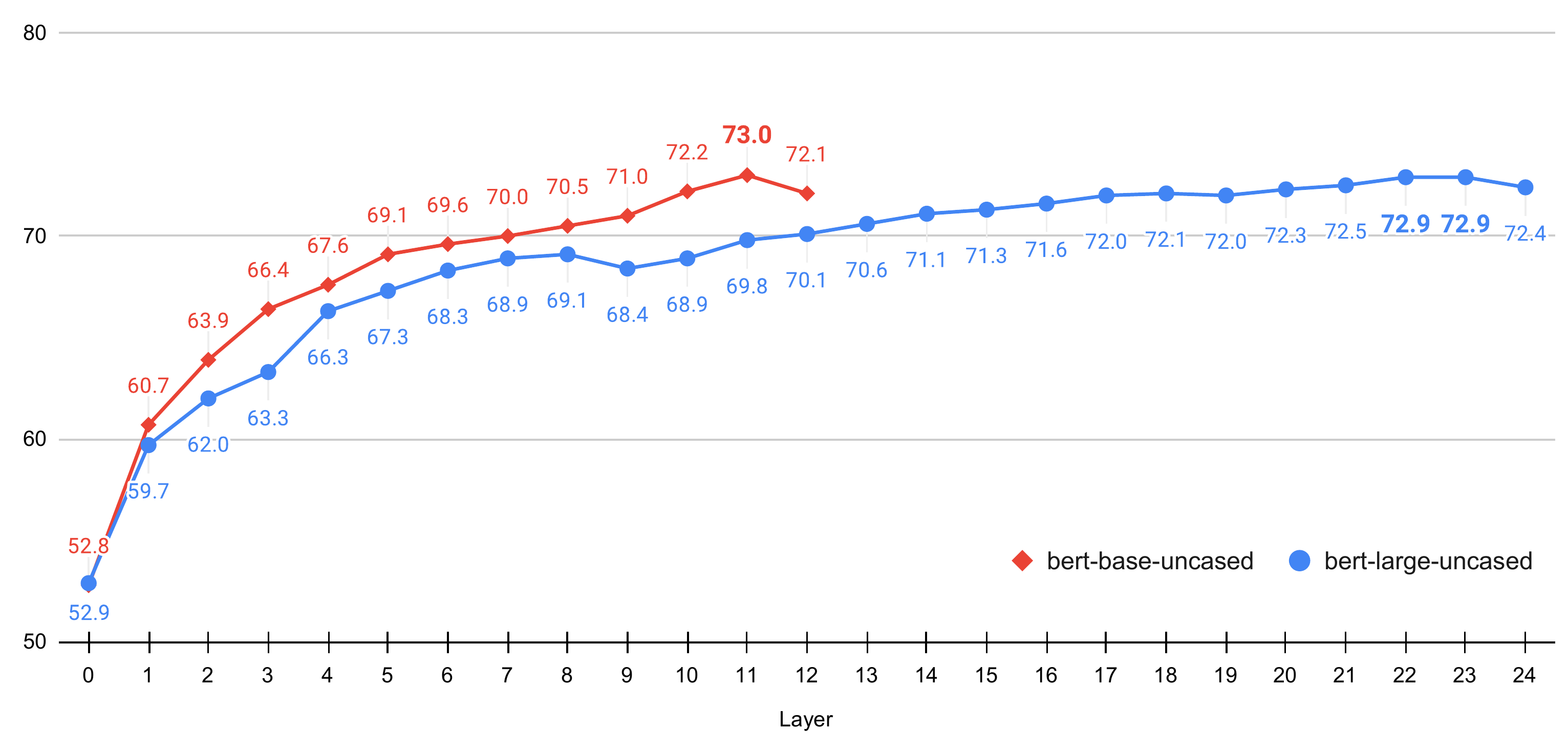}
\caption{F-measure performance on a restricted version of the MASC corpus \cite{ide2008masc} for representations derived from individual layers of the two BERT models used in our experiments.
\label{fig:layer_perf}}
\end{figure}

\subsubsection{Layer performance}
\label{sec:layer}

Current BERT-based 1NN WSD methods (see Section \ref{sec:nn}), such as LMMS and SensEmBERT, apply a \textit{pooling} procedure to combine representations extracted from various layers of the model.
The convention is to sum the embeddings from the last four layers, following the Named Entity Recognition experiments reported by \citet{devlin-etal-2019-bert}. 
It is generally understood that lower layers are closer to their static representations (i.e., initialization) and, conversely, upper layers better match the modelling objectives \cite{tenney-etal-2019-bert}. 
Still, \citet{reif2019visualizing} have shown that this relation is not monotonic when it comes to sense representations from BERT.
Additional probing studies have also pointed to irregular progression of context-specificity and token identity across the layers \cite{ethayarajh-2019-contextual,voita-etal-2019-bottom}, two important pre-requisites for sense representation.

Given our focus on measuring BERT's adeptness for WSD, and the known variability in layer specialization, we performed an analysis to reveal which layers produce representations that are most effective for WSD. 
This analysis involved obtaining sense representations learned from SemCor for each layer individually using the process described in Section \ref{sec:nn}. 

Figure \ref{fig:layer_perf} shows the performance of each layer using a restricted version of the MASC corpus \cite{ide2008masc} as a validation set where only annotations for senses that occur in SemCor are considered.
Any sentence that contained annotations for senses not occurring in SemCor was removed, restricting this validation set to 14,645 annotations out of 113,518.
We restrict the MASC corpus so that our analysis isn't affected by strategies for inferring senses (e.g. Network Propagation) or fallbacks (e.g. Most Frequent Sense).
This restricted version of MASC is based on the release introduced in \citet{vial-etal-2018-ufsac}, which mapped annotations to Princeton WordNet (3.0).

Similarly to \citet{reif2019visualizing}, we find that lower layers are not as effective for disambiguation as upper layers. 
However, our experiment specifically targets WSD and its results suggest a different distribution of the best performing layers than those reported by \citet{reif2019visualizing}. 
Nevertheless, this analysis shows that the current convention of using the sum of last four layers for sense representations is sensible, even if not optimal.

Several model probing works have revealed that the scalar mixing method introduced by \citet{peters-etal-2018-deep} allows for combining information from all layers with improved performance on lexico-semantic tasks \cite{liu-etal-2019-linguistic,Tenney-et-al-2019,de-vries-etal-2020-whats}.
However, scalar mixing essentially involves training a learned probe, which can limit attempts at analysing the inherent semantic space represented by NLMs \cite{mickus-etal-2020-what}.

\subsubsection{Analysis by Part-of-Speech}
\label{sec:pos}
Table \ref{tab:table_pos} shows the results of BERT and the comparison systems by part of speech.\footnote{For this table we only included systems for which we got access to their system outputs.} 
The results clearly show that verbs are substantially more difficult to model, which corroborates the findings of \citet{raganato-etal-2017-word}, while adverbs are the least problematic in terms of disambiguation. 
For example, in the fine-grained setting, BERT-Large achieves an overall F1 of 75.1\% on nouns vs. 63.2\% on verbs (85.3\% on adverbs). 
The same trend is observed for other models, including hybrid ones. This may also be related to the electrophysiological evidence suggesting humans process nouns and verbs differently \cite{10.1093/brain/123.12.2552}. 
Another more concrete reason to this gap is due to the fine granularity of verb senses in WordNet. 
For instance, the verb \textit{run} has 41 sense entries in WordNet, twelve of which denote some kind of motion. 

\begin{table}[]
\setlength{\tabcolsep}{9pt}
\scalebox{0.8}{
\begin{tabular}{lllcccccccc}
\toprule
  & \multirow{2}{*}{\bf Type} & \bf \multirow{2}{*}{\bf System} 
  { \textbf{}} &
  \multicolumn{2}{c}{{ \textbf{Nouns}}} &
  \multicolumn{2}{c}{{ \textbf{Verbs}}} &
  \multicolumn{2}{c}{{ \textbf{Adjectives}}} &
  \multicolumn{2}{c}{{ \textbf{Adverbs}}} \\
  \cmidrule(lr){4-11}
   &
    { } &
  { \textbf{}} &
  { \textbf{FN}} &
  { \textbf{CS}} &
  { \textbf{FN}} &
  { \textbf{CS}} &
  { \textbf{FN}} &
  { \textbf{CS}} &
  { \textbf{FN}} &
  { \textbf{CS}} \\
  \midrule
  
\multicolumn{2}{l}{\multirow{3}{*}{\textbf{KB}}} 
&
  {UKB} &
  71.2 &
  \multicolumn{1}{l}{80.5} &
  50.7 &
  \multicolumn{1}{l}{69.2} &
  75.0 &
  \multicolumn{1}{l}{82.7} &
  77.7 &
  \multicolumn{1}{l}{91.3} \\
 &
 &
  {Lesk$_{ext}$+emb} &
  69.8 &
  79.0 &
  51.2 &
  69.2 &
  51.7 &
  62.4 &
  80.6 &
  92.8 \\
  &
&
  {Babelfy$\dagger$} &
  68.6 &
  78.9 &
  49.9 &
  67.6 &
  73.2 &
  82.1 &
  79.8 &
  91.6 \\
    \midrule
    \midrule
  
   \multirow{7}{*}{\textbf{\begin{sideways}Supervised\end{sideways}}} 
  & \multirow{4}{*}{{ \textbf{1NN}}} 
{ } &
  { {Context2vec}} &
  { 71.0} &
  { 80.5} &
  { 57.6} &
  { 72.9} &
  { 75.2} &
  { 83.1} &
  { 82.7} &
  { 92.5} \\
  &
{ } &
  { {ELMo}} &
  { 70.9} &
  \multicolumn{1}{l}{{ 80.0}} &
  { 57.3} &
  \multicolumn{1}{l}{{ 73.5}} &
  { 77.4} &
  \multicolumn{1}{l}{{ 85.4}} &
  { 82.4} &
  \multicolumn{1}{l}{{ 92.8}} \\
  &
{ } &
  { {BERT-Base}} &
  { 74.0} &
  \multicolumn{1}{l}{{ 83.0}} &
  { 61.7} &
  \multicolumn{1}{l}{{ 75.3}} &
  { 77.7} &
  \multicolumn{1}{l}{{ 84.9}} &
  { \textbf{85.8}} &
  \multicolumn{1}{l}{{ 93.9}} \\
  &
  &
  { {BERT-Large}} &
  { 75.1} &
  \multicolumn{1}{l}{{ 83.7}} &
  { 63.2} &
  \multicolumn{1}{l}{{ 76.6}} &
  { 79.5} &
  \multicolumn{1}{l}{{ 85.4}} &
  { 85.3} &
  \multicolumn{1}{l}{{ \textbf{94.2}}} \\
  \cmidrule(lr){2-11}
  &
  \multirow{2}{*}{\textbf{SVM}} &
  {IMS} &
  70.4 &
  79.4 &
  56.1 &
  72.5 &
  75.6 &
  84.1 &
  82.9 &
  93.1 \\
  
  &
 &
  {IMS+emb} &
  71.9 &
  80.5 &
  56.9 &
  73.1 &
  75.9 &
  83.8 &
  84.7 &
  93.4 \\
 \midrule
 \midrule
 
 \multicolumn{2}{l}{\multirow{2}{*}{{ \textbf{Hybrid}}}} &
  {LMMS$_{2048}$} &
  { \textbf{78.0}} &
  { \textbf{86.2}} &
  { 64.0} &
  { 76.5} &
  { \textbf{80.7}} &
  { \textbf{86.7}} &
  { 83.5} &
  { 92.8} \\
  
  &
  &
  {KnowBert$\dagger$ \scriptsize{WN+WK}} &
  { 77.0} &
  \multicolumn{1}{l}{{ 85.0}} &
  { \textbf{66.4}} &
  \multicolumn{1}{l}{{ \textbf{78.8}}} &
  { 78.3} &
  \multicolumn{1}{l}{{ 86.1}} &
  { 84.7} &
  \multicolumn{1}{l}{{ 93.9}} \\
  \midrule
  \midrule

  &
\textbf{-} &
  \textit{MFS Baseline} &
  67.6 &
  77.0 &
  49.6 &
  67.2 &
  73.1 &
  82.0 &
  80.5 &
  92.9 \\
  \bottomrule
\end{tabular}
}
\caption{\label{tab:table_pos} F-Measure performance in the concatenation of all datasets of the unified WSD evaluation framework \cite{raganato-etal-2017-word}, split by Part-of-Speech. As in Table \ref{tab:evWSD} systems marked with $\dagger$ make use of external resources other than WordNet.}
\end{table}

The coarsening of sense inventory does help in bridging this gap, with the best models performing in the 75\% ballpark. Nonetheless, the lower performance is again found in verb instances, with noun, adjective and adverb performance being above 80\% on the BERT-based models (above 90\% in the case of adverbs). 
One problem with the existing coarsening methods is that they usually exploit domain-level information, whereas in some cases verbs do not belong to clear domains.
For our example verb \textit{run}, some of the twelve senses denoting motion are clustered into different domains, which eases the task for automatic models due to having fewer number of classes. 
However, one could argue that this clustering is artificial as all senses of the verb belong to the same domain. 

Indeed, while the sense clustering provided by CSI \cite{lacerra2020csi} covers all PoS categories, it extends BabelDomains \cite{babeldomains:2017}, a domain clustering resource that covers mainly nouns. While out of scope for this paper, in the future it would be interesting to investigate verb-specific clustering methods, e.g., \cite{peterson2018bayesian}. 

In the remainder of this article we focus on noun ambiguity, and check the extent to which language models can solve coarse-grained WSD in ideal settings. 
In Section \ref{sec:discussion}, we extend the discussion about sense granularity in WSD.

\section{CoarseWSD-20 Dataset}
\label{sec:coarsewsd-20}

Standard WSD benchmarks mostly rely on WordNet. This makes the evaluations carried out on these datasets and the conclusions drawn from them specific to this resource only. Moreover, sense distinctions in WordNet are generally known to be too fine-grained (see more details about the fine granularity of WordNet in the discussion of Section \ref{sec:discussion}) and annotations are scarce given the knowledge-acquisition bottleneck \cite{Galeetal:92c,pasini2020knowledge}. This prevents from testing the limits of language models in WSD, which is one of the main motivations of this paper.

To this end, we devise a new dataset, CoarseWSD-20 henceforth, in an attempt to solve the aforementioned limitations. CoarseWSD-20 is aimed at providing a benchmark for the qualitative analysis of certain types of easily-interpretable sense distinctions. Our dataset also serves as a tool for testing the limits of WSD models in ideal training scenarios, i.e., with plenty of training data available per word.

In the following we describe the procedure we followed to construct CoarseWSD-20 (Section \ref{cons-coarse-dataset}). Then, we present an estimation of the human performance (Section \ref{sec:human}) and outline some relevant statistics (Section \ref{sec:statistics}). 
Finally, we discuss the out-of-domain test set we built as a benchmark for experiments in Section \ref{outofdomain:results}.

\subsection{Dataset construction}
\label{cons-coarse-dataset}

CoarseWSD-20 targets noun ambiguity\footnote{There are arguably more types of ambiguity, including word categories (e.g., \textit{play} as a noun or as a verb). Nevertheless, this type of ambiguity can be solved to a good extent by using state-of-the-art PoS taggers, which are able to achieve performances above 97\% for English in general settings \cite{akbik-etal-2018-contextual}.} for which, thanks to Wikipedia, data is more easily available. 
The dataset focuses on the coarse-grained disambiguation setting, which is more interpretable by humans \cite{lacerra2020csi}. 
To this end, twenty words\footnote{The main justification to select twenty words (and no more) was the extent of experiments and the computation required to run a deep qualitative analysis (see Section \ref{experimental}). A larger number of words would have prevented us from running the analyses at the depth we envisaged: twenty provided a good trade-off between having a heterogeneous set of words and a deep qualitative analysis.} and their corresponding senses were selected by a group of two expert computational linguists in order to provide a diverse dataset. Wikipedia\footnote{We used the Wikipedia dump of May 2016.} was used as reference inventory and corpus. 
In this case, each Wikipedia page corresponds to an unambiguous sense. 
Sentences where a given Wikipedia page is referred to via a hyperlink are considered to be its corresponding sense-annotated sentences. 
The process to select twenty ambiguous words and their corresponding sense-annotated sentences was as follows:

\begin{enumerate}
     
\item A larger set of a few hundred ambiguous words that had a minimum of thirty occurrences\footnote{This threshold was selected for the goal of testing the language models under close-to-ideal conditions. A real setting should also include senses with even lower frequency, the so called \textit{long tail} \cite{ilievski2018systematic,blevins-zettlemoyer-2020-moving}, which would clearly harm automatic models.} (i.e., sentences where one of their senses is referred to via a hyperlink) was selected. 

\item Two experts selected twenty words based on a variety of criteria: type of ambiguity (e.g., spanning across domains or not), polysemy, overall frequency, distribution of instances across senses of the word, and interpretability. This process was performed semi-automatically, as initially the experts filtered words and senses manually providing a reduced set of words and associated senses. The main goal of this filtering was to discard those senses that were not easily interpretable or distinguishable by humans.

\end{enumerate}

Once these twenty words were selected, we tokenized and lowercased the English Wikipedia and extracted all sentences that contained them and their selected senses as hyperlinks. All sentences were then semi-automatically verified so as to remove duplicate and noisy sentences. Finally, for each word we created a single dataset based on a standard 60/40 train/test split.

\subsection{Human performance estimation}
\label{sec:human}

As explained above, this WSD dataset was designed to be simple for humans to annotate. In other words, the senses considered for CoarseWSD-20 are easily interpretable. As a sanity check, we performed a disambiguation exercise with 1,000 instances randomly sampled from the test set (50 for each word). Four annotators\footnote{All annotators were fluent English speakers and understood the pre-defined senses for their assigned words.} were asked to disambiguate a given target word in context using the CoarseWSD-20 sense inventory. Each annotator completed the task for five words. In the following section we provide details about the results of this annotation exercise, as well as general statistics of CoarseWSD-20.

\subsection{Statistics}
\label{sec:statistics}

Table \ref{per-word-stats} shows the list of words, their associated senses and the frequency of each word sense in CoarseWSD-20, along with the ratio of the first sense with respect to the rest (F2E), normalized entropy\footnote{Computed as $\sum f_i ~ log(f_i)$ normalized by $log(n)$ where $n$ is the number of senses.} (Ent.) and an estimation of the human accuracy (see Section \ref{sec:human}).
The number of senses per word varies from 2 to 5 (eleven words with two associated senses, six with three, two with four and one with five) while the overall frequency ranges from 110 instances (68 for training) for \textit{digit} to 9,240 (6,421 for training) for \textit{pitcher}. As for the human performance, we can see how annotators did not have special difficulty in assigning the right sense for each word in context. Annotators achieve an accuracy of over 96\% in all cases except for a couple of senses with slightly finer grained distinctions such as \textit{club} and \textit{bass}.

Normalized entropy ranges from 0.04 to 0.99 (higher entropy shows more balanced sense distribution). 
While some words contain a roughly balanced distribution of senses (e.g. \textit{crane} or \textit{java}), other words' distribution are highly skewed (see normalized entropy values, e.g., for \textit{pitcher} or \textit{bank}).

Finally, in the appendix we include more information for each of the senses available in CoarseWSD-20, including definitions and an example sentence from the dataset.

\begin{table}[t]
\setlength{\tabcolsep}{4.2pt}
\renewcommand{\arraystretch}{1.2}
\resizebox{\columnwidth}{!}{
\begin{tabular}{ll}
\begin{tabular}{lllllll}
\bf Word & \bf F2R & \bf Ent. & \bf Hum & \bf Senses &   \bf Frequency \\ 
\cmidrule[1pt](lr){1-6}
\multirow{2}{*}{\bf apple}  &   \multirow{2}{*}{1.6} & \multirow{2}{*}{0.96} &  \multirow{2}{*}{100} & apple\_inc      &   1466/634  & \\
                        & & & & apple             &   892/398     &\\
\cmidrule(lr){1-6}
\multirow{2}{*}{\bf arm}    & \multirow{2}{*}{2.8} & \multirow{2}{*}{0.83} &  \multirow{2}{*}{100} &   arm\_architecture    &   311/121    &\\
                        & & & &  arm                &   112/43  & \\ 
\cmidrule(lr){1-6}

\multirow{2}{*}{\bf bank}   &  \multirow{2}{*}{23.1} & \multirow{2}{*}{0.28} &  \multirow{2}{*}{98} & bank            &   1061/433    & \\  
                        & & & &  bank\_(geography)   &   46/22   & \\                          
\cmidrule(lr){1-6}
\multirow{3}{*}{\bf bass}   & \multirow{3}{*}{2.9} & \multirow{3}{*}{0.67}  &  \multirow{3}{*}{90}  &   bass\_guitar        &   2356/1005  & \\
                           & & & &   bass\_(voice\_type) &   609/298 & \\
                        & & &  &   double\_bass        &   208/88  & \\
\cmidrule(lr){1-6}
\multirow{3}{*}{\bf bow}    &   \multirow{3}{*}{1.0} & \multirow{3}{*}{0.87} &  \multirow{3}{*}{98} & bow\_ship           &   266/117 &  \\                           
                        &  & & & bow\_and\_arrow     &   185/72  & \\  
                        &  & & & bow\_(music)        &   72/26 & \\                             
\cmidrule(lr){1-6}
\multirow{2}{*}{\bf chair}      &   \multirow{2}{*}{1.4} & \multirow{2}{*}{0.91} &  \multirow{2}{*}{98} & chairman            &   156/88  & \\
                            & & & &  chair               &   115/42 & \\ 
\cmidrule(lr){1-6}
\multirow{3}{*}{\bf club}   &  \multirow{3}{*}{0.9}  & \multirow{3}{*}{0.85} &   \multirow{3}{*}{86} & club                     &   186/108 & \\
                        & &  & & nightclub               &   148/73    & \\
                        & & & & club\_(weapon)           &   54/21    & \\ 
\cmidrule(lr){1-6}
\multirow{2}{*}{\bf crane}  &   \multirow{2}{*}{1.3} & \multirow{2}{*}{0.99} &   \multirow{2}{*}{98} & crane\_(machine)        &   211/81  & \\
                        & & & &   crane\_(bird)           &   161/76    & \\ 
\cmidrule(lr){1-6}
\multirow{2}{*}{\bf deck}   &  \multirow{2}{*}{8.4} & \multirow{2}{*}{0.37} & \multirow{2}{*}{96} & deck\_(ship)            &  152/92 & \\ 
                        & & & & deck\_(building)        &   18/7 &    \\ 
\cmidrule(lr){1-6}    
\multirow{2}{*}{\bf digit}  & \multirow{2}{*}{2.2} & \multirow{2}{*}{0.74} &   \multirow{2}{*}{100} &  numerical\_digit        &  47/33  & \\
                        & & & &  digit\_(anatomy)        &   21/9    & \\ 
\cmidrule(lr){1-6}

\multirow{3}{*}{\bf hood}   & \multirow{3}{*}{1.6} & \multirow{3}{*}{0.88} &   \multirow{3}{*}{98} &  hood\_(comics)          &   105/47  & \\
                        & & & & hood\_(vehicle)         &   42/13 & \\ 
                        & & & & hood\_(headgear)        &   24/22   & \\ 
\end{tabular}
\begin{tabular}{lllllll}
& \bf Word & \bf F2R & \bf Ent. & \bf Hum & \bf Senses &   \bf Frequency \\ 
\cmidrule[1pt](lr){2-7}    
& \multirow{2}{*}{\bf java}   & \multirow{2}{*}{1.4} & \multirow{2}{*}{0.96} &   \multirow{2}{*}{100} &  java                    &   2641/1180   \\
&                        & & & &   java\_(progr.\_lang.)   &   1863/749    \\ 
 \cmidrule(lr){2-7}                     
&\multirow{5}{*}{\bf mole}    & \multirow{5}{*}{0.4} & \multirow{5}{*}{0.93} &   \multirow{5}{*}{98} &  mole\_(animal)          &   148/77   \\ 
&                        & & & &  mole\_(espionage)       &   120/44  \\
&                        & & & & mole\_(unit)              &   108/42       \\
&                        & & & & mole\_sauce             &   53/23  \\ 
&                        & & & & mole\_(architecture)      &   51/20    \\ 
\cmidrule(lr){2-7}
&\multirow{2}{*}{\bf pitcher}& \multirow{2}{*}{355.7} & \multirow{2}{*}{0.04} &   \multirow{2}{*}{100} &  pitcher                 &   6403/2806   \\
&                        & & & & pitcher\_(container)    &   18/13   \\ 
\cmidrule(lr){2-7}
&\multirow{2}{*}{\bf pound}  & \multirow{2}{*}{6.2} & \multirow{2}{*}{0.48} &   \multirow{2}{*}{100} &  pound\_mass             &   160/87  \\
&                        & & & & pound\_(currency)       &   26/10    \\ 
\cmidrule(lr){2-7}
&\multirow{4}{*}{\bf seal}   & \multirow{4}{*}{0.5} & \multirow{4}{*}{0.87} &   \multirow{4}{*}{100} &  pinniped               &   305/131 \\
&                        & & & &  seal\_(musician)        &   267/106    \\ 
&                        & & & & seal\_(emblem)          &   265/114 \\
&                        & & & & seal\_(mechanical)      &   38/12 \\ 
\cmidrule(lr){2-7}
&\multirow{3}{*}{\bf spring}    & \multirow{3}{*}{0.9} & \multirow{3}{*}{0.91} &   \multirow{3}{*}{100} &  spring\_(hidrology)  &   516/236 \\
&                            & & & &  spring\_(season)    &   389/148    \\ 
&                            & & & & spring\_(device)      &   159/73 \\ 
\cmidrule(lr){2-7}
&\multirow{4}{*}{\bf square}    &  \multirow{4}{*}{1.1} & \multirow{4}{*}{0.83} &   \multirow{4}{*}{96} & square               &   264/103 \\
&                            & & & & square\_(company)     &   167/62  \\ 
&                            & & & & town\_square         &   56/29   \\
&                            & & & & square\_number      &   21/13  \\ 
\cmidrule(lr){2-7}
&\multirow{3}{*}{\bf trunk}     & \multirow{3}{*}{1.3} & \multirow{3}{*}{0.85} &   \multirow{3}{*}{100} &  trunk\_(botany)      &   93/47   \\ 
&                            & & & & trunk\_(automobile) &   36/16  \\ 
&                            & & & & trunk\_(anatomy)    &   35/14  \\ 
\cmidrule(lr){2-7}
&\multirow{2}{*}{\bf yard}       & \multirow{2}{*}{5.3} & \multirow{2}{*}{0.62} &   \multirow{2}{*}{100} &  yard                &   121/61  \\
&                            & & & & yard\_(sailing)     &   23/11    \\
                            
\end{tabular}
\end{tabular}
}
\caption{\label{per-word-stats} Target words and their associated senses, represented by their Wikipedia page title, with their overall associated frequency in CoarseWSD-20 (train/test). \textit{F2R} denotes the ratio of instances for first sense to the rest, while \textit{Ent.} is the normalized entropy of sense distribution. Moreover, the \textit{Human} performance is reported in terms of accuracy.}
\end{table}

\subsection{Out of domain test set}
\label{outofdomain}

The CoarseWSD-20 dataset was constructed exclusively based on Wikipedia. Therefore, the variety of language present in the dataset might be limited. To verify the robustness of WSD models in a different setting, we constructed an out-of-domain test set from existing WordNet-based datasets.

\begin{table}[t]
\centering
\setlength{\tabcolsep}{14.0pt}
\resizebox{1.0\columnwidth}{!}{%
\begin{tabular}{l cccl}
\toprule
&\bf Polysemy 
&\bf Normalized entropy
&\bf No. of instances
&\bf Sense distribution \\
\midrule
bank  &   2  &   0.87  &   48  &  ~~~~~~~~34/14 \\
chair  &   2  &   0.47  &   40  &  ~~~~~~~~4/36 \\
pitcher  &   2  &   0.52  &   17  &  ~~~~~~~~15/2 \\
pound  &   2  &   0.43  &   46  &  ~~~~~~~~42/4 \\
spring  &   3  &   0.63  &   31  &  ~~~~~~~~3/24/4 \\
square  &   3  &   0.49  &   26  &  ~~~~~~~~22/2/2 \\
club  &   2  &   0.39  &   13  &  ~~~~~~~~12/1 \\
\bottomrule
\end{tabular}
}
\caption{Statistics of the out of domain dataset. The two rightmost columns show the number of instances for each of the seven words and their distribution across senses.} \label{tab:out-domain-table}
\end{table}

To construct this test set, we leveraged BabelNet mappings from Wikipedia to WordNet \cite{NavigliPonzetto:12aij} to link the Wikipedia-based CoarseWSD-20 to WordNet senses. After a manual verification of all senses, we retrieved all sentences containing one of the target words in either SemCor \cite{Milleretal:93} or any of the Senseval/SemEval evaluation datasets from \citet{raganato-etal-2017-word}. Finally, we only kept those target words for which all the associated senses were present in the WordNet-based sense annotated corpora and occurred at least 10 times. This resulted in a test set with seven target words (i.e., bank, chair, pitcher, pound, spring, square and club). Table \ref{tab:out-domain-table} shows the relevant statistics of this out-of-domain test set.

\section{Evaluation}
\label{sec:evaluation}

In this section we report on our quantitative evaluation in the coarse-grained WSD setting on CoarseWSD-20. 
We describe the experimental setting in Section \ref{experimental} and then present the main results on CoarseWSD-20 (Section \ref{results}) and the out-of-domain test set (Section \ref{outofdomain:results}).

\subsection{Experimental setting}
\label{experimental}

CoarseWSD-20 consists of 20 separate sets, each containing sentences for different senses of the corresponding target word. 
Therefore, the evaluation can be framed as a standard classification task for each word.

Given the classification nature of the CoarseWSD-20 datasets, we can perform experiments with our 1NN BERT system and compare it with a standard fine-tuned BERT model (see Section \ref{languagemodelsWSD} for more details on the LM-based WSD approaches). 
Note that fine-tuning for individual target words results in many models (one per word). Therefore, this setup would not be computationally feasible in a general WSD setting, as the number of models would approach the vocabulary size. 
However, in our experiments we are interested in verifying the limits of BERT, without any other confounds or model-specific restrictions.

To ensure our conclusions are generalizable, we also report 1NN and fine-tuning results using ALBERT.
In spite of substantial operational differences, BERT and ALBERT have the most similar training objectives and tokenization methods out of several other prominent Transformer-based models \cite{XLNet:2019,liu2019roberta}, thus being the most directly comparable.
Given the similar performance between BERT-Large and ALBERT-XXLarge on the main CoarseWSD-20 dataset, we proceed with further experiments using only BERT.

We also include two FastText linear classifiers \cite{joulin-etal-2017-bag} as baselines: FTX-B (base model without pre-trained embeddings) and FTX-C (using pre-trained embeddings from Common Crawl). We chose FastText as baseline given its efficiency and competitive results for sentence classification.

\paragraph{Configuration} Our experiments with BERT and ALBERT used the Transformers framework (v2.5.1) developed by \citet{Wolf2019HuggingFacesTS}, and we used the uncased pre-trained base and large models released by \citet{devlin-etal-2019-bert} for BERT, and the xxlarge (v2) models released by \citet{lan2019albert} for ALBERT.
We use the uncased variants of Transformers models to match the casing in CoarseWSD-20 (except for ALBERT, which is only available in cased variants).
Following previous feature extraction works (including our experiment in Section \ref{bertmodels}), with CoarseWSD-20 we also average sub-word representations and use the sum of the last four layers when extracting contextual embeddings. For fine-tuning experiments, we used a concatenation of the average embedding of target word's sub-words with the embedding of the [CLS] token, and fed them to a classifier. We used the same default hyper-parameter configuration for all the experiments. Given the fluctuation of results with fine-tuning, all the experiments are based on the average of three independent runs. Our experiments with FastText used the official package\footnote{\url{https://fasttext.cc/}} (v0.9.1), with FastText-Base corresponding to the default supervised classification pipeline using randomly-initialized vectors, and FastText-Crawl corresponding to the same pipeline but starting with pre-trained 300-dimensional vectors based on Common Crawl. Following \citet{joulin-etal-2017-bag}, classification with FastText is performed using multinomial logistic regression and averaged sub-word representations.

\paragraph{Evaluation measures} In a classification setting, the performance of a model is measured by various metrics, among which precision, recall and F-score are the most popular.
Let $TP_i$ (true-positive) and $FP_i$ (false-positive) be the number of instances correctly / incorrectly classified as class $c_i$ respectively. Also, let $TN_i$ (true-negative) and $FN_i$ (false-negative) be the number of instances correctly / incorrectly classified as class $c_j$ for any $j \neq i$. Therefore, for class $c_i$, precision $P_i$ and recall $R_i$ are defined as follows:
\vspace{-20pt}
\begin{multicols}{2}
\begin{equation}
P_i = \frac{TP_i}{TP_i + FP_i}
\end{equation}\break
\begin{equation}
R_i = \frac{TP_i}{TP_i + FN_i}
\end{equation}
\end{multicols}
\noindent In other words, precision is the fraction of relevant instances among the retrieved instances, while recall is the fraction of the total number of relevant instances that were actually retrieved. 
The F-score  $F_i$ for class $c_i$ is then defined as the harmonic mean of its precision and recall values:
\begin{equation}
F_i = \frac{2}{P_i^{-1}+R_i^{-1}} = 2  \frac{P_i . R_i}{P_i + R_i}
\end{equation}
\noindent In order to have a single value to measure the overall performance of the model, we can take the weighted average of these computed values over all the classes, which is referred to as average micro, if the weights are set to be the number of instances for each class, and macro if the weights are set to be equal. For our experiments we mainly report \textit{Macro-F1} and \textit{Micro-F1}.

\paragraph{Number of experiments} To provide an idea of the experiments run on (including the analysis in Section \ref{sec:analysis}, in the following we detail the number of computations required. We evaluated six models, each of them trained and tested separately for each word (there are twenty of them). The same models are also trained with balanced datasets (Section \ref{an:distribution}). In total, 240 models trained and tested for the main results (excluding multiple runs). Then, the computationally more demanding models (BERT-Large) are also evaluated on the out of domain test set, and trained with different training data sizes (Section \ref{trainingsize}) and with fixed number of examples (Section \ref{an:nshot}). In the latter case being BERT-base and FastText models also considered (sometimes with multiple runs). As a rough estimate, all the experiments took over 1500 hours on a Tesla K80 GPU. These experiments do not include the experiments run in the standard benchmarks (Section \ref{evaluation}) and all the extra-analyses and prior experimental tests that did not make into the paper.

\begin{table*}[]
\setlength{\tabcolsep}{0.06in}
\scalebox{0.86}{%
\begin{tabular}{l@{\hspace{0.25in}}cc@{\hspace{0.25in}}cc@{\hspace{0.25in}}ccc@{\hspace{0.25in}}ccc}
\toprule
\multirow{2}{*}{\textbf{Word}} & \multirow{2}{*}{\textbf{Human}} & \multirow{2}{*}{\textbf{MFS}} & \multicolumn{2}{c}{\textbf{Static emb.}} & \multicolumn{3}{c}{\textbf{1NN}} & \multicolumn{3}{c}{\textbf{Fine-tune}~~~~~~~} \\
\cmidrule(r@{0.25in}){4-5}
\cmidrule(r@{0.25in}){6-8}
\cmidrule(r@{0.25in}){9-11}
& & & FTX-B & FTX-C & BRT-B	& BRT-L & ALBRT & BRT-B & BRT-L & ALBRT\\ 
\midrule
\multicolumn{11}{c}{\bf Micro-F1 (Accuracy)} \\
\cmidrule(lr){2-11}
\textbf{crane}   &  \textit{98.0}                        & 51.6 & \cellcolor[HTML]{F8DCD9}91.7 & \cellcolor[HTML]{FCF0EF}94.9  & \cellcolor[HTML]{FAE8E6}93.6  & \cellcolor[HTML]{FEFCFC}96.8  & \cellcolor[HTML]{CBEADB}98.1  & \cellcolor[HTML]{F2FAF6}97.5  & \cellcolor[HTML]{CBEADB}98.1  & \cellcolor[HTML]{FEFCFC}96.8  \\
\textbf{java}    &            \textit{100}~~              & 61.2 & \cellcolor[HTML]{A0D9BD}98.8 & \cellcolor[HTML]{7DCBA5}99.4  & \cellcolor[HTML]{6EC49A}99.6  & \cellcolor[HTML]{71C69C}99.6  & \cellcolor[HTML]{71C69C}99.6  & \cellcolor[HTML]{67C295}99.7  & \cellcolor[HTML]{6AC398}99.7  & \cellcolor[HTML]{75C79F}99.5  \\
\textbf{apple}   & \textit{100}~~                  & 61.4 & \cellcolor[HTML]{FEFAFA}96.5 & \cellcolor[HTML]{BBE4D0}98.4  & \cellcolor[HTML]{92D3B3}99.0  & \cellcolor[HTML]{87CFAB}99.2  & \cellcolor[HTML]{7BCAA3}99.4  & \cellcolor[HTML]{71C69C}99.6  & \cellcolor[HTML]{6FC59B}99.6  & \cellcolor[HTML]{81CCA7}99.3  \\
\textbf{mole}    &   \textit{98.0}                       & 37.4 & \cellcolor[HTML]{F3C1BC}87.4 & \cellcolor[HTML]{FAE5E3}93.2  & \cellcolor[HTML]{FEFEFE}97.1  & \cellcolor[HTML]{B0DFC8}98.5  & \cellcolor[HTML]{CDEBDD}98.1  & \cellcolor[HTML]{9CD7BA}98.9  & \cellcolor[HTML]{9CD7BA}98.9  & \cellcolor[HTML]{B0DFC8}98.5  \\
\textbf{spring}  & \textit{100}~~                         & 51.6 & \cellcolor[HTML]{F8DDDB}91.9 & \cellcolor[HTML]{FBEEEC}94.5  & \cellcolor[HTML]{F7FCF9}97.4  & \cellcolor[HTML]{DCF1E7}97.8  & \cellcolor[HTML]{7FCCA6}99.3  & \cellcolor[HTML]{CFECDE}98.0  & \cellcolor[HTML]{BDE5D1}98.3  & \cellcolor[HTML]{C2E6D4}98.2  \\
\textbf{chair}   & \textit{98.0} & 67.7 & \cellcolor[HTML]{EC9C95}81.5 & \cellcolor[HTML]{F4C7C4}88.5  & \cellcolor[HTML]{FDF8F7}96.2  & \cellcolor[HTML]{FDF8F7}96.2  & \cellcolor[HTML]{FCF3F2}95.4  & \cellcolor[HTML]{FEFBFB}96.7  & \cellcolor[HTML]{FDF8F7}96.2  & \cellcolor[HTML]{FBEBE9}94.1  \\
\textbf{hood}    & \textit{98.0}                   & 57.3 & \cellcolor[HTML]{EA958E}80.5 & \cellcolor[HTML]{F5CBC7}89.0  & \cellcolor[HTML]{A1D9BE}98.8  & \cellcolor[HTML]{57BB8A}100~~ & \cellcolor[HTML]{A1D9BE}98.8  & \cellcolor[HTML]{D3EDE0}98.0  & \cellcolor[HTML]{70C69C}99.6  & \cellcolor[HTML]{A1D9BE}98.8  \\
\textbf{seal}    & \textit{100}~~                  & 36.1 & \cellcolor[HTML]{F4C9C5}88.7 & \cellcolor[HTML]{FCF1F0}95.0  & \cellcolor[HTML]{FEF9F9}96.4  & \cellcolor[HTML]{CCEBDC}98.1  & \cellcolor[HTML]{EEF8F3}97.5  & \cellcolor[HTML]{95D4B5}99.0  & \cellcolor[HTML]{95D4B5}99.0  & \cellcolor[HTML]{BCE4D0}98.3  \\
\textbf{bow}     & \textit{98.0}                   & 54.4 & \cellcolor[HTML]{F6D0CC}89.8 & \cellcolor[HTML]{FDF6F5}95.8  & \cellcolor[HTML]{FDF9F8}96.3  & \cellcolor[HTML]{FCF3F2}95.3  & \cellcolor[HTML]{FEFBFB}96.7  & \cellcolor[HTML]{EEF8F3}97.5  & \cellcolor[HTML]{B6E2CC}98.5  & \cellcolor[HTML]{E5F5ED}97.7  \\
\textbf{club}    &                 \textit{86.0}         & 53.5 & \cellcolor[HTML]{E98D85}79.2 & \cellcolor[HTML]{EB978F}80.7  & \cellcolor[HTML]{EB9A93}81.2  & \cellcolor[HTML]{F0B3AD}85.1  & \cellcolor[HTML]{EDA39D}82.7  & \cellcolor[HTML]{F0B3AD}85.2  & \cellcolor[HTML]{EFAFAA}84.7  & \cellcolor[HTML]{EFADA8}84.3  \\
\textbf{trunk}   & \textit{100}~~                  & 61.0 & \cellcolor[HTML]{EFAEA8}84.4 & \cellcolor[HTML]{F7D7D4}90.9  & \cellcolor[HTML]{FDF7F7}96.1  & \cellcolor[HTML]{A6DBC1}98.7  & \cellcolor[HTML]{A6DBC1}98.7  & \cellcolor[HTML]{DBF1E6}97.8  & \cellcolor[HTML]{C0E6D4}98.3  & \cellcolor[HTML]{8CD1AF}99.1  \\
\textbf{square}  & \textit{96.0}                   & 49.8 & \cellcolor[HTML]{F2BEB9}87.0 & \cellcolor[HTML]{F6D3D0}90.3  & \cellcolor[HTML]{FCF2F1}95.2  & \cellcolor[HTML]{FDF8F7}96.1  & \cellcolor[HTML]{FBEBEA}94.2  & \cellcolor[HTML]{FDF6F5}95.8  & \cellcolor[HTML]{FDF5F4}95.7  & \cellcolor[HTML]{FEFAF9}96.5  \\
\textbf{arm}     & \textit{100}~~                  & 73.8 & \cellcolor[HTML]{FBEDEC}94.5 & \cellcolor[HTML]{C6E8D8}98.2  & \cellcolor[HTML]{7CCAA4}99.4  & \cellcolor[HTML]{7CCAA4}99.4  & \cellcolor[HTML]{7CCAA4}99.4  & \cellcolor[HTML]{7DCAA4}99.4  & \cellcolor[HTML]{7DCAA4}99.4  & \cellcolor[HTML]{70C59C}99.6  \\
\textbf{digit}   & \textit{100}~~                  & 78.6 & \cellcolor[HTML]{F9E3E1}92.9 & \cellcolor[HTML]{57BB8A}100~~ & \cellcolor[HTML]{57BB8A}100~~ & \cellcolor[HTML]{57BB8A}100~~ & \cellcolor[HTML]{57BB8A}100~~ & \cellcolor[HTML]{87CFAC}99.2  & \cellcolor[HTML]{57BB8A}100~~ & \cellcolor[HTML]{57BB8A}100~~ \\
\textbf{bass}    & \textit{90.0}                   & 72.3 & \cellcolor[HTML]{FAE9E8}93.9 & \cellcolor[HTML]{FBECEA}94.2  & \cellcolor[HTML]{EB9790}80.7  & \cellcolor[HTML]{EFAEA9}84.5  & \cellcolor[HTML]{F0B5B0}85.5  & \cellcolor[HTML]{FCF4F3}95.5  & \cellcolor[HTML]{FDF6F5}95.8  & \cellcolor[HTML]{FDF5F4}95.7  \\
\textbf{yard}    & \textit{100}~~                  & 84.7 & \cellcolor[HTML]{F1B9B4}86.1 & \cellcolor[HTML]{FBEDEC}94.4  & \cellcolor[HTML]{E67C73}76.4  & \cellcolor[HTML]{F4CAC6}88.9  & \cellcolor[HTML]{F9E4E2}93.1  & \cellcolor[HTML]{ACDEC5}98.6  & \cellcolor[HTML]{73C79E}99.5  & \cellcolor[HTML]{74C79E}99.5  \\
\textbf{pound}   & \textit{100}~~                  & 89.7 & \cellcolor[HTML]{F3C2BE}87.6 & \cellcolor[HTML]{F3C2BE}87.6  & \cellcolor[HTML]{F2BCB7}86.6  & \cellcolor[HTML]{F5CFCC}89.7  & \cellcolor[HTML]{FDF6F5}95.9  & \cellcolor[HTML]{FCF0EF}94.9  & \cellcolor[HTML]{FCF0EF}94.9  & \cellcolor[HTML]{FEFAFA}96.6  \\
\textbf{deck}    & \textit{96.0}                   & 92.9 & \cellcolor[HTML]{F8DDDB}91.9 & \cellcolor[HTML]{FBEAE8}93.9  & \cellcolor[HTML]{F6D0CD}89.9  & \cellcolor[HTML]{F8DDDB}91.9  & \cellcolor[HTML]{FCF0EF}94.9  & \cellcolor[HTML]{FEFBFA}96.6  & \cellcolor[HTML]{FCF2F1}95.3  & \cellcolor[HTML]{FEFDFD}97.0  \\
\textbf{bank}    & \textit{98.0}                   & 95.2 & \cellcolor[HTML]{FEFDFC}96.9 & \cellcolor[HTML]{CFECDE}98.0  & \cellcolor[HTML]{72C69D}99.6  & \cellcolor[HTML]{65C194}99.8  & \cellcolor[HTML]{65C194}99.8  & \cellcolor[HTML]{6EC59A}99.6  & \cellcolor[HTML]{84CDA9}99.3  & \cellcolor[HTML]{80CCA6}99.3  \\
\textbf{pitcher} &           \textit{100}~~               & 99.5 & \cellcolor[HTML]{6DC499}99.6 & \cellcolor[HTML]{69C296}99.7  & \cellcolor[HTML]{5EBE8F}99.9  & \cellcolor[HTML]{5CBD8D}99.9  & \cellcolor[HTML]{57BB8A}100~~ & \cellcolor[HTML]{5BBD8D}100~~ & \cellcolor[HTML]{5BBD8D}100~~ & \cellcolor[HTML]{61BF91}99.8  \\
\midrule
\textbf{AVG}     &                          & 66.5 & 90.0                         & 93.8                          & 94.0                          & 95.8                          & 96.4                          & 97.4                          & 97.5                          & 97.4\\
\bottomrule
\toprule
\multicolumn{11}{c}{\bf Macro-F1} \\
\cmidrule(lr){2-11}
\textbf{crane}   &       --               & 34.0 & \cellcolor[HTML]{FCF1F0}91.7 & \cellcolor[HTML]{FDF9F9}94.8  & \cellcolor[HTML]{FDF6F5}93.5  & \cellcolor[HTML]{FEFEFE}96.7  & \cellcolor[HTML]{B9E3CE}98.1  & \cellcolor[HTML]{DBF1E6}97.5 & \cellcolor[HTML]{BAE3CF}98.1  & \cellcolor[HTML]{FCFEFD}96.8  \\
\textbf{java}    &    --                  & 38.0 & \cellcolor[HTML]{99D6B8}98.7 & \cellcolor[HTML]{78C9A1}99.4  & \cellcolor[HTML]{68C296}99.7  & \cellcolor[HTML]{6AC398}99.6  & \cellcolor[HTML]{6AC398}99.6  & \cellcolor[HTML]{65C194}99.7 & \cellcolor[HTML]{69C296}99.7  & \cellcolor[HTML]{72C69D}99.5  \\
\textbf{apple}   &  --                    & 38.1 & \cellcolor[HTML]{FEFDFD}96.2 & \cellcolor[HTML]{B7E2CD}98.1  & \cellcolor[HTML]{8AD0AE}99.0  & \cellcolor[HTML]{84CEAA}99.1  & \cellcolor[HTML]{7AC9A2}99.3  & \cellcolor[HTML]{6EC59A}99.6 & \cellcolor[HTML]{6DC499}99.6  & \cellcolor[HTML]{7CCAA4}99.3  \\
\textbf{mole}    &      --                & 10.9 & \cellcolor[HTML]{F8DDDA}84.4 & \cellcolor[HTML]{FBEFEE}91.0  & \cellcolor[HTML]{D1EDDF}97.6  & \cellcolor[HTML]{8AD0AD}99.0  & \cellcolor[HTML]{ACDEC6}98.4  & \cellcolor[HTML]{8ED2B1}98.9 & \cellcolor[HTML]{83CDA9}99.2  & \cellcolor[HTML]{98D6B7}98.8  \\
\textbf{spring}  &        --              & 22.7 & \cellcolor[HTML]{FCEFEE}91.1 & \cellcolor[HTML]{FEF9F9}94.9  & \cellcolor[HTML]{DCF1E7}97.4  & \cellcolor[HTML]{CAEADA}97.8  & \cellcolor[HTML]{82CDA8}99.2  & \cellcolor[HTML]{C7E9D8}97.8 & \cellcolor[HTML]{BAE3CF}98.1  & \cellcolor[HTML]{B2E0CA}98.2  \\
\textbf{chair}   & -- & 40.4 & \cellcolor[HTML]{F6CFCC}79.5 & \cellcolor[HTML]{F9E2E1}86.5  & \cellcolor[HTML]{FDF9F8}94.7  & \cellcolor[HTML]{FDF9F8}94.7  & \cellcolor[HTML]{FDF9F9}94.7  & \cellcolor[HTML]{FEFDFD}96.1 & \cellcolor[HTML]{FEFBFB}95.5  & \cellcolor[HTML]{FDF5F4}93.3  \\
\textbf{hood}    &          --            & 24.3 & \cellcolor[HTML]{F1B7B2}70.5 & \cellcolor[HTML]{F7DAD7}83.2  & \cellcolor[HTML]{A6DBC1}98.5  & \cellcolor[HTML]{57BB8A}100~~ & \cellcolor[HTML]{A6DBC1}98.5  & \cellcolor[HTML]{C8E9D9}97.8 & \cellcolor[HTML]{6BC398}99.6  & \cellcolor[HTML]{B0DFC8}98.3  \\
\textbf{seal}    &          --            & 13.3 & \cellcolor[HTML]{F2BDB8}72.7 & \cellcolor[HTML]{FCF3F2}92.6  & \cellcolor[HTML]{E2F3EB}97.3  & \cellcolor[HTML]{A4DAC0}98.5  & \cellcolor[HTML]{BCE4D0}98.1  & \cellcolor[HTML]{8ED2B1}98.9 & \cellcolor[HTML]{9ED8BC}98.6  & \cellcolor[HTML]{C5E8D7}97.9  \\
\textbf{bow}     &          --            & 23.5 & \cellcolor[HTML]{F7DAD7}83.3 & \cellcolor[HTML]{FDF6F6}93.7  & \cellcolor[HTML]{F2FAF6}97.0  & \cellcolor[HTML]{FEFCFC}95.7  & \cellcolor[HTML]{E3F4EC}97.3  & \cellcolor[HTML]{D7EFE4}97.5 & \cellcolor[HTML]{A0D9BD}98.6  & \cellcolor[HTML]{FEFFFE}96.8  \\
\textbf{club}    &          --            & 23.2 & \cellcolor[HTML]{F2BEBA}73.2 & \cellcolor[HTML]{F6D2CF}80.5  & \cellcolor[HTML]{F8DDDB}84.6  & \cellcolor[HTML]{FAE9E7}88.7  & \cellcolor[HTML]{F9E4E2}87.1  & \cellcolor[HTML]{F8DCDA}84.3 & \cellcolor[HTML]{F8DCDA}84.1  & \cellcolor[HTML]{F8DCD9}84.0  \\
\textbf{trunk}   &           --           & 25.3 & \cellcolor[HTML]{F4C6C2}76.0 & \cellcolor[HTML]{F9E1DF}85.9  & \cellcolor[HTML]{C5E8D7}97.9  & \cellcolor[HTML]{7CCAA4}99.3  & \cellcolor[HTML]{7CCAA4}99.3  & \cellcolor[HTML]{D4EEE1}97.6 & \cellcolor[HTML]{BEE5D2}98.0  & \cellcolor[HTML]{8AD0AE}99.0  \\
\textbf{square}  &          --            & 16.6 & \cellcolor[HTML]{EFAFAA}67.7 & \cellcolor[HTML]{F4C7C3}76.3  & \cellcolor[HTML]{FCF3F2}92.5  & \cellcolor[HTML]{FDF9F9}94.7  & \cellcolor[HTML]{FBEBEA}89.7  & \cellcolor[HTML]{FCF2F1}92.2 & \cellcolor[HTML]{FCF0EF}91.4  & \cellcolor[HTML]{FDF6F5}93.5  \\
\textbf{arm}     &           --           & 42.5 & \cellcolor[HTML]{FCF3F2}92.5 & \cellcolor[HTML]{BEE5D2}98.0  & \cellcolor[HTML]{6DC499}99.6  & \cellcolor[HTML]{6DC499}99.6  & \cellcolor[HTML]{6DC499}99.6  & \cellcolor[HTML]{80CCA7}99.2 & \cellcolor[HTML]{80CCA7}99.2  & \cellcolor[HTML]{72C69D}99.5  \\
\textbf{digit}   &            --          & 44.0 & \cellcolor[HTML]{F7DAD7}83.3 & \cellcolor[HTML]{57BB8A}100~~ & \cellcolor[HTML]{57BB8A}100~~ & \cellcolor[HTML]{57BB8A}100~~ & \cellcolor[HTML]{57BB8A}100~~ & \cellcolor[HTML]{97D5B7}98.8 & \cellcolor[HTML]{57BB8A}100~~ & \cellcolor[HTML]{57BB8A}100~~ \\
\textbf{bass}    &          --            & 28.0 & \cellcolor[HTML]{F6D1CE}80.2 & \cellcolor[HTML]{F6D4D1}81.3  & \cellcolor[HTML]{F5CECB}79.1  & \cellcolor[HTML]{F8DCD9}84.0  & \cellcolor[HTML]{F9E4E2}87.1  & \cellcolor[HTML]{FAE5E3}87.5 & \cellcolor[HTML]{FAE5E4}87.6  & \cellcolor[HTML]{F9E4E2}86.9  \\
\textbf{yard}    &          --            & 45.9 & \cellcolor[HTML]{E88B83}54.5 & \cellcolor[HTML]{F7D6D3}81.8  & \cellcolor[HTML]{F9E1DF}86.1  & \cellcolor[HTML]{FDF5F5}93.4  & \cellcolor[HTML]{FEFCFC}95.9  & \cellcolor[HTML]{E7F6EF}97.2 & \cellcolor[HTML]{87CFAC}99.1  & \cellcolor[HTML]{87CFAC}99.1  \\
\textbf{pound}   &           --           & 47.3 & \cellcolor[HTML]{E67C73}48.9 & \cellcolor[HTML]{E8887F}53.3  & \cellcolor[HTML]{FCF3F2}92.5  & \cellcolor[HTML]{FDF8F7}94.3  & \cellcolor[HTML]{CEECDD}97.7  & \cellcolor[HTML]{F8DDDA}84.4 & \cellcolor[HTML]{F8DBD9}83.9  & \cellcolor[HTML]{FBEDEC}90.4  \\
\textbf{deck}    &            --          & 48.2 & \cellcolor[HTML]{E98F88}56.1 & \cellcolor[HTML]{EA928B}57.1  & \cellcolor[HTML]{FAE6E5}88.0  & \cellcolor[HTML]{FEFCFB}95.7  & \cellcolor[HTML]{F8DCD9}84.1  & \cellcolor[HTML]{F8DAD7}83.4 & \cellcolor[HTML]{F5CBC8}78.0  & \cellcolor[HTML]{F8DFDD}85.2  \\
\textbf{bank}    &          --            & 48.8 & \cellcolor[HTML]{F0B0AB}68.2 & \cellcolor[HTML]{F6CFCC}79.5  & \cellcolor[HTML]{FEFBFB}95.5  & \cellcolor[HTML]{CDEBDC}97.7  & \cellcolor[HTML]{CDEBDC}97.7  & \cellcolor[HTML]{C2E7D5}97.9 & \cellcolor[HTML]{FEFBFB}95.6  & \cellcolor[HTML]{FEFDFD}96.3  \\
\textbf{pitcher} &           --           & 49.9 & \cellcolor[HTML]{EC9E98}61.5 & \cellcolor[HTML]{F0B3AE}69.2  & \cellcolor[HTML]{5ABD8C}99.9  & \cellcolor[HTML]{59BC8C}100~~ & \cellcolor[HTML]{57BB8A}100~~ & \cellcolor[HTML]{E2F3EB}97.3 & \cellcolor[HTML]{E4F5ED}97.3  & \cellcolor[HTML]{FBEAE8}89.2  \\
\midrule
\textbf{AVG}     &           --           & 33.2 & 76.5                         & 84.9                          & 94.5                          & 96.4                          & 96.1                          & 95.2                         & 95.1                          & 95.1  \\               \bottomrule        
\end{tabular}
}
\caption{Micro-F1 (top) and macro-F1 (bottom) performance on the full CoarseWSD-20 dataset for eight different models: FastText-Base (FTX-B) and -Crawl (FTX-C), 1NN and fine-tuned BERT-Base (BRT-B), -Large (BRT-L), and ALBERT-XXL (ALBRT). An estimation of the human performance (see Section \ref{sec:human} for more details) and the most frequent sense (MFS) baseline are also reported for each word.
Rows in each table are sorted by the entropy of sense distribution (see Table \ref{per-word-stats}), in descending order. Table cells are highlighted (from red to green) for better interpretability.}
\label{tab:evCoarseWSD}
\end{table*}

\subsection{Results}
\label{results}

Word-specific results for different configurations of BERT and ALBERT as well as the FastText baseline are shown in Table \ref{tab:evCoarseWSD}. In general, results are high for all Transformer-based models, over 90\% in most cases. This reinforces the potential of language models for WSD, both in its light-weight 1NN and in the fine-tuning settings. 
While BERT-Large slightly improves over BERT-Base, the performance of the former is very similar to that of ALBERT-XXL across different configurations, despite having different architectures, number of parameters, and training objectives.
Overall, performance variations in different models are similar to those for the human baseline. For instance, words such as \textit{java} and \textit{digit} seem easy for both humans and models to disambiguate, whereas words such as \textit{bass} and \textit{club} are challenging perhaps due to their more fine-grained distinctions.\footnote{Given that the human performance is estimated based on a small subset of the test set, and given the skewed distribution of sense frequencies, macro-F1 values can be highly sensitive to less-frequent senses (which might even have no instance in the subset); hence, we do not report macro-F1 for human performance.}
As a perhaps surprising result, having more training instances does not necessarily lead to better performance, indicated by the very low Pearson correlation (0.2 or lower) of the number of training instances with results in all BERT configurations. Also, higher polysemy is not a strong indicator of lower performance (see Table \ref{sec:statistics} for statistics of the twenty words, including polysemy), as one would expect from a classification task with higher number of classes (near zero average correlation across settings). In the following we also discuss other relevant points with respect to Most Frequent Sense (MFS) bias and fine-tuning.

\paragraph{MFS Bias}
    As expected, macro F1 results degrade for the purely supervised classification models (FastText and fine-tuned BERT), indicating the inherent sense biases captured by the model which lead to lowered performance for the obscure senses (see the work by \citet{postma-etal-2016-addressing} for a more thorough analysis on this issue).
    However, BERT proves to be much more robust with this respect whereas FastText suffers heavily (highlighted in the macro setting).

\paragraph{Impact of fine-tuning} 
   By average, fine-tuning improves the performance for BERT-Large by 1.6 points in terms of micro-F1 (from 95.8\% to 97.5\%) but decreases on macro-F1 (from 96.4\% to 95.1\%). While BERT-Base significantly correlates with BERT-Large in the 1NN setting (Pearson correlation above 0.9 for both micro and macro), it has a relatively low correlation with the fine-tuned BERT-Base (0.60 on Micro-F1 and 0.75 on macro-F1). The same trend is observed for BERT-Large, where the correlation between fine-tuning and 1NN  is 0.71 and 0.63 on micro-F1 and macro-F1, respectively.
  The operating principles behind both approaches are significantly different, which may explain this relatively low correlation. While fine-tuning is optimizing a loss function during training, the 1NN approach is simply memorizing states. By optimizing losses, fine-tuning is more susceptible to overfit on the MFS. In contrast, by memorizing states, 1NN models senses independently and disregards sense distributions entirely. These differences can explain the main discrepancies between the two strategies, reflected for both micro and macro scores (macro F1 penalizes models which are not as good for less frequent senses). 
  The differences between 1NN and fine-tuned models will be analyzed in more detail in our analysis section (Section \ref{sec:analysis}).
   
   In our error analysis we will show, among others, that there are some cases which are difficult even for humans to disambiguate, e.g., the intended meaning of {\it apple} (fruit vs. company) or {\it club} (nightclub vs. association) in the following contexts taken from the test set: ``it also likes apple'' and ``she was discovered in a club by the record producer peter harris''.

\subsection{Out of domain}
\label{outofdomain:results}

To verify the robustness of BERT and to see if the conclusions can be extended to other settings, we carried out a set of cross-domain evaluations in which the same BERT models (trained on CoarseWSD-20) were evaluated on the out-of-domain dataset described in Section \ref{outofdomain}.

\begin{table}[t]
\setlength{\tabcolsep}{10.0pt}
\resizebox{\columnwidth}{!}{%
\begin{tabular}{l rrrr | rrrr}
\toprule

 &
\multicolumn{4}{c}{\bf Micro F1} &
\multicolumn{4}{c}{\bf Macro F1} \\
\cmidrule(lr){2-5}
\cmidrule(lr){6-9}
&
\multicolumn{2}{c}{\bf 1NN}   &
\multicolumn{2}{c}{\bf  F-Tune}  &
\multicolumn{2}{c}{\bf 1NN} &
\multicolumn{2}{c}{\bf F-Tune} \\ 
\cmidrule(lr){2-3}
\cmidrule(lr){4-5}
\cmidrule(lr){6-7}
\cmidrule(lr){8-9}

  & \multicolumn{1}{r}{BRT-B}   &   \multicolumn{1}{r}{BRT-L}   &   \multicolumn{1}{r}{BRT-B}   & 
  \multicolumn{1}{r}{BRT-L}   & 
  \multicolumn{1}{r}{BRT-B}     &
   \multicolumn{1}{r}{BRT-L}   &   \multicolumn{1}{r}{BRT-B}   &
  \multicolumn{1}{r}{BRT-L} \\ 
\midrule
bank	&	97.9	&	100	&  92.4 & 93.1 &	96.4	&	100 &  89.8 & 	90.5 \\
chair	&	100	    &	100	&  98.3 & 99.2 &	100	&	100 & 94.8 & 	97.4 \\
pitcher	&	82.4	&	100	&  100 & 100 &	90.0 &	100 & 100 & 	100 \\
pound	&	89.1	&	87.0 & 96.4 & 94.9 &	94.0    &	81.5 & 85.5 & 	77.5 \\
spring	&	100		&	96.8 & 94.6	& 96.8 &	100     &	91.7 & 91.2 & 	90.5 \\
square	&	73.1	&	73.1 & 93.6	& 96.2 &	89.4    &	89.4 & 83.2 & 	92.6 \\
club	&	100	    &	100	 & 100 & 100 &	100   &   100 & 100 & 	100\\
\midrule
AVG     &   91.8    &   93.8   &  96.5 & \textbf{97.2}  &   \textbf{95.7}    &   94.7 &   92.1 & 	92.6 \\
\bottomrule
\end{tabular}
}
\caption{Out of domain WSD results: Models trained on the CoarseWSD-20 training set and tested on the out-of-domain test set.}
\label{tab:domain-table}
\end{table}

Table \ref{tab:domain-table} shows the results.
The performance trend is largely in line with that presented in Table \ref{tab:evCoarseWSD}, with some cases even having higher performance in this out-of-domain test set. 
Despite the relatively limited size of this test set, these results seem to corroborate previous findings and highlight the generalization capability of language models to perform WSD in different contexts. 
The fine-tuned version of BERT clearly achieves the highest micro-F1 scores, in line with previous experiments.
Perhaps more surprisingly, BERT-Base 1NN achieves the best macro F1 performance, also highlighting its competitiveness with respect to BERT-Large in this setting.
As explained before, the 1NN strategy seems less prone to biases than the fine-tuned model, and this experiment shows the same conclusion extends to domain specificity as well, therefore the higher figures according to the macro metric.
Interestingly, BERT-Base produces better results according to macro-F1 in the 1NN setting, despite lagging behind according to micro-F1. 
This suggests that data-intensive methods (e.g., fine-tuning) do not generally lead to significantly better results.
Indeed, the results in Table \ref{tab:evCoarseWSD} also confirm that the gains using a larger BERT model are not massive.

\section{Analysis}
\label{sec:analysis}

In this section we perform an analysis on different aspects relevant to WSD on the CoarseWSD-20 dataset. In particular, we first present a qualitative analysis on the type of contextualized embeddings learned by BERT (Section \ref{an:contemb}) and then analyze the impact of sense distribution of the training data (Section \ref{an:distribution}) as well as its size (Section \ref{an:nshot}) on WSD performance. 
Finally, we carry out an analysis on the inherent sense biases present in the pre-trained BERT models (Section \ref{sec:bias}).

\begin{figure}[t]
\centering
   \includegraphics[trim={0 -1cm 0 0},width=1.0\textwidth]{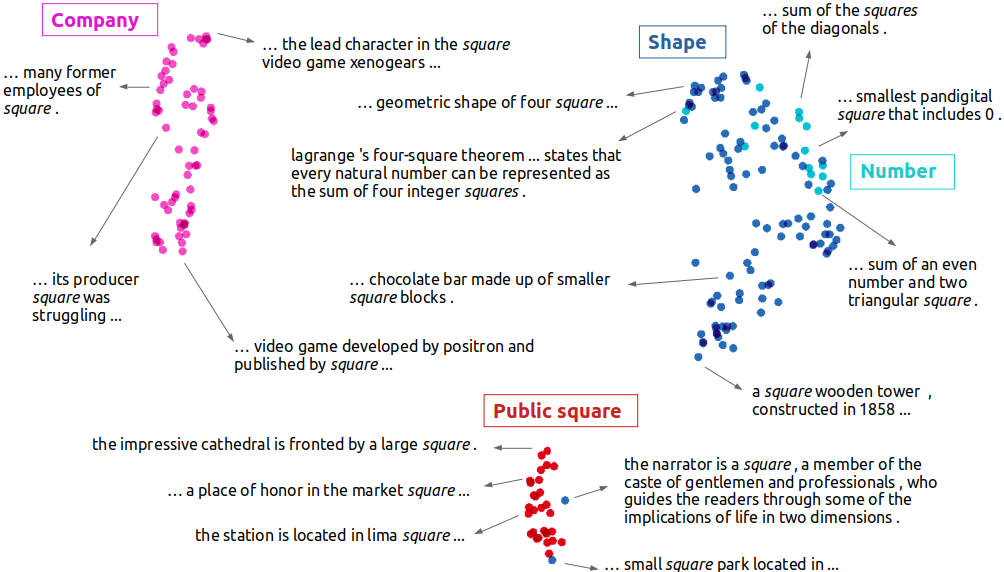}
   \caption{2-D visualizations of contextualized representations for different occurrences of  \textit{square} in the test set. While the company and public-square senses are grouped into distinct clusters, the numerical and geometrical meanings mostly overlap. Using UMAP for dimensionality reduction. \label{fig:square-space}}
\end{figure}

\begin{figure}[t]
\centering
   \includegraphics[trim={0 -1cm 0 0},width=1.0\textwidth]{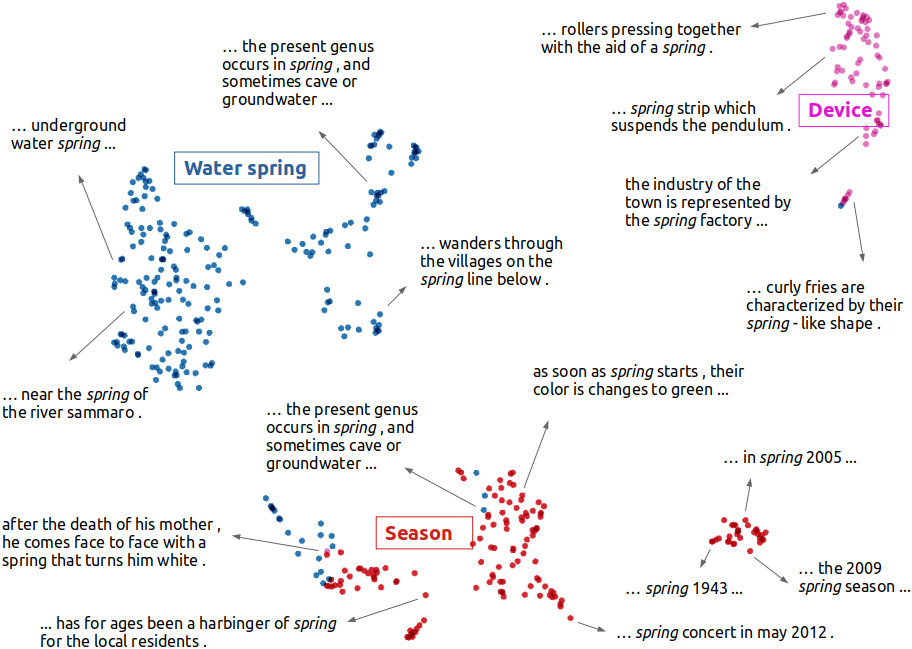}
   \caption{2-D visualizations of contextualized representations for different occurrences of  \textit{spring}. A fine-grained distinction can be observed for the season meaning of \textit{spring}, with a distinct cluster (on the right) denoting the spring of a specific year. Using PCA for dimensionality reduction. \label{fig:spring-space}}
\end{figure}

\begin{figure}[t]
\centering
\includegraphics[width=1.0\linewidth]{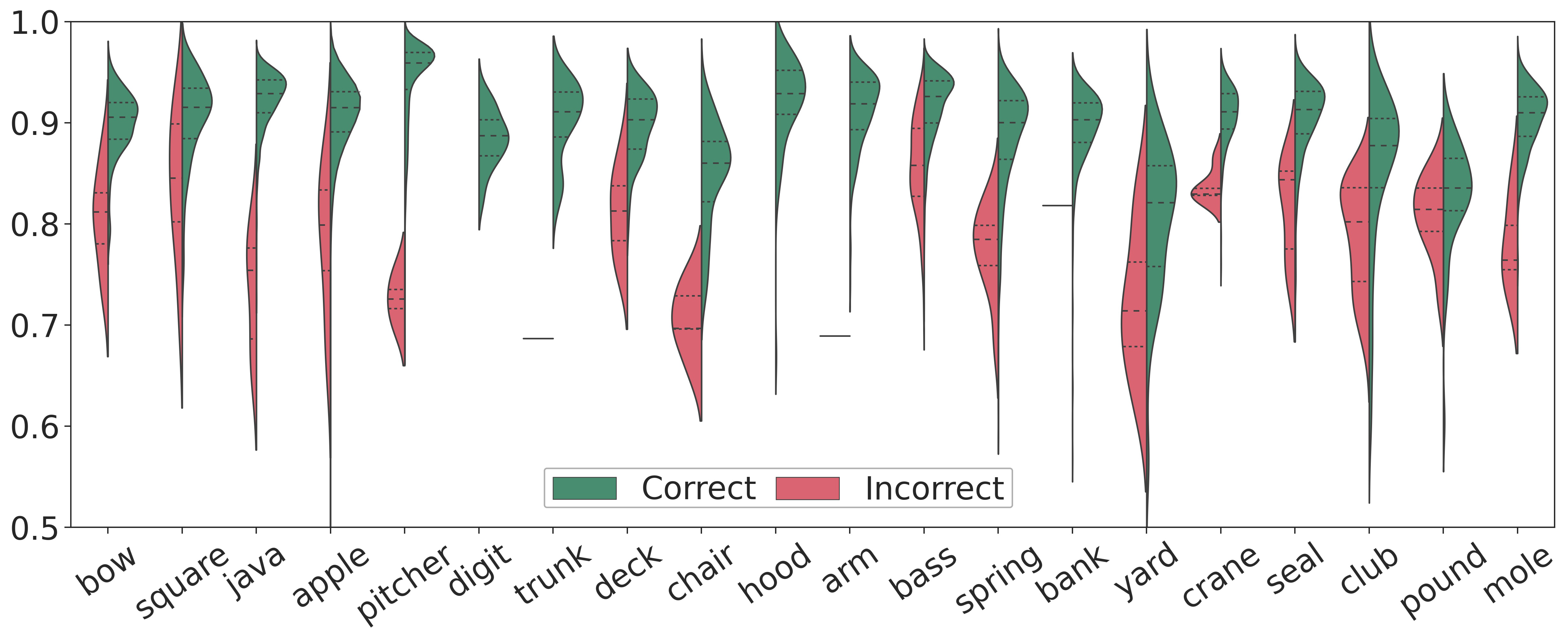}
\caption{Distribution of cosine similarities between contextual embeddings (BERT-Large) of words to be disambiguated (in test set) and their corresponding closest sense embeddings learned from training data, for each word in the CoarseWSD-20 dataset, grouped by correct and incorrect prediction.
\label{fig:similarities}}
\end{figure}

\subsection{Contextualized embeddings}
\label{an:contemb}

The strong performance of BERT-based 1NN WSD method reported for both fine and coarse-grained WSD proves that the representations produced by BERT are sufficiently precise to allow for effective disambiguation. 
Figures \ref{fig:square-space} and \ref{fig:spring-space} illustrate the 2-D semantic space for contextualized representations of two target words (\textit{square} and {\it spring}) in the test set.
For each case, we applied the dimensionality technique that produced the most interpretable visualization, considering UMAP \cite{McInnes2018} and Principal Component Analysis (PCA), although similar observations could be made using either of these two techniques.
BERT is able to correctly distinguish and place most occurrences in distinct clusters. Few challenging exceptions exist, e.g., two geometric senses of \textit{square} are misclassified as public-square, highlighted in the figure (``... small \textit{square} park located in ...'' and `` ... the narrator is a \textit{square} ...'').
Another interesting observation is for the season meaning of \textit{spring}. BERT not only places all the contextualized representations for this sense in the same proximity in the space, it also makes a fine-grained distinction for the spring season of a specific year (e.g., ``... in \textit{spring} 2005 ...'').

Beyond simply checking whether the nearest neighbor corresponds to the correct sense, there is still the question of the extent to which these representations are differentiated. 
In order to quantitatively analyse this, we plotted the distribution of cosine similarities between the contextual embeddings of the target word (to be disambiguated) from the test set and the closest predicted sense embedding learned from the training set. 
In Figure \ref{fig:similarities} we grouped these similarities by correct and incorrect predictions, revealing substantially different distributions. While incorrect prediction spans across the 0.5-0.9 interval, correct predictions are in the main higher than 0.75 for most words (over 97\% of all predictions using BERT-Large with similarity higher than 0.75 are correct, for example).  
Consequently, this analysis also shows that a simple threshold could be used for effectively discarding false matches, increasing the precision of 1NN methods.

\subsection{Role of training data}
\label{an:trainingdata}

In order to gain insights on the role of training data, we perform two types of analysis: (1) distribution of training data and, in particular, a comparison between skewed and balanced training sets (Section \ref{an:distribution}), and (2) the size of the training set (Section \ref{trainingsize}).

\subsubsection{Distribution}
\label{an:distribution}

To verify the impact of the distribution of the training data, we created a balanced training dataset for each word by randomly removing instances for the more frequent senses in order to have a balanced distribution over all senses. 
Note that the original CoarseWSD-20 dataset has a skewed sense distribution, given that it is constructed based on naturally-occurring texts.

\begin{table}[]
\setlength{\tabcolsep}{0.03in}
\resizebox{\textwidth}{!}{
\begin{tabular}{l@{\hspace{0.25in}}cccccc@{\hspace{0.25in}}cccccc}
\toprule
\multicolumn{1}{l}{} & \multicolumn{6}{c}{\textbf{Micro F1}}                         & \multicolumn{6}{c}{\textbf{Macro F1}}         \\
\cmidrule(lr@{0.25in}){2-7}
\cmidrule(lr){8-13}
\multicolumn{1}{l}{} & \multicolumn{2}{c}{\textbf{Static emb.}}                       & \multicolumn{2}{c}{\textbf{1NN}}                                & \multicolumn{2}{c}{\textbf{F-Tune}~~~~~~}                             & \multicolumn{2}{c}{\textbf{Static emb.}}                       & \multicolumn{2}{c}{\textbf{1NN}}                                & \multicolumn{2}{c}{\textbf{F-Tune}}                               \\
\cmidrule(lr){2-3}
\cmidrule(lr){4-5}
\cmidrule(lr@{0.25in}){6-7}
\cmidrule(lr){8-9}
\cmidrule(lr){10-11}
\cmidrule(lr){12-13}
\multicolumn{1}{l}{} & FTX-B & FTX-C & BRT-B & BRT-L & BRT-B & BRT-L & FTX-B & FTX-C & BRT-B & BRT-L & BRT-B & BRT-L \\
\midrule
\textbf{crane}       & \cellcolor[HTML]{FCF2F1}-3.8  & \cellcolor[HTML]{FDFEFE}~0.0   & \cellcolor[HTML]{FAFDFC}~0.6  & \cellcolor[HTML]{FDFEFE}~0.0  & \cellcolor[HTML]{FDFEFE}~0.0  & \cellcolor[HTML]{FDFEFE}~0.0  & \cellcolor[HTML]{FCF2F1}-3.7  & \cellcolor[HTML]{FDFEFE}~0.0   & \cellcolor[HTML]{FAFDFC}~0.6  & \cellcolor[HTML]{FDFEFE}~0.0  & \cellcolor[HTML]{FDFEFE}~0.0   & \cellcolor[HTML]{FDFEFE}~0.0   \\
\textbf{java}        & \cellcolor[HTML]{FDFFFE}-0.1  & \cellcolor[HTML]{FDFEFD}~0.1   & \cellcolor[HTML]{FDFEFE}~0.0  & \cellcolor[HTML]{FDFEFE}~0.0  & \cellcolor[HTML]{FDFEFE}~0.0  & \cellcolor[HTML]{FDFFFE}-0.1 & \cellcolor[HTML]{E88A82}-30.3 & \cellcolor[HTML]{F4C6C2}-15.1 & \cellcolor[HTML]{FDFEFD}~0.1  & \cellcolor[HTML]{FDFEFE}~0.0  & \cellcolor[HTML]{FDFEFE}~0.0   & \cellcolor[HTML]{FDFFFE}-0.1  \\
\textbf{apple}       & \cellcolor[HTML]{FEFFFE}-0.2  & \cellcolor[HTML]{FFFFFF}-0.6     & \cellcolor[HTML]{FDFEFE}~0.0  & \cellcolor[HTML]{FDFEFE}~0.0  & \cellcolor[HTML]{FDFEFE}~0.0  & \cellcolor[HTML]{FDFFFE}-0.1 & \cellcolor[HTML]{FBFEFD}~0.4   & \cellcolor[HTML]{FFFFFF}-0.4     & \cellcolor[HTML]{FDFEFE}~0.0  & \cellcolor[HTML]{FDFEFE}~0.0  & \cellcolor[HTML]{FDFEFE}~0.0   & \cellcolor[HTML]{FDFFFE}-0.1  \\
\textbf{mole}        & \cellcolor[HTML]{F7D5D2}-11.2 & \cellcolor[HTML]{FEFBFB}-1.5  & \cellcolor[HTML]{FDFEFE}~0.0  & \cellcolor[HTML]{FDFEFE}~0.0  & \cellcolor[HTML]{FEFEFE}-0.7 & \cellcolor[HTML]{FEFEFE}-0.7 & \cellcolor[HTML]{FEFDFD}-0.9  & \cellcolor[HTML]{F4FBF7}2.0   & \cellcolor[HTML]{FDFEFE}~0.0  & \cellcolor[HTML]{FDFEFE}~0.0  & \cellcolor[HTML]{FFFFFF}-0.5     & \cellcolor[HTML]{FEFEFE}-0.7  \\
\textbf{spring}      & \cellcolor[HTML]{FBEDEC}-5.0  & \cellcolor[HTML]{FDF9F9}-2.0  & \cellcolor[HTML]{FDFEFE}~0.0  & \cellcolor[HTML]{FCFEFD}~0.2  & \cellcolor[HTML]{FEFDFC}-1.1 & \cellcolor[HTML]{FEFDFD}-0.9 & \cellcolor[HTML]{F6D1CE}-12.3 & \cellcolor[HTML]{F6FCF9}1.5   & \cellcolor[HTML]{FEFFFE}-0.2 & \cellcolor[HTML]{FCFEFD}~0.1  & \cellcolor[HTML]{FEFDFD}-1.0  & \cellcolor[HTML]{FEFEFE}-0.7  \\
\textbf{chair}       & \cellcolor[HTML]{FAE9E7}-6.2  & \cellcolor[HTML]{FDF5F4}-3.1  & \cellcolor[HTML]{FDFEFE}~0.0  & \cellcolor[HTML]{FDFEFE}~0.0  & \cellcolor[HTML]{FEFDFD}-1.0 & \cellcolor[HTML]{FCFEFD}~0.3  & \cellcolor[HTML]{FCEFEE}-4.5  & \cellcolor[HTML]{FDF8F7}-2.3  & \cellcolor[HTML]{FDFEFE}~0.0  & \cellcolor[HTML]{FDFEFE}~0.0  & \cellcolor[HTML]{FEFCFC}-1.2  & \cellcolor[HTML]{FCFEFD}~0.3   \\
\textbf{hood}        & \cellcolor[HTML]{F9E4E2}-7.3  & \cellcolor[HTML]{FEFCFC}-1.2  & \cellcolor[HTML]{FDFEFE}~0.0  & \cellcolor[HTML]{FEFCFC}-1.2 & \cellcolor[HTML]{FFFFFF}-0.4    & \cellcolor[HTML]{FDFEFE}~0.0  & \cellcolor[HTML]{C6E8D7}12.2  & \cellcolor[HTML]{E9F6F0}4.4   & \cellcolor[HTML]{FEFEFE}-0.8 & \cellcolor[HTML]{FEFBFB}-1.5 & \cellcolor[HTML]{FEFDFD}-0.9  & \cellcolor[HTML]{FEFFFF}-0.3  \\
\textbf{seal}        & \cellcolor[HTML]{EEA6A0}-23.1 & \cellcolor[HTML]{FAE5E3}-7.2  & \cellcolor[HTML]{FCFEFD}~0.3  & \cellcolor[HTML]{FDFEFE}~0.0  & \cellcolor[HTML]{FDF6F5}-2.9 & \cellcolor[HTML]{FEFEFE}-0.7 & \cellcolor[HTML]{F8DEDB}-9.0  & \cellcolor[HTML]{F6D4D1}-11.5 & \cellcolor[HTML]{FCFEFD}~0.2  & \cellcolor[HTML]{FDFEFE}~0.0  & \cellcolor[HTML]{F9E4E2}-7.3  & \cellcolor[HTML]{FDF8F7}-2.4  \\
\textbf{bow}         & \cellcolor[HTML]{F8DCDA}-9.3  & \cellcolor[HTML]{FCF2F1}-3.7  & \cellcolor[HTML]{FDFEFE}~0.0  & \cellcolor[HTML]{FDFEFE}~0.0  & \cellcolor[HTML]{FEFBFB}-1.4 & \cellcolor[HTML]{FEFEFE}-0.8 & \cellcolor[HTML]{FDF8F7}-2.3  & \cellcolor[HTML]{FDF9F9}-2.0  & \cellcolor[HTML]{FDFEFE}~0.0  & \cellcolor[HTML]{FDFEFE}~0.0  & \cellcolor[HTML]{FEFAF9}-1.8  & \cellcolor[HTML]{FEFBFB}-1.5  \\
\textbf{club}        & \cellcolor[HTML]{F2BFBA}-16.8 & \cellcolor[HTML]{FBEAE8}-5.9  & \cellcolor[HTML]{FDFEFE}~0.0  & \cellcolor[HTML]{FEFBFB}-1.5 & \cellcolor[HTML]{FEFEFE}-0.8 & \cellcolor[HTML]{FDF5F5}-3.0 & \cellcolor[HTML]{F9DFDD}-8.6  & \cellcolor[HTML]{FEFEFE}-0.6  & \cellcolor[HTML]{FEFFFF}-0.3 & \cellcolor[HTML]{FEFBFB}-1.5 & \cellcolor[HTML]{FFFFFF}-0.4     & \cellcolor[HTML]{FDF7F7}-2.4  \\
\textbf{trunk}       & \cellcolor[HTML]{F5CECB}-13.0 & \cellcolor[HTML]{F8DDDB}-9.1  & \cellcolor[HTML]{FCF2F1}-3.9 & \cellcolor[HTML]{FDFEFE}~0.0  & \cellcolor[HTML]{FEFDFD}-0.9 & \cellcolor[HTML]{FEFAFA}-1.7 & \cellcolor[HTML]{FAE8E6}-6.4  & \cellcolor[HTML]{FCF0EF}-4.3  & \cellcolor[HTML]{FDF8F8}-2.1 & \cellcolor[HTML]{FDFEFE}~0.0  & \cellcolor[HTML]{FEFDFD}-0.9  & \cellcolor[HTML]{FEFAFA}-1.7  \\
\textbf{square}      & \cellcolor[HTML]{EDA49E}-23.7 & \cellcolor[HTML]{F9E1DF}-8.2  & \cellcolor[HTML]{FAE6E5}-6.8 & \cellcolor[HTML]{F9E3E1}-7.7 & \cellcolor[HTML]{FBEFED}-4.7 & \cellcolor[HTML]{FEFCFC}-1.3 & \cellcolor[HTML]{F7FCF9}1.4   & \cellcolor[HTML]{D1EDDF}9.6   & \cellcolor[HTML]{FCF4F3}-3.4 & \cellcolor[HTML]{FCF2F1}-3.9 & \cellcolor[HTML]{FBEEED}-4.8  & \cellcolor[HTML]{F8FDFA}1.1   \\
\textbf{arm}         & \cellcolor[HTML]{FDF7F7}-2.4  & \cellcolor[HTML]{FEFCFC}-1.2  & \cellcolor[HTML]{FDFEFE}~0.0  & \cellcolor[HTML]{FDFEFE}~0.0  & \cellcolor[HTML]{FDFEFE}~0.0  & \cellcolor[HTML]{FDFEFE}~0.0  & \cellcolor[HTML]{FAFDFC}~0.6   & \cellcolor[HTML]{FEFEFE}-0.8  & \cellcolor[HTML]{FDFEFE}~0.0  & \cellcolor[HTML]{FDFEFE}~0.0  & \cellcolor[HTML]{FDFEFE}~0.0   & \cellcolor[HTML]{FDFEFE}~0.0   \\
\textbf{digit}       & \cellcolor[HTML]{F2C0BB}-16.7 & \cellcolor[HTML]{FAE5E3}-7.1  & \cellcolor[HTML]{FDFEFE}~0.0  & \cellcolor[HTML]{FDFEFE}~0.0  & \cellcolor[HTML]{F9FDFB}~0.8  & \cellcolor[HTML]{FDFEFE}~0.0  & \cellcolor[HTML]{F6FCF9}1.5   & \cellcolor[HTML]{FCEFEE}-4.5  & \cellcolor[HTML]{FDFEFE}~0.0  & \cellcolor[HTML]{FDFEFE}~0.0  & \cellcolor[HTML]{F7FCFA}1.2   & \cellcolor[HTML]{FDFEFE}~0.0   \\
\textbf{bass}        & \cellcolor[HTML]{F8DDDB}-9.1  & \cellcolor[HTML]{F9E1DF}-8.2  & \cellcolor[HTML]{FBFEFD}~0.4  & \cellcolor[HTML]{F9FDFB}~0.8  & \cellcolor[HTML]{FBEDEC}-5.1 & \cellcolor[HTML]{FCF0EF}-4.4 & \cellcolor[HTML]{DEF2E8}6.8   & \cellcolor[HTML]{DFF3E9}6.5   & \cellcolor[HTML]{FBFDFC}~0.5  & \cellcolor[HTML]{F9FDFB}~0.9  & \cellcolor[HTML]{FBEBEA}-5.6  & \cellcolor[HTML]{FCF1F0}-4.0  \\
\textbf{yard}        & \cellcolor[HTML]{F6D0CD}-12.5 & \cellcolor[HTML]{FBEBEA}-5.6  & \cellcolor[HTML]{FDF6F5}-2.8 & \cellcolor[HTML]{FCF1F0}-4.2 & \cellcolor[HTML]{FAE9E8}-6.0 & \cellcolor[HTML]{FDF8F7}-2.3 & \cellcolor[HTML]{ABDDC5}18.2  & \cellcolor[HTML]{C8E9D9}11.6  & \cellcolor[HTML]{FEFAFA}-1.6 & \cellcolor[HTML]{FDF7F7}-2.5 & \cellcolor[HTML]{F8DEDC}-8.9  & \cellcolor[HTML]{FCF2F1}-3.9  \\
\textbf{pound}       & \cellcolor[HTML]{E67C73}-34.0 & \cellcolor[HTML]{ECA099}-24.7 & \cellcolor[HTML]{FDFEFE}~0.0  & \cellcolor[HTML]{FEFDFD}-1.0 & \cellcolor[HTML]{F8DEDC}-8.9 & \cellcolor[HTML]{FEFBFB}-1.4 & \cellcolor[HTML]{AADDC4}18.5  & \cellcolor[HTML]{57BB8A}36.7  & \cellcolor[HTML]{DBF1E6}7.5  & \cellcolor[HTML]{FFFFFF}-0.6    & \cellcolor[HTML]{F8DEDC}-8.8  & \cellcolor[HTML]{F4FBF7}2.0   \\
\textbf{deck}        & \cellcolor[HTML]{EB9A93}-26.3 & \cellcolor[HTML]{F8DDDB}-9.1  & \cellcolor[HTML]{FDF9F9}-2.0 & \cellcolor[HTML]{FEFDFD}-1.0 & \cellcolor[HTML]{FBEAE9}-5.7 & \cellcolor[HTML]{FCF2F1}-3.7 & \cellcolor[HTML]{C6E8D7}12.3  & \cellcolor[HTML]{7ECBA6}28.1  & \cellcolor[HTML]{FEFDFC}-1.1 & \cellcolor[HTML]{FFFFFF}-0.5    & \cellcolor[HTML]{FBEDEC}-5.0  & \cellcolor[HTML]{F4FBF7}2.1   \\
\textbf{bank}        & \cellcolor[HTML]{F2BDB8}-17.4 & \cellcolor[HTML]{F7D8D6}-10.3 & \cellcolor[HTML]{FCFEFD}~0.2  & \cellcolor[HTML]{FDFEFE}~0.0  & \cellcolor[HTML]{FDF6F6}-2.6 & \cellcolor[HTML]{FEF9F9}-1.9 & \cellcolor[HTML]{CEECDD}10.3  & \cellcolor[HTML]{D1EDDF}9.7   & \cellcolor[HTML]{F3FAF7}2.3  & \cellcolor[HTML]{FDFEFE}~0.0  & \cellcolor[HTML]{F7D7D5}-10.6 & \cellcolor[HTML]{FAE8E6}-6.5  \\
\textbf{pitcher}     & \cellcolor[HTML]{F5CECB}-13.0 & \cellcolor[HTML]{FAE8E6}-6.4  & \cellcolor[HTML]{FDFFFE}-0.1 & \cellcolor[HTML]{FDFEFE}~0.0  & \cellcolor[HTML]{FEFCFC}-1.3 & \cellcolor[HTML]{FFFFFF}-0.4    & \cellcolor[HTML]{B1E0C9}16.8  & \cellcolor[HTML]{98D6B7}22.4  & \cellcolor[HTML]{FDFEFE}~0.0  & \cellcolor[HTML]{FDFEFE}~0.0  & \cellcolor[HTML]{EB9891}-26.7 & \cellcolor[HTML]{F5CFCC}-12.7 \\
\midrule
\textbf{AVG}         & -12.6 & -5.8 & -0.7 & -0.8 & -2.1 & -1.1 & 1.0 & 4.6 & 0.1 & -0.5 & -4.2 & -1.6 \\
\bottomrule
\end{tabular}}
\caption{Performance drop or increase when using a fully balanced training set instead of the original CoarseWSD-20 skewed training set.\label{tab:balanceall}}
\end{table}

Table \ref{tab:balanceall} shows the performance drop or increase when using a fully balanced training set instead of the original CoarseWSD-20 skewed training set (tested on the original skewed test set). 
Performance is generally similar across the two settings for the less entropic words (on top) that tend to have more uniform distributions. 
For the more entropic words (e.g., \textit{deck}, \textit{bank} or \textit{pitcher}), even though balancing the data inevitably reduces the overall number of training instances to a large extent, it can result in improved macro results for FastText, and even improved macro-recall results for fine-tuning, as we will see in Table \ref{tab:precisionrecall-balanced}. 

This can be attributed to the better encoding of the least frequent senses, which corroborates the findings of \citet{postma-etal-2016-always} for conventional supervised WSD models, such as IMS or, in this case, FastText. In contrast, the micro-averaged results clearly depend on accurately knowing the original distribution in both the supervised and fine-tuning settings, as was also discussed in previous works \cite{bennett-etal-2016-lexsemtm,pasini2018two}. 
Moreover, the feature extraction procedure (1NN in this case) is much more robust to training distribution changes. 
Indeed, being solely based on vector similarities, the 1NN strategy is not directly influenced by the number of occurrences of each sense in the CoarseWSD-20 training set.

\begin{table}[]
\setlength{\tabcolsep}{11.0pt}
\resizebox{\textwidth}{!}{
\begin{tabular}{lrrrrrrrrr }
\toprule
\multicolumn{1}{l}{} &
  \multicolumn{4}{c}{\textbf{F-Tune (BRT-L)}} &&
  \multicolumn{4}{c}{\textbf{1NN (BRT-L)}} \\
  \cmidrule(lr){2-5}
  \cmidrule(lr){7-10}
 &
  \multicolumn{2}{c}{\textbf{Precision}} &
  \multicolumn{2}{c}{\textbf{Recall}} &&
  \multicolumn{2}{c}{\textbf{Precision}} &
  \multicolumn{2}{c}{\textbf{Recall}} \\
  \cmidrule(lr){2-3}
  \cmidrule(lr){4-5}
  \cmidrule(lr){7-8}
  \cmidrule(lr){9-10}
 &
  \multicolumn{1}{c}{\textbf{MFS}} &
  \multicolumn{1}{c}{\textbf{LFS}} &
  \multicolumn{1}{c}{\textbf{MFS}} &
  \multicolumn{1}{c}{\textbf{LFS}} &&
  \textbf{MFS} &
  \textbf{LFS} &
  \textbf{MFS} &
  \textbf{LFS} \\
  \midrule
{crane} &
  0.4 &
  \cellcolor[HTML]{FEFDFD}-0.4 &
  \cellcolor[HTML]{FEFDFD}-0.4 &
  0.4 &&
  0.0 &
  0.0 &
  0.0 &
  0.0 \\
{java} &
  0.0 &
  \cellcolor[HTML]{FEFEFE}-0.3 &
  \cellcolor[HTML]{FEFEFE}-0.2 &
  0.0 &&
  0.0 &
  0.0 &
  0.0 &
  0.0 \\
{apple} &
  \cellcolor[HTML]{FEFEFE}-0.1 &
  \cellcolor[HTML]{FEFEFE}-0.1 &
  \cellcolor[HTML]{FEFEFE}-0.1 &
  \cellcolor[HTML]{FEFEFE}-0.2 &&
  0.0 &
  0.0 &
  0.0 &
  0.0 \\
{mole} &
  \cellcolor[HTML]{FEFCFC}-0.9 &
  \cellcolor[HTML]{FEFCFC}-0.8 &
  \cellcolor[HTML]{FEFCFC}-0.9 &
  \cellcolor[HTML]{FEFAFA}-1.5 &&
  0.0 &
  0.0 &
  0.0 &
  0.0 \\
{spring} &
  \cellcolor[HTML]{FEFDFD}-0.6 &
  \cellcolor[HTML]{FEFCFB}-1.0 &
  \cellcolor[HTML]{FEFBFB}-1.3 &
  \cellcolor[HTML]{FEFBFA}-1.4 &&
  0.0 &
  \cellcolor[HTML]{FDFFFE}0.6 &
  0.0 &
  0.0 \\
{chair} &
  0.0 &
  \cellcolor[HTML]{FDFEFE}0.9 &
  0.4 &
  0.0 &&
  0.0 &
  0.0 &
  0.0 &
  0.0 \\
{hood} &
  \cellcolor[HTML]{FDFFFE}0.7 &
  0.0 &
  0.0 &
  \cellcolor[HTML]{FDF7F7}-2.6 &&
  \cellcolor[HTML]{FDF9F8}-2.1 &
  0.0 &
  0.0 &
  0.0 \\
{seal} &
  \cellcolor[HTML]{FEFEFE}-0.3 &
  0.0 &
  \cellcolor[HTML]{FEFDFD}-0.5 &
  0.0 &&
  0.0 &
  0.0 &
  0.0 &
  0.0 \\
{bow} &
  \cellcolor[HTML]{FDFEFE}0.8 &
  \cellcolor[HTML]{FEFCFC}-1.0 &
  \cellcolor[HTML]{FEFDFD}-0.6 &
  \cellcolor[HTML]{FEFBFA}-1.4 &&
  0.0 &
  0.0 &
  0.0 &
  0.0 \\
{club} &
  \cellcolor[HTML]{FDF5F4}-3.4 &
  \cellcolor[HTML]{FEFAFA}-1.6 &
  \cellcolor[HTML]{FDF8F8}-2.2 &
  \cellcolor[HTML]{FBEBEA}-6.9 &&
  \cellcolor[HTML]{FCF4F3}-3.9 &
  0.0 &
  \cellcolor[HTML]{FDFEFD}0.9 &
  \cellcolor[HTML]{FCEFEE}-5.5 \\
{trunk} &
  \cellcolor[HTML]{FDFFFE}0.7 &
  \cellcolor[HTML]{FAE9E8}-7.4 &
  \cellcolor[HTML]{FDF4F4}-3.6 &
  0.0 &&
  0.0 &
  0.0 &
  0.0 &
  0.0 \\
{square} &
  \cellcolor[HTML]{EBF7F1}6.5 &
  \cellcolor[HTML]{FEFDFD}-0.5 &
  \cellcolor[HTML]{F9E4E2}-9.4 &
  0.0 &&
  \cellcolor[HTML]{FEFDFD}-0.4 &
  0.0 &
  \cellcolor[HTML]{F6D2CF}-15.5 &
  0.0 \\
{arm} &
  0.0 &
  0.0 &
  0.0 &
  0.0 &&
  0.0 &
  0.0 &
  0.0 &
  0.0 \\
{digit} &
  0.0 &
  0.0 &
  0.0 &
  0.0 &&
  0.0 &
  0.0 &
  0.0 &
  0.0 \\
{bass} &
  \cellcolor[HTML]{F7FCFA}2.5 &
  \cellcolor[HTML]{FEFEFD}-0.4 &
  \cellcolor[HTML]{FAE6E4}-8.6 &
  \cellcolor[HTML]{FEFCFC}-0.8 &&
  \cellcolor[HTML]{FEFDFC}-0.7 &
  \cellcolor[HTML]{FAFDFB}1.8 &
  \cellcolor[HTML]{FDFFFE}0.7 &
  \cellcolor[HTML]{FCFEFD}1.1 \\
{yard} &
  \cellcolor[HTML]{FEFFFE}0.5 &
  \cellcolor[HTML]{F7D7D4}-14.0 &
  \cellcolor[HTML]{FDF5F5}-3.3 &
  \cellcolor[HTML]{F6FCF9}3.0 &&
  0.0 &
  \cellcolor[HTML]{FAE8E6}-7.9 &
  \cellcolor[HTML]{FCF0F0}-4.9 &
  0.0 \\
{pound} &
  \cellcolor[HTML]{F3FAF6}4.0 &
  \cellcolor[HTML]{F2BCB7}-23.4 &
  \cellcolor[HTML]{FBEEED}-5.8 &
  \cellcolor[HTML]{8AD0AE}36.7 &&
  \cellcolor[HTML]{FDF8F7}-2.4 &
  0.0 &
  0.0 &
  \cellcolor[HTML]{FEFBFB}-1.1 \\
{deck} &
  \cellcolor[HTML]{F3FAF7}3.9 &
  \cellcolor[HTML]{F0B1AC}-27.1 &
  \cellcolor[HTML]{FAE8E6}-8.0 &
  \cellcolor[HTML]{57BB8A}52.4 &&
  \cellcolor[HTML]{FDF6F6}-2.9 &
  0.0 &
  0.0 &
  \cellcolor[HTML]{FEFBFB}-1.1 \\
{bank} &
  \cellcolor[HTML]{FDFFFE}0.8 &
  \cellcolor[HTML]{EDA29C}-32.5 &
  \cellcolor[HTML]{FDF7F6}-2.8 &
  \cellcolor[HTML]{CFECDE}15.2 &&
  0.0 &
  0.0 &
  0.0 &
  0.0 \\
{pitcher} &
  0.1 &
  \cellcolor[HTML]{E67C73}-46.0 &
  \cellcolor[HTML]{FEFDFD}-0.4 &
  \cellcolor[HTML]{DFF2E9}10.3 &&
  0.0 &
  0.0 &
  0.0 &
  0.0 \\
  \midrule
{AVG} &
  \multicolumn{1}{c}{0.8} &
  \multicolumn{1}{c}{-7.8} &
  \multicolumn{1}{c}{-2.4} &
  \multicolumn{1}{c}{5.2} &&
  -0.6 &
  -0.3 &
  -0.9 &
  -0.3 \\
 \bottomrule
\end{tabular}
}
\caption{Precision and recall drop or increase on the Most Frequent Sense (MFS) and Least Frequent Sense (LFS) classes when using a fully balanced training set. \label{tab:precisionrecall-balanced}}
\end{table}

To complement these results, Table \ref{tab:precisionrecall-balanced} shows the performance difference on the MFS (Most Frequent Class) and LFS (Least Frequent Class) classes when using the balanced training set. The most interesting takeaway from this experiment is the marked difference between precision and recall for the LFS in entropic words (bottom). While the recall of the BERT-Large fine-tuned model increases significantly (up to 52.4 points in the case of \textit{deck}), the precision decreases (e.g. -27.1 points for \textit{deck}). 
This means that the model is clearly less biased towards the MFS with a balanced training set, as we could expect. 
However, the precision for LFS is also lower, due to the model's lower sensitivity for higher-frequency senses. In general, these results suggest that the fine-tuned BERT model is overly sensitive to the distribution of the training data, while its feature extraction counterpart suffers considerably less from this issue.
In Section \ref{sec:bias} we will extend the analysis on the bias present in each of the models.

\begin{table}[]
\setlength{\tabcolsep}{6pt}
\renewcommand{\arraystretch}{1.2}
\resizebox{\textwidth}{!}{
\begin{tabular}{lccccccccccccc}
\toprule
\multicolumn{1}{l}{} &
  \multicolumn{6}{c}{\textbf{Fine-tuning (BRT-L)}} & &
  \multicolumn{6}{c}{\textbf{1NN (BRT-L)}} \\
\cmidrule(lr){2-7}
\cmidrule(lr){9-14}
\multicolumn{1}{l}{} &
  \textbf{1\%} &
  \textbf{5\%} &
  \textbf{10\%} &
  \textbf{25\%} &
  \textbf{50\%} &
  \textbf{ALL\%} &&
  \textbf{1\%} &
  \textbf{5\%} &
  \textbf{10\%} &
  \textbf{25\%} &
  \textbf{50\%} &
  \textbf{ALL\%} \\
\midrule
\textbf{Macro} &	74.2	&	81.6	&	85.8	&	91.5	&	94.2	&	95.1	&&	94.4	&	95.3	&	95.6	&	95.8	&	96.0	&	96.4	\\
\textbf{Micro} 	&	89.0	&	93.5	&	95.3	&	96.3	&	97.0	&	97.5	&&	95.5	&	95.8	&	95.7	&	95.7	&	95.6	&	95.8	\\
\midrule
\textbf{MFS} 	&	91.9	&	95.3	&	96.4	&	97.2	&	97.5	&	98.0	&&	95.8	&	95.8	&	95.6	&	95.6	&	95.4	&	95.4	\\
\textbf{LFS} 	&	52.1	&	64.3	&	71.9	&	83.4	&	88.5	&	91.0	&&	91.6	&	93.3	&	94.1	&	94.6	&	95.5	&	96.6	\\
\bottomrule
\end{tabular}
}
\caption{Macro- and micro-F1 \% performance for the two BERT-Large models. The last two rows indicate the F1 performance on the Most Frequent Sense (MFS) and Least Frequent Sense (LFS) classes.
\label{tab:sizesummary}}

\end{table}

\subsubsection{Size}
\label{trainingsize}

We performed an additional experiment to investigate the impact of training data size on the performance for the most and least frequent senses.
To this end, we shrank the training dataset for all words, while preserving their original distribution. 
Table \ref{tab:sizesummary} shows a summary of the aggregated micro-F1 and macro-F1 results, including the performance on the most and least frequent senses.\footnote{In the appendix we include detailed results for each word and their MFS and LFS performance.} Clearly, the 1NN model performs considerably better than fine-tuning in settings with low training data (e.g., 74.2\% to 94.4\% macro-F1 with 1\% of the training data). Interestingly, the 1NN's performance does not deteriorate with few training data, as the results with 1\% and 100\% of the training data do not vary much (less than two absolute points decrease in performance for micro-F1 and 0.3 in terms of micro-F1). Even for the LFS, the overall performance with 1\% of the training data is above 90 (i.e., 91.6).
This is an encouraging behaviour, as in real settings sense-annotated data is generally scarce.

\begin{table}[]
\setlength{\tabcolsep}{6.0pt}
\resizebox{\textwidth}{!}{
\begin{tabular}{lrrrrrrlrrrrrr}
\toprule
                 & \multicolumn{6}{c}{\textbf{Fine-tuning (BRT-L)}}            &  & \multicolumn{6}{c}{\textbf{1NN (BRT-L)}}      \\
\cmidrule(lr){2-7}
\cmidrule(lr){8-14}
                 & \multicolumn{1}{c}{\textbf{1\%}}      & \multicolumn{1}{c}{\textbf{5\%}} & \multicolumn{1}{c}{\textbf{10\%}} & \multicolumn{1}{c}{\textbf{25\%}} & \multicolumn{1}{c}{\textbf{50\%}} & \multicolumn{1}{c}{\textbf{ALL}} &  & \multicolumn{1}{c}{\textbf{1\%}}                   & \multicolumn{1}{c}{\textbf{5\%}}                   & \multicolumn{1}{c}{\textbf{10\%}} & \multicolumn{1}{c}{\textbf{25\%}} & \multicolumn{1}{c}{\textbf{50\%}} & \multicolumn{1}{c}{\textbf{ALL}} \\
\cmidrule(lr){2-7}
\cmidrule(lr){8-14}
crane &
  \cellcolor[HTML]{F8DBD9}83.3 &
  \cellcolor[HTML]{FEFCFC}95.7 &
  \cellcolor[HTML]{FEFCFC}95.7 &
  \cellcolor[HTML]{F1FAF5}96.8 &
  \cellcolor[HTML]{FEFCFC}95.5 &
  \cellcolor[HTML]{B3E1CA}98.1 &
   &
  \cellcolor[HTML]{FEFEFE}96.4 &
  \cellcolor[HTML]{FCFEFD}96.6 &
  \cellcolor[HTML]{F5FBF8}96.7 &
  \cellcolor[HTML]{F5FBF8}96.7 &
  \cellcolor[HTML]{F5FBF8}96.7 &
  \cellcolor[HTML]{F5FBF8}96.7 \\
java &
  \cellcolor[HTML]{86CFAB}99.0 &
  \cellcolor[HTML]{85CEAA}99.1 &
  \cellcolor[HTML]{6AC397}99.6 &
  \cellcolor[HTML]{70C59C}99.5 &
  \cellcolor[HTML]{6CC499}99.6 &
  \cellcolor[HTML]{67C296}99.7 &
   &
  \cellcolor[HTML]{69C397}99.6 &
  \cellcolor[HTML]{69C397}99.6 &
  \cellcolor[HTML]{69C397}99.6 &
  \cellcolor[HTML]{69C397}99.6 &
  \cellcolor[HTML]{69C397}99.6 &
  \cellcolor[HTML]{69C397}99.6 \\
apple &
  \cellcolor[HTML]{78C9A1}99.3 &
  \cellcolor[HTML]{77C8A0}99.4 &
  \cellcolor[HTML]{73C79E}99.4 &
  \cellcolor[HTML]{73C79E}99.4 &
  \cellcolor[HTML]{70C59C}99.5 &
  \cellcolor[HTML]{6BC398}99.6 &
   &
  \cellcolor[HTML]{81CCA7}99.1 &
  \cellcolor[HTML]{81CCA7}99.1 &
  \cellcolor[HTML]{81CCA7}99.1 &
  \cellcolor[HTML]{81CCA7}99.1 &
  \cellcolor[HTML]{81CCA7}99.1 &
  \cellcolor[HTML]{81CCA7}99.1 \\
mole &
  \cellcolor[HTML]{F6D2CF}79.8 &
  \cellcolor[HTML]{FEFAFA}94.8 &
  \cellcolor[HTML]{CAEADA}97.6 &
  \cellcolor[HTML]{7AC9A3}99.3 &
  \cellcolor[HTML]{7AC9A2}99.3 &
  \cellcolor[HTML]{80CCA7}99.2 &
   &
  \cellcolor[HTML]{9BD7B9}98.6 &
  \cellcolor[HTML]{84CDA9}99.1 &
  \cellcolor[HTML]{86CEAB}99.0 &
  \cellcolor[HTML]{86CEAB}99.0 &
  \cellcolor[HTML]{86CEAB}99.0 &
  \cellcolor[HTML]{86CEAB}99.0 \\
spring &
  \cellcolor[HTML]{FEFAFA}94.8 &
  \cellcolor[HTML]{CBEADB}97.6 &
  \cellcolor[HTML]{F0F9F5}96.8 &
  \cellcolor[HTML]{EBF7F1}96.9 &
  \cellcolor[HTML]{C2E7D5}97.8 &
  \cellcolor[HTML]{B3E1CA}98.1 &
   &
  \cellcolor[HTML]{BDE5D1}97.9 &
  \cellcolor[HTML]{BCE4D1}97.9 &
  \cellcolor[HTML]{BCE4D1}97.9 &
  \cellcolor[HTML]{BDE5D1}97.9 &
  \cellcolor[HTML]{BEE5D2}97.9 &
  \cellcolor[HTML]{C2E6D4}97.8 \\
chair &
  \cellcolor[HTML]{F4C9C5}76.2 &
  \cellcolor[HTML]{FCF3F2}92.2 &
  \cellcolor[HTML]{FEFBFB}95.2 &
  \cellcolor[HTML]{FEFDFD}96.1 &
  \cellcolor[HTML]{FEFEFE}96.4 &
  \cellcolor[HTML]{FEFCFC}95.5 &
   &
  \cellcolor[HTML]{FDF9F8}94.3 &
  \cellcolor[HTML]{FEF9F9}94.6 &
  \cellcolor[HTML]{FEFAF9}94.7 &
  \cellcolor[HTML]{FEFAF9}94.7 &
  \cellcolor[HTML]{FEFAF9}94.7 &
  \cellcolor[HTML]{FEFAF9}94.7 \\
hood &
  \cellcolor[HTML]{EB968F}57.2 &
  \cellcolor[HTML]{FBEBEA}89.3 &
  \cellcolor[HTML]{FCF3F3}92.3 &
  \cellcolor[HTML]{FCFEFD}96.6 &
  \cellcolor[HTML]{C5E8D7}97.7 &
  \cellcolor[HTML]{6AC397}99.6 &
   &
  \cellcolor[HTML]{FEFAF9}94.7 &
  \cellcolor[HTML]{9CD7BA}98.6 &
  \cellcolor[HTML]{7CCAA4}99.2 &
  \cellcolor[HTML]{70C59B}99.5 &
  \cellcolor[HTML]{57BB8A}100 &
  \cellcolor[HTML]{57BB8A}100 \\
seal &
  \cellcolor[HTML]{F6D3D0}80.3 &
  \cellcolor[HTML]{FEFDFC}95.8 &
  \cellcolor[HTML]{FEFFFF}96.5 &
  \cellcolor[HTML]{AEDEC7}98.2 &
  \cellcolor[HTML]{B8E3CE}98.0 &
  \cellcolor[HTML]{99D6B8}98.6 &
   &
  \cellcolor[HTML]{9DD8BB}98.6 &
  \cellcolor[HTML]{9AD6B9}98.6 &
  \cellcolor[HTML]{97D5B6}98.7 &
  \cellcolor[HTML]{9AD7B9}98.6 &
  \cellcolor[HTML]{99D6B8}98.6 &
  \cellcolor[HTML]{9ED8BC}98.5 \\
bow &
  \cellcolor[HTML]{E78178}49.3 &
  \cellcolor[HTML]{FAE5E3}86.8 &
  \cellcolor[HTML]{FEFCFC}95.7 &
  \cellcolor[HTML]{FEFDFD}96.0 &
  \cellcolor[HTML]{CFECDE}97.5 &
  \cellcolor[HTML]{9BD7BA}98.6 &
   &
  \cellcolor[HTML]{FDF7F6}93.5 &
  \cellcolor[HTML]{FEFDFD}96.0 &
  \cellcolor[HTML]{FEFEFE}96.2 &
  \cellcolor[HTML]{FEFDFD}95.9 &
  \cellcolor[HTML]{FEFCFC}95.7 &
  \cellcolor[HTML]{FEFCFC}95.7 \\
club &
  \cellcolor[HTML]{F1B8B3}70.1 &
  \cellcolor[HTML]{F5CCC8}77.4 &
  \cellcolor[HTML]{F5CBC7}77.0 &
  \cellcolor[HTML]{F6D3D0}80.0 &
  \cellcolor[HTML]{F8DBD8}83.0 &
  \cellcolor[HTML]{F8DEDB}84.1 &
   &
  \cellcolor[HTML]{F9E1DF}85.6 &
  \cellcolor[HTML]{F9E4E2}86.5 &
  \cellcolor[HTML]{FAE6E4}87.4 &
  \cellcolor[HTML]{FAE7E5}87.6 &
  \cellcolor[HTML]{FAE8E6}88.0 &
  \cellcolor[HTML]{FBEAE8}88.7 \\
trunk &
  \cellcolor[HTML]{F5CDCA}77.9 &
  \cellcolor[HTML]{F8DFDD}84.6 &
  \cellcolor[HTML]{D0ECDF}97.5 &
  \cellcolor[HTML]{99D6B8}98.6 &
  \cellcolor[HTML]{99D6B8}98.6 &
  \cellcolor[HTML]{B7E2CD}98.0 &
   &
  \cellcolor[HTML]{C6E8D7}97.7 &
  \cellcolor[HTML]{A7DCC2}98.3 &
  \cellcolor[HTML]{96D5B6}98.7 &
  \cellcolor[HTML]{7AC9A2}99.3 &
  \cellcolor[HTML]{7AC9A2}99.3 &
  \cellcolor[HTML]{7AC9A2}99.3 \\
square &
  \cellcolor[HTML]{F0B4AF}68.4 &
  \cellcolor[HTML]{F1B7B2}69.6 &
  \cellcolor[HTML]{F3C1BD}73.5 &
  \cellcolor[HTML]{F4CAC6}76.6 &
  \cellcolor[HTML]{F6D1CE}79.4 &
  \cellcolor[HTML]{FCF1F0}91.4 &
   &
  \cellcolor[HTML]{FAE5E3}86.7 &
  \cellcolor[HTML]{FAE8E6}88.0 &
  \cellcolor[HTML]{FAE7E6}87.8 &
  \cellcolor[HTML]{FAE8E6}88.1 &
  \cellcolor[HTML]{FCF0EF}91.1 &
  \cellcolor[HTML]{FEFAF9}94.7 \\
arm &
  \cellcolor[HTML]{FBEDEC}90.1 &
  \cellcolor[HTML]{B3E0CA}98.1 &
  \cellcolor[HTML]{7DCBA5}99.2 &
  \cellcolor[HTML]{7DCBA5}99.2 &
  \cellcolor[HTML]{7DCBA5}99.2 &
  \cellcolor[HTML]{7DCBA5}99.2 &
   &
  \cellcolor[HTML]{6BC498}99.6 &
  \cellcolor[HTML]{6BC498}99.6 &
  \cellcolor[HTML]{6BC498}99.6 &
  \cellcolor[HTML]{6BC498}99.6 &
  \cellcolor[HTML]{6BC498}99.6 &
  \cellcolor[HTML]{6BC498}99.6 \\
digit &
  \cellcolor[HTML]{FCF3F3}92.4 &
  \cellcolor[HTML]{F6D2CF}79.7 &
  \cellcolor[HTML]{FCF3F2}92.1 &
  \cellcolor[HTML]{92D3B4}98.8 &
  \cellcolor[HTML]{57BB8A}100 &
  \cellcolor[HTML]{57BB8A}100 &
   &
  \cellcolor[HTML]{84CDA9}99.1 &
  \cellcolor[HTML]{57BB8A}100 &
  \cellcolor[HTML]{57BB8A}100 &
  \cellcolor[HTML]{57BB8A}100 &
  \cellcolor[HTML]{57BB8A}100 &
  \cellcolor[HTML]{57BB8A}100 \\
bass &
  \cellcolor[HTML]{F2BEB9}72.2 &
  \cellcolor[HTML]{F6D1CE}79.4 &
  \cellcolor[HTML]{F8DEDC}84.3 &
  \cellcolor[HTML]{FAE4E3}86.7 &
  \cellcolor[HTML]{FAE7E6}87.8 &
  \cellcolor[HTML]{FAE7E5}87.6 &
   &
  \cellcolor[HTML]{F8DBD8}83.1 &
  \cellcolor[HTML]{F8DDDB}83.8 &
  \cellcolor[HTML]{F8DEDC}84.4 &
  \cellcolor[HTML]{F9DFDD}84.8 &
  \cellcolor[HTML]{F9DFDD}84.8 &
  \cellcolor[HTML]{F8DDDB}84.0 \\
yard &
  \cellcolor[HTML]{F7DAD7}82.7 &
  \cellcolor[HTML]{F9E2E0}85.7 &
  \cellcolor[HTML]{FAE9E7}88.3 &
  \cellcolor[HTML]{FDF9F8}94.3 &
  \cellcolor[HTML]{84CEAA}99.1 &
  \cellcolor[HTML]{84CEAA}99.1 &
   &
  \cellcolor[HTML]{FDF6F6}93.4 &
  \cellcolor[HTML]{FDF6F6}93.4 &
  \cellcolor[HTML]{FDF5F4}92.8 &
  \cellcolor[HTML]{FDF4F3}92.6 &
  \cellcolor[HTML]{FCF3F2}92.2 &
  \cellcolor[HTML]{FDF6F6}93.4 \\
pound &
  \cellcolor[HTML]{E98C84}53.5 &
  \cellcolor[HTML]{E7847B}50.4 &
  \cellcolor[HTML]{E67C73}47.3 &
  \cellcolor[HTML]{E88A82}52.6 &
  \cellcolor[HTML]{F8DBD9}83.2 &
  \cellcolor[HTML]{F8DDDB}83.9 &
   &
  \cellcolor[HTML]{FAE5E4}87.0 &
  \cellcolor[HTML]{FCF4F3}92.4 &
  \cellcolor[HTML]{FDF6F6}93.3 &
  \cellcolor[HTML]{FDF6F5}93.2 &
  \cellcolor[HTML]{FDF9F8}94.3 &
  \cellcolor[HTML]{FDF9F8}94.3 \\
deck &
  \cellcolor[HTML]{EA948D}56.7 &
  \cellcolor[HTML]{E67E75}48.2 &
  \cellcolor[HTML]{E67E75}48.2 &
  \cellcolor[HTML]{F1B9B4}70.2 &
  \cellcolor[HTML]{F5CBC8}77.2 &
  \cellcolor[HTML]{F5CDCA}78.0 &
   &
  \cellcolor[HTML]{F9E1DF}85.5 &
  \cellcolor[HTML]{F9E0DE}85.1 &
  \cellcolor[HTML]{FBEAE9}88.9 &
  \cellcolor[HTML]{FCF0EF}91.1 &
  \cellcolor[HTML]{FCF3F2}92.1 &
  \cellcolor[HTML]{FEFCFC}95.7 \\
bank &
  \cellcolor[HTML]{E7837B}50.2 &
  \cellcolor[HTML]{EA928B}55.9 &
  \cellcolor[HTML]{F4C5C1}74.9 &
  \cellcolor[HTML]{E4F5ED}97.1 &
  \cellcolor[HTML]{FEFCFC}95.7 &
  \cellcolor[HTML]{FEFCFC}95.6 &
   &
  \cellcolor[HTML]{E9F6F0}97.0 &
  \cellcolor[HTML]{99D6B8}98.6 &
  \cellcolor[HTML]{8ED2B0}98.9 &
  \cellcolor[HTML]{A0D9BD}98.5 &
  \cellcolor[HTML]{C4E8D6}97.7 &
  \cellcolor[HTML]{C4E8D6}97.7 \\
pitcher &
  \cellcolor[HTML]{E7827A}49.9 &
  \cellcolor[HTML]{E88981}52.3 &
  \cellcolor[HTML]{EEA8A2}63.9 &
  \cellcolor[HTML]{FEFEFE}96.5 &
  \cellcolor[HTML]{78C9A1}99.3 &
  \cellcolor[HTML]{DAF0E6}97.3 &
   &
  \cellcolor[HTML]{59BC8B}100 &
  \cellcolor[HTML]{59BC8B}100 &
  \cellcolor[HTML]{59BC8B}100 &
  \cellcolor[HTML]{59BC8B}100 &
  \cellcolor[HTML]{59BC8B}100 &
  \cellcolor[HTML]{59BC8C}100 \\
\midrule
\it Average
&
  \cellcolor[HTML]{F3C3BF}74.2 &
  \cellcolor[HTML]{F7D7D4}81.6 &
  \cellcolor[HTML]{F9E2E0}85.8 &
  \cellcolor[HTML]{FCF1F0}91.5 &
  \cellcolor[HTML]{FDF8F8}94.2 &
  \cellcolor[HTML]{FEFBFA}95.1 &
   &
  \multicolumn{1}{c}{\cellcolor[HTML]{FDF9F8}94.4} &
  \multicolumn{1}{c}{\cellcolor[HTML]{FEFBFB}95.3} &
  \cellcolor[HTML]{FEFCFC}95.6 &
  \cellcolor[HTML]{FEFDFC}95.8 &
  \cellcolor[HTML]{FEFDFD}96.0 &
  \cellcolor[HTML]{FEFEFE}96.4 \\
  \bottomrule
\end{tabular}
}
\caption{Macro-F1 results on the CoarseWSD-20 test set using training sets of different sizes sampled from the original training set. \label{table:macro_results}}
\end{table}

To get a more detailed picture for each word, Table \ref{table:macro_results} shows the macro-F1 results for each word and training size.\footnote{In the appendix we include the same table for the micro-F1 results.} Again, we can observe a large drop for the most entropic words in the fine-tuning setting. Examples of words with a considerable degrading performance are \textit{pitcher} or \textit{bank}, which decrease from macro-F1 scores higher than 95\% in both cases (97.3 and 95.6, respectively) to as low as 49.9 and 50.2 (almost random chance) with 1\% of the training data, and still lower than 75\% with 10\% of the training data (63.9 and 74.9, respectively). This trend clearly highlights the need for gathering reasonable amounts of training data for the obscure senses. Moreover, this establishes a trade-off between balancing or preserving the original skewed distribution depending on the end goal, as discussed in Section \ref{an:distribution}.

\begin{figure}[h]
\centering
\includegraphics[width=1.0\linewidth]{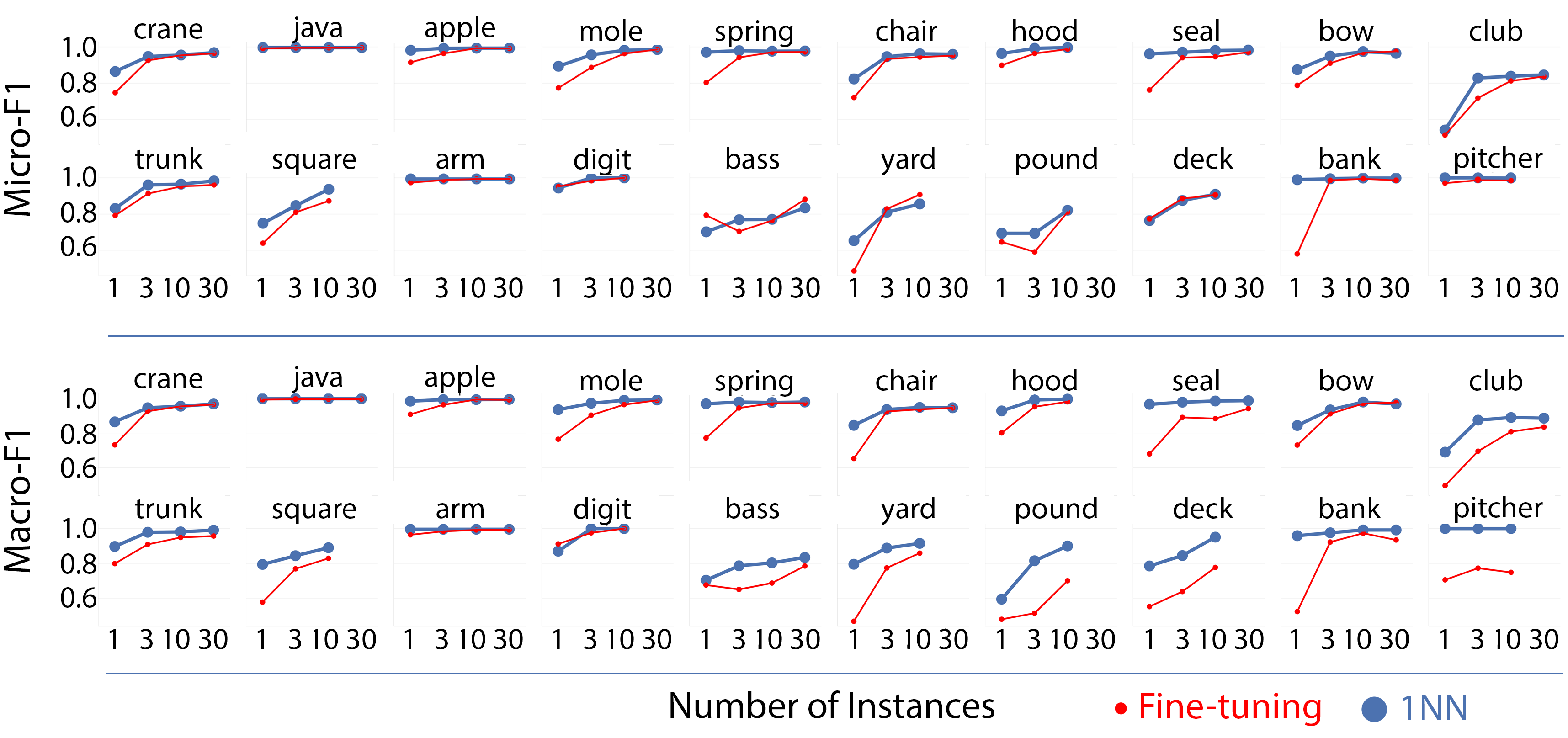}  

\caption{Micro and macro F-scores for different values of $n$ in the $n$-shot setting, for all the words and for the two WSD strategies. Results are averaged from three runs over three different samples.
\label{fig:line-plots}}
\end{figure}

\subsection{\textit{n}-shot learning}
\label{an:nshot}

\begin{figure}
\centering
\includegraphics[width=0.8\linewidth]{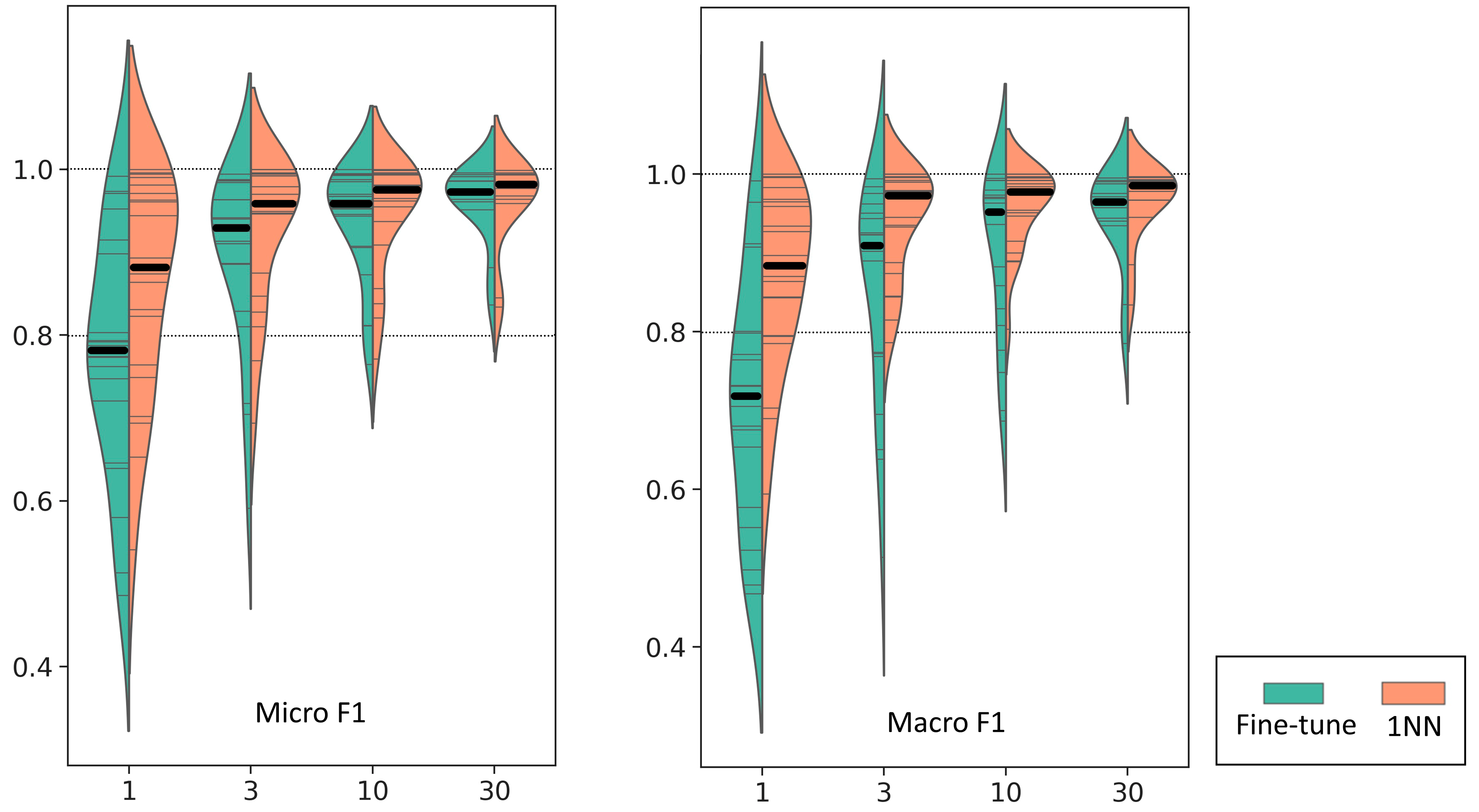}  

\caption{Distribution of performance scores for all 20 words according to micro and macro F1 in the two WSD strategies (left: fine-tuning, right: 1NN) and for different values of $n$, i.e., 1, 3, 10, 30 (if available).
\label{fig:nshot}}
\end{figure}

Given the results of the previous section, one may wonder how many instances would be enough for BERT to perform well in coarse-grained WSD. 
To verify this, we fine-tuned BERT on limited amounts of training data, with uniform distribution over word senses, each having between 1 (i.e., one-shot) and 30 instances. 
Figure \ref{fig:line-plots} shows the performance of both 1NN and Fine-Tuning strategies on this set of experiments.
Perhaps surprisingly, we can see how having only three instances per sense is enough for achieving a competitive result. 
Then, only small improvements can be obtained by adding more instances. 
This is relevant in the context of WSD, as generally current sense-annotated corpora follow the Zipf's law \cite{Zipf:49}, and therefore contain many repeated senses that are very frequent. Significant improvements may therefore be obtained by simply getting a few sense annotations for less frequent instances.
Figure \ref{fig:nshot} summarizes Figure \ref{fig:line-plots} by showing the distribution of words according to their performance in the two strategies.
In the case of Fine-Tuning, the performance is generally better in terms of micro compared to macro F-score.
This further corroborates the previous observation, that there is a bias towards the most frequent sense (cf. Section \ref{an:distribution}). 
Additionally, in contrast to 1NN, Fine-tuning greatly benefits from the increase in the training-data size, which also indicates the more robust behaviour of 1NN strategy compared to its counterpart (cf. Section \ref{an:distribution}).

\definecolor{c1}{rgb}{0.55,0.8,0.65}
\definecolor{c2}{rgb}{0.90,0.58,0.55}

\subsection{Bias analysis}
\label{sec:bias}

Supervised classifiers are known to have label bias towards more frequent classes, i.e., those that are seen more frequently in the training data \cite{NIPS2016_6374}, and this is particularly noticeable in WSD \cite{postma-etal-2016-always,blevins-zettlemoyer-2020-moving}.
Label bias is a reasonable choice for maximizing performance when the distribution of classes is skewed, particularly for classification tasks with a small number of categories (which is often the case in WSD).
For the same reason, many of the knowledge-based systems are coupled with the MFS back-off strategy: when the system is not confident in its disambiguation, it backs off to the most frequent sense (MFS) of the word (instead of resorting to the low-confidence decision).

We were interested in investigating the inherent sense biases in the two BERT-based WSD strategies.
We opted for the $n$-shot setting given that it provides a suitable setting for evaluating the relationship between sense bias and training data size.
Moreover, given that the training data in the $n$-shot setting is uniformly distributed (balanced), the impact of sense-annotated training data in introducing sense bias is minimized.
This analysis is mainly focused on two questions: (1) how do the two strategies (fine-tuning and 1NN) compare in terms of sense bias?, and (2) what are the inherent sense biases (if any) in the pre-trained BERT language model?

\subsubsection{Sense bias definition}

We propose the following procedure for computing the disambiguation bias towards a specific sense.\footnote{The procedure can presumably be used for quantifying bias in other similar classification settings.}
For a word with polysemy $n$, we are interested in computing the disambiguation bias $B_j$ towards its $j^{th}$ sense ($s_j$).
Let $n_{ij}$ be the total number of test instances with the gold label $s_i$ that were mistakenly disambiguated as $s_j$ ($i \ne j$). 
We first normalize $n_{ij}$ by the total number of (gold-labeled) instances for $s_i$, i.e., $\Sigma_j{n_{ij}}$, to obtain bias $b_{ij}$, which is the bias from sense $i$ to sense $j$.
In other words, $b_{ij}$ denotes the ratio of $s_i$-labeled instances which were misclassified as $s_j$.
The total bias towards a specific sense, $B_j$, is then computed as:
\begin{equation}
B_j=\sum_{\substack{i=1\\ i \ne j}}^{n} {(\frac{n_{ij}}{\Sigma_j{n_{ij}}})}
\end{equation}
The value of $B_j$ denotes the tendency of the disambiguation system to disambiguate a word with the intended sense of $s_k$, $k \ne j$ incorrectly as $s_j$.
The higher the value of $B_j$, the more the disambiguation model is biased towards $s_j$.
We finally compute the \textbf{sense bias $B$} as the \textit{maximum} $B_j$ value towards different senses of a specific word, i.e., $max {(B_j)}, j \in [1,n]$.
Given fluctuations in the results, particularly for the case of small training data, we take the median of three runs to compute $B_{j}$.

In our coarse-grained disambiguation setting, the bias $B$ can be mostly attributed to the case where the system did not have enough evidence to distinguish $s_j$ from other senses and had pre-training bias towards $s_j$.
One intuitive explanation for this would be that the language model is biased towards $s_j$ because it has seen the target word more often with this intended sense than other $s_{k,j \ne k}$ senses.

\subsubsection{Results}

\begin{figure}
\includegraphics[width=1.0\linewidth]{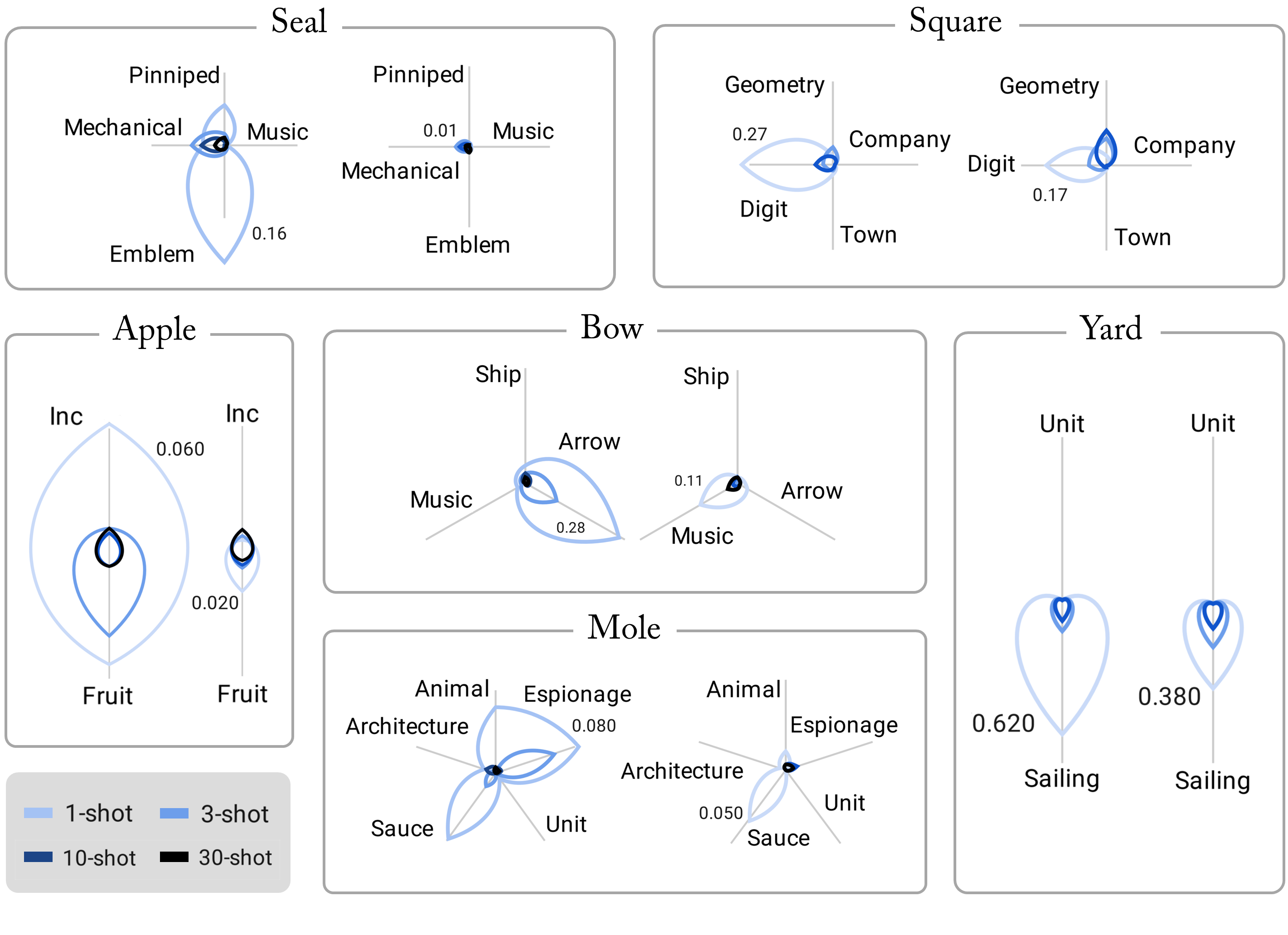}  

\caption{Sense bias for a few representative cases from each polysemy class for the two WSD strategies (left: fine-tuning, right: 1NN) and for different values of $n$, i.e., 1, 3, 10, 30 (if available).
\label{fig:bias}}
\end{figure}

Table \ref{tab:bias} reports the average sense bias values ($B$) for the two WSD strategies and for different values of $n$ (training data size) in the $n$-shot setting.
We also illustrate using radar charts in Figure \ref{fig:bias} the sense bias for a few representative cases.
The numbers reported in the figure (in parentheses) represent the bias value $B$ for the corresponding setting (word, WSD strategy, and $n$'s value).

\begin{table}[]
\setlength{\tabcolsep}{9.5pt}
\renewcommand{\arraystretch}{1.2}
\resizebox{\textwidth}{!}{
\begin{tabular}{ccccccccccc}
\toprule
  \multicolumn{2}{c}{\textbf{One-shot}} &&
  \multicolumn{2}{c}{\textbf{3-shot}} &&
  \multicolumn{2}{c}{\textbf{10-shot}} &&
  \multicolumn{2}{c}{\textbf{30-shot}} \\
\cmidrule(lr){1-2}
\cmidrule(lr){4-5}
\cmidrule(lr){7-8}
\cmidrule(lr){10-11}

F-Tune & 1NN &&
F-Tune & 1NN &&
F-Tune & 1NN &&
F-Tune & 1NN \\

\midrule

0.232   &
0.137	&&
0.111	&
0.078	&&
0.050	&
0.052	&&
0.021	&
0.025 \\

\bottomrule
\end{tabular}
}
\caption{Average sense bias values ($B$) for the two WSD strategies and for different values of $n$.\label{tab:bias}}

\end{table}

Based on our observations, we draw the following general conclusions.

\paragraph{Bias and training size}
There is a consistent pattern across all words and for both the strategies: sense bias rapidly reduces with increase in the training data.
Specifically, the average bias $B$ approximately reduces by half with each step of increase in the training size.
This is supported by the radar charts in Figure \ref{fig:bias} (see, for instance, \textit{apple}, \textit{yard}, and \textit{bow}).
The WSD system tends to be heavily biased in the one-shot setting (particularly in the fine-tuning setting), but the bias often improves significantly with just 3 instances in the training data (3-shot).

\paragraph{Disambiguation strategy: 1NN vs. Fine-tuning}
Among the two WSD strategies, the 1NN approach proves to be more robust with respect to sense biases.
This is particularly highlighted in the one-shot setting where the average sense bias value is 0.137 for 1NN in comparison to 0.232 for fine-tuning.
The trend is also clearly visible for almost all words in the radar charts in Figure \ref{fig:bias}.
This corroborates our findings in Section \ref{an:nshot} that the 1NN strategy is the preferable choice particularly with limited data.
For higher values of $n$ (larger training sizes) the difference between the two strategies diminishes, with both settings proving robust with respect to sense bias.

It is also notable that the two strategies, despite being usually similar in behaviour, might not necessarily have matching biases towards the same senses.
For instance, the fine-tuning setting shows bias only towards the arrow sense of \textit{bow}, whereas 1NN is instead (slightly) biased towards its music sense.
Another example is for the word \textit{digit} for which with the same set of training instances in the one-shot setting (one sentence for each of the two senses), all the mistakes (5 in total) of the fine-tuning model are numerical digit incorrectly tagged as anatomical, whereas all the mistakes in the 1NN setting (5 in total) are the reverse.

Finally, we also observed that for cases with subtle disambiguation, both the strategies failed consistently in the one-shot setting. For instance, a common mistake shared by the two strategies was for cases where the context contained semantic cues for multiple senses, e.g., ``the English word \textit{digit} as well as its translation in many languages is also the anatomical term for fingers and toes.'' in which the intended meaning of \textit{digit} is the numerical one (both strategies failed on disambiguation this).
This observation is in line with the analysis of \citet{reif2019visualizing} which highlighted the failure of BERT in identifying semantic boundaries of words.

\paragraph{Pre-training label bias}
In most of the conventional supervised WSD classifiers (such as IMS), which rely on sense-annotated training data as their main source of information, the source of sense bias is usually the skewed distribution of instances for different senses of a word \cite{PilehvarNavigli:2014b}.
For instance, the word \textit{digit} would appear much more frequently with its numerical meaning than the finger meaning in an open-domain text.
Therefore, a sense-annotated corpus that is sampled from open-domain texts shows a similar sense distribution, resulting in a bias towards more frequent senses in the classification.

Given that in the $n$-shot setting we restrict the training datasets to have a uniform distribution of instances, sense bias in this scenario can be indicative of inherent sense biases in BERT's pre-training.
We observed that the pre-trained BERT indeed exhibits sense biases, often consistently across the two WSD strategies.
For instance, we observed the following biases towards (often) more frequent senses of words: \textit{java} towards its programming sense (rather than island), \textit{deck} towards ship deck (rather than building deck), \textit{yard} towards its sailing meaning (rather than measure unit), and \textit{digit} and \textit{square} towards their numerical meanings.
We also observed some contextual cues that misled the WSD system, especially in the one-shot setting.
For instance, we observed that our BERT-based WSD system had tendency to classify \textit{square} as its digit meaning whenever there was a \textit{number} in its context, e.g., ``marafor is a roman square with two temples attached'' or ``it has 4 trapezoid and 2 square faces''.
Not surprisingly, the source of most bias towards the digit sense of \textit{square} is from its geometrical sense (which has domain relatedness).
Also, classification for \textit{digit} was often biased towards its numerical meaning.
Similarly to the case of \textit{square}, the existence of a number in context seems to bias the model towards numerical meanings, e.g., ``There were five {\it digit} on each hand and four on each foot''.

\paragraph{Sensitivity to initialization}
We observed a high variation in the results, especially for the one-shot setting, suggesting the high sensitivity of the model with little evidence from training to the initialization point.
For instance, in the one-shot experiment for the fine-tuning model and the word \textit{bank}, in three runs, 1\%, 60\%, and 70\% of the test instances for the financial bank are incorrectly classified as river bank. Similarly, for \textit{crane}, 12\%, 25\%, and 72\% of the machine instances are misclassified as bird in three runs.
The 1NN strategy, in addition to being less prone to sense biases, is generally more robust across multiple runs. For the above two examples, the figures are 2\%, 0\%, and 0\% for \textit{bank} and 15\%, 0\%, and 27\% for \textit{crane}.
Other than the extent of bias, we observed that the direction can also change dramatically from run to run. For example, in the one-shot 1NN setting and for the word \textit{apple}, almost all the mistakes in the first two runs (37 of 38 and 12 of 14) were incorporation for fruit, whereas in the third run, almost all (6 of 7) were fruit for incorporation.

\section{Discussion}
\label{sec:discussion}

In the previous sections we have run an extensive set of experiments to investigate various properties of language models when adapted to the task of WSD. 
In the following we discuss some of the general conclusions and open questions arising from our analysis.

\paragraph{Fine-grained vs. coarse-grained}
A well-known issue of WordNet is the fine granularity of its sense distinctions \cite{navigli:09}. 
For example, the noun \textit{star} has 8 senses in WordNet, 
two of which refer to a ``celestial body'', only differing in if they are visible from the Earth or not.
Both meanings translate to \textit{estrella} in Spanish and therefore this sense distinction serves no advantage in MT, for example. 
In fact, it has been shown that coarse-grained distinctions are generally more suited to downstream applications \cite{rud-etal-2011-piggyback,severyn-etal-2013-learning,flekova-gurevych-2016-supersense,pilehvar-etal-2017-towards}. However, the coarsening of sense inventories is certainly not a solved task. While in this paper we relied either on experts for selecting senses from Wikipedia (given the reduced number of selected words) or domain labels from lexical resources for WordNet \cite{lacerra2020csi}, there are other strategies for coarsening sense inventories \cite{mccarthy2016word,hauer2020one}, for instance, based on translations or parallel corpora \cite{resnik1999distinguishing,apidianaki2008translation,bansal2012unsupervised}. This is generally an open problem, especially for verbs \cite{peterson2018bayesian}, which have not been analyzed in-depth in this article due to lack of effective techniques for an interpretable coarsening. Indeed, while in this work we have shown how contextualized embeddings encode meaning to a similar extent as humans do, for fine-grained distinctions these have been shown to correlate to a much lesser extent, an area that requires further exploration \cite{haber2020word}.

\paragraph{Fine-tuning vs. feature extraction (1NN)} The distinction between fine-tuning and feature extraction has been already studied in the literature for different tasks \cite{peters2019tune}. The general assumption is that fine-tuned models perform better when reasonable amounts of training data are available. In the case of WSD, however, feature extraction (specifically the 1NN strategy explained in this paper) is the more solid choice on general grounds, even when training data is available. The advantages of feature extraction (1NN) with respect to fine-tuning are threefold: 

\begin{enumerate}
    \item It is significantly less expensive to train as it simply relies on extracting contextualized embeddings from the training data. This is especially relevant when the WSD model is to be used in an all-words setting.
    \item It is more robust to changes in the training distribution (see Section \ref{an:distribution}).  
    \item It works reasonably well for limited amounts of training data (see Section \ref{trainingsize}), even in few-shot settings (see Section \ref{an:nshot}).
\end{enumerate}    

\paragraph{Few-shot learning} An important limitation of supervised WSD models is their dependence on sense-annotated corpora, which is expensive to construct, i.e., the so-called knowledge-acquisition bottleneck \cite{gale1992method,pasini2020knowledge}. Therefore, being able to learn from a limited set of examples is a desirable property of WSD models.
Encouragingly, as mentioned above, the simple 1NN method studied in this article shows robust results even with as few as three training examples per word sense. In the future it would be interesting to investigate models relying on knowledge from lexical resources that can perform WSD with no training instances available (i.e., zero-shot), in the line of \citet{kumar-etal-2019-zero} and \citet{blevins-zettlemoyer-2020-moving}.

\section{Conclusions}
\label{sec:conclusions}

In this paper we have provided an extensive analysis on how pre-trained language models (particularly BERT) capture lexical ambiguity. 
Our aim was to inspect the capability of BERT in predicting different usages of the same word depending on its context, similarly as humans do \cite{doi:10.1177/1745691619885860}.
The general conclusion we draw is that in the ideal setting of having access to enough amounts of training data and computing power, BERT can approach human-level performance for coarse-grained noun WSD, even in cross-domain scenarios. 
However, this ideal setting rarely occurs in practice, and challenges remain to make these models more efficient and less reliant on sense-annotated data. 
As an encouraging finding, feature extraction-based models (referred to as 1NN throughout the article) show strong performance even with a handful of examples per word sense. 
As future work it would be interesting to focus on the internal representation of the Transformer architecture by, e.g., carrying out an in-depth study of layer distribution \cite{tenney-etal-2019-bert}, investigating the importance of each attention head \cite{clark-etal-2019-bert}, or analyzing the differences for modeling concepts, entities and other categories of words, e.g., verbs. Moreover, our analysis could be extended to additional Transformer-based models, such as RoBERTa \cite{liu2019roberta} and T5 \cite{raffel2019exploring}.

To enable further analysis of this type, another contribution of the paper is the release of the CoarseWSD-20 dataset (Section \ref{sec:coarsewsd-20}), which also includes the out-of-domain test set (Section \ref{outofdomain}).
This dataset can be reliably used for quantitative and qualitative analyses in coarse-grained WSD, as we performed.
We hope that future research in WSD will take inspiration on the types of analyses performed in this paper, as they help shed light on the advantages and limitations of each approach. In particular, few-shot and bias analysis along with training distribution variations are key aspects to understand the versatility and robustness of any given approach.

Finally, WSD is clearly not a solved problem, even in the coarse-grained setting, due to a few challenges: (1) it is an arduous process to manually create high-quality full-coverage training data; therefore, future research should also focus on reliable ways of automising this process \cite{taghipour-ng-2015-one,dellibovietal:17,scarlini-etal-2019-just,pasini2019train,loureiro-camacho-collados-2020-dont,scarlini-etal-2020-contexts} and/or leveraging specific knowledge from lexical resources \cite{luo-etal-2018-incorporating,kumar-etal-2019-zero,huang-etal-2019-glossbert}; and (2) the existing sense-coarsening approaches are mainly targeted at nouns, and verb sense modelling remains an important open research challenge.

\section{Acknowledgments}

We would like to thank Claudio Delli Bovi and Miguel Ballesteros for early pre-BERT discussions on the topic of ambiguity and language models. We would also like to thank the anonymous reviewers for their comments and suggestions that helped improve the article. 
\\

\starttwocolumn
\bibliography{compling_style}
\onecolumnnew

\section*{APPENDIX}

\subsection*{Word in Context Evaluation}
\label{sec:wic}

Word-in-Context \cite[WiC]{pilehvar2019-wic} is a binary classification task from the SuperGLUE language understanding benchmark \cite{wang2019superglue} aimed at testing the ability of models to distinguish between different senses of the same word without relying on a pre-defined sense inventory. 
In particular, given a target word (either a verb or a noun) and two contexts where such target word occurs, the task consists of deciding whether the two target words in context refer to the same sense or not. Even though no sense inventory is explicitly given, this dataset was also constructed based on WordNet.
Table \ref{tab:wic-examples} shows a few examples from the dataset.

\paragraph{BERT-based model} Given that the task in WiC is a binary classification, the 1NN model is not applicable since a training to learn sense margins is necessary.
Therefore, we experimented with the BERT model fine-tuned on WiC's training data.
We followed \citet{wang2019superglue} and fused the two sentences and fed them as input to BERT. A classifier was then trained on the concatenation of the resulting BERT contextual embeddings.

\paragraph{Baselines} In addition to our BERT-based model, we include results for two FastText supervised classifiers \cite{joulin-etal-2017-bag} as baselines: a basic one with random initialization (FastText-B) and another initialized with FastText embeddings trained on the Common Crawl (FastText-C). 
As other indicative reference points, we added two language models which are enriched with WordNet \cite{levine2019sensebert,loureiro-jorge-2019-liaad} and another with WordNet and Wikipedia \cite{peters-etal-2019-knowledge}.

\paragraph{Results} Table \ref{tab:resultswic} shows the result of BERT models and the other baselines on the WiC benchmark.\footnote{Data and results from comparison systems taken from \url{https://pilehvar.github.io/wic/}} We can see that BERT significantly outperforms the FastText static word embedding. The two versions of BERT (Base and Large) perform equally well on this task, achieving results close to the state of the art. As with fine-grained all-words WSD, the additional knowledge drawn from WordNet proves to be beneficial, as shown by the results for KnowBERT and SenseBERT.

\begin{table}[t!]
\setlength{\tabcolsep}{4.0pt}
  \begin{tabular}{c p{0.95\textwidth}}
    \toprule
    \bf F &
    There's a lot of trash on the \textit{bed} of the river |
    I keep a glass of water next to my \textit{bed} when I sleep \\
    \bf F & 
    \textit{Justify} the margins |
    The end \textit{justifies} the means    \\
    \bf T &
    \textit{Air} pollution |
    Open a window and let in some \textit{air} \\
    \bf T &
    The expanded \textit{window} will give us time to catch the thieves |
    You have a two-hour \textit{window} of clear weather to finish working on the lawn \\
    \bottomrule    
  \end{tabular}
  \caption{Sample positive (T) and negative (F) pairs from the WiC dataset (target word in \textit{italics}).}
  \label{tab:wic-examples}
\end{table} 

\begin{table}
\small
\setlength{\tabcolsep}{22pt}
\resizebox{\textwidth}{!}{
\resizebox{\columnwidth}{!}{
\begin{tabular}{llc}
\toprule
\bf Type & \bf Model  &  \bf Accuracy  \\


\midrule
 \multirow{3}{*}{\bf Hybrid}    &  KnowBERT \cite{peters-etal-2019-knowledge}  & 70.9     
  \\
 &  SenseBERT \cite{levine2019sensebert}  & 72.1    \\ 
  &  LMMS-LR \cite{loureiro-jorge-2019-liaad}  & 68.1   \\

 \midrule
 \multirow{4}{*}{\bf \begin{minipage}{0.65in}Fine-tuned/ Supervised\end{minipage}} 
&  BERT-Base  & 69.6    \\ 
&  BERT-Large & 69.6    \\ 
&  FastText-B  & 52.3  \\
&  FastText-C  & 54.7   
  \\
   \midrule
   \bf Lowerbound	 &  \it Most Frequent Class & 50.0 \\  
   \bf Upperbound &  \it Human performance & 80.0  \\  
\bottomrule
\end{tabular}
}
}
\caption{\label{tab:resultswic} Accuracy (\%) performance of different models on the WiC dataset.}
\end{table}

\subsection*{CoarseWSD-20: Sense Information} 

Table \ref{tab:sense-definitions} shows for each sense their ID (as per their Wikipedia page title), definition and example usage from the dataset.

\subsection*{Complementary Results in CoarseWSD-20} 

\begin{enumerate}

\item Table \ref{table:micro_results} shows micro-F1 results for the experiment with different training data sizes sampled from the original CoarseWSD-20 training set (cf. Section \ref{trainingsize} of the paper).

\begin{table}[]
\setlength{\tabcolsep}{6.0pt}
\resizebox{\textwidth}{!}{
\begin{tabular}{lrrrrrrlrrrrrr}
\toprule
                 & \multicolumn{6}{c}{\textbf{Fine-tuning (BRT-L)}}            &  & \multicolumn{6}{c}{\textbf{1NN (BRT-L)}}      \\
\cmidrule(lr){2-7}
\cmidrule(lr){8-14}
                 & \multicolumn{1}{c}{\textbf{1\%}}      & \multicolumn{1}{c}{\textbf{5\%}} & \multicolumn{1}{c}{\textbf{10\%}} & \multicolumn{1}{c}{\textbf{25\%}} & \multicolumn{1}{c}{\textbf{50\%}} & \multicolumn{1}{c}{\textbf{ALL}} &  & \multicolumn{1}{c}{\textbf{1\%}}                   & \multicolumn{1}{c}{\textbf{5\%}}                   & \multicolumn{1}{c}{\textbf{10\%}} & \multicolumn{1}{c}{\textbf{25\%}} & \multicolumn{1}{c}{\textbf{50\%}} & \multicolumn{1}{c}{\textbf{ALL}} \\
\cmidrule(lr){2-7}
\cmidrule(lr){8-14}

{crane}   & \cellcolor[HTML]{F8DDDB}84.1\        & \cellcolor[HTML]{FEFCFC}95.8\   & \cellcolor[HTML]{FEFCFC}95.8\    & \cellcolor[HTML]{F2FAF6}96.8\    & \cellcolor[HTML]{FEFCFC}95.5\    & \cellcolor[HTML]{B4E1CB}98.1\   &  & \cellcolor[HTML]{FEFEFE}96.5\                     & \cellcolor[HTML]{F9FDFB}96.7\                     & \cellcolor[HTML]{F3FAF6}96.8\    & \cellcolor[HTML]{F3FAF6}96.8\    & \cellcolor[HTML]{F3FAF6}96.8\    & \cellcolor[HTML]{F3FAF6}96.8\   \\
{java}    & \cellcolor[HTML]{85CEAA}99.1\        & \cellcolor[HTML]{83CDA9}99.1\   & \cellcolor[HTML]{69C397}99.6\    & \cellcolor[HTML]{6FC59B}99.5\    & \cellcolor[HTML]{6BC398}99.6\    & \cellcolor[HTML]{67C295}99.7\   &  & \cellcolor[HTML]{6CC499}99.6\                     & \cellcolor[HTML]{6CC499}99.6\                     & \cellcolor[HTML]{6CC499}99.6\    & \cellcolor[HTML]{6CC499}99.6\    & \cellcolor[HTML]{6CC499}99.6\    & \cellcolor[HTML]{6CC499}99.6\   \\
{apple}   & \cellcolor[HTML]{77C8A1}99.4\        & \cellcolor[HTML]{75C89F}99.4\   & \cellcolor[HTML]{72C69D}99.5\    & \cellcolor[HTML]{72C69D}99.5\    & \cellcolor[HTML]{6FC59B}99.5\    & \cellcolor[HTML]{6AC398}99.6\   &  & \cellcolor[HTML]{7DCBA5}99.2\                     & \cellcolor[HTML]{7DCBA5}99.2\                     & \cellcolor[HTML]{7DCBA5}99.2\    & \cellcolor[HTML]{7DCBA5}99.2\    & \cellcolor[HTML]{7DCBA5}99.2\    & \cellcolor[HTML]{7DCBA5}99.2\   \\
{mole}    & \cellcolor[HTML]{F6D3D0}80.1\        & \cellcolor[HTML]{FEFDFD}96.0\   & \cellcolor[HTML]{C6E8D7}97.7\    & \cellcolor[HTML]{87CFAB}99.0\    & \cellcolor[HTML]{87CFAB}99.0\    & \cellcolor[HTML]{8FD2B1}98.9\   &  & \cellcolor[HTML]{C6E8D7}97.7\                     & \cellcolor[HTML]{9AD6B9}98.6\                     & \cellcolor[HTML]{9ED8BC}98.5\    & \cellcolor[HTML]{9ED8BC}98.5\    & \cellcolor[HTML]{9ED8BC}98.5\    & \cellcolor[HTML]{9ED8BC}98.5\   \\
{spring}  & \cellcolor[HTML]{FEFAFA}95.0\        & \cellcolor[HTML]{D0ECDF}97.5\   & \cellcolor[HTML]{F0F9F5}96.9\    & \cellcolor[HTML]{F4FBF7}96.8\    & \cellcolor[HTML]{C2E7D5}97.8\    & \cellcolor[HTML]{A9DDC3}98.3\   &  & \cellcolor[HTML]{BBE4D0}98.0\                     & \cellcolor[HTML]{BAE3CF}98.0\                     & \cellcolor[HTML]{BAE3CF}98.0\    & \cellcolor[HTML]{BBE4D0}98.0\    & \cellcolor[HTML]{BDE4D1}97.9\    & \cellcolor[HTML]{C2E7D5}97.8\   \\
{chair}   & \cellcolor[HTML]{F8DAD7}82.8\        & \cellcolor[HTML]{FDF7F6}93.6\   & \cellcolor[HTML]{FEFDFD}95.9\    & \cellcolor[HTML]{FAFDFB}96.7\    & \cellcolor[HTML]{EDF8F3}96.9\    & \cellcolor[HTML]{FEFDFD}96.2\   &  & \cellcolor[HTML]{FEFBFA}95.1\                     & \cellcolor[HTML]{FEFDFD}95.8\                     & \cellcolor[HTML]{FEFDFD}96.2\    & \cellcolor[HTML]{FEFDFD}96.2\    & \cellcolor[HTML]{FEFDFD}96.2\    & \cellcolor[HTML]{FEFDFD}96.2\   \\
{hood}    & \cellcolor[HTML]{F5CCC9}77.6\        & \cellcolor[HTML]{FCEFEE}90.7\   & \cellcolor[HTML]{FDF6F6}93.5\    & \cellcolor[HTML]{E2F4EB}97.2\    & \cellcolor[HTML]{CEECDD}97.6\    & \cellcolor[HTML]{6BC498}99.6\   &  & \cellcolor[HTML]{E2F4EB}97.2\                     & \cellcolor[HTML]{87CFAC}99.0\                     & \cellcolor[HTML]{75C89F}99.4\    & \cellcolor[HTML]{6BC498}99.6\    & \cellcolor[HTML]{57BB8A}100\   & \cellcolor[HTML]{57BB8A}100\  \\
{seal}    & \cellcolor[HTML]{FCF3F3}92.4\        & \cellcolor[HTML]{CCEBDC}97.6\   & \cellcolor[HTML]{B5E2CC}98.1\    & \cellcolor[HTML]{91D3B3}98.8\    & \cellcolor[HTML]{9FD8BC}98.5\    & \cellcolor[HTML]{89CFAD}99.0\   &  & \cellcolor[HTML]{B3E1CA}98.1\                     & \cellcolor[HTML]{B0DFC8}98.2\                     & \cellcolor[HTML]{ABDDC5}98.3\    & \cellcolor[HTML]{B1E0C9}98.2\    & \cellcolor[HTML]{AFDFC7}98.2\    & \cellcolor[HTML]{B5E2CC}98.1\   \\
{bow}     & \cellcolor[HTML]{F3C3BF}74.1\        & \cellcolor[HTML]{FCF3F3}92.4\   & \cellcolor[HTML]{FEFDFD}96.1\    & \cellcolor[HTML]{F6FCF9}96.7\    & \cellcolor[HTML]{D0ECDF}97.5\    & \cellcolor[HTML]{A3DABF}98.5\   &  & \cellcolor[HTML]{FEFAFA}94.9\                     & \cellcolor[HTML]{FEFDFD}95.9\                     & \cellcolor[HTML]{FEFDFC}95.8\    & \cellcolor[HTML]{FEFCFC}95.5\    & \cellcolor[HTML]{FEFBFB}95.3\    & \cellcolor[HTML]{FEFBFB}95.3\   \\
{club}    & \cellcolor[HTML]{F2BFBB}72.8\        & \cellcolor[HTML]{F5CFCC}78.7\   & \cellcolor[HTML]{F5CFCC}78.7\    & \cellcolor[HTML]{F6D3D0}80.4\    & \cellcolor[HTML]{F8DCD9}83.5\    & \cellcolor[HTML]{F8DFDD}84.7\   &  & \cellcolor[HTML]{F7D8D5}82.0\                     & \cellcolor[HTML]{F8DAD8}82.9\                     & \cellcolor[HTML]{F8DDDA}83.8\    & \cellcolor[HTML]{F8DDDB}84.0\    & \cellcolor[HTML]{F8DEDC}84.4\    & \cellcolor[HTML]{F9E0DE}85.1\   \\
{trunk}   & \cellcolor[HTML]{F9E3E1}86.2\        & \cellcolor[HTML]{FBEAE8}88.7\   & \cellcolor[HTML]{C1E6D4}97.8\    & \cellcolor[HTML]{97D5B7}98.7\    & \cellcolor[HTML]{97D5B7}98.7\    & \cellcolor[HTML]{ACDEC5}98.3\   &  & \cellcolor[HTML]{C1E6D4}97.8\                     & \cellcolor[HTML]{B0DFC8}98.2\                     & \cellcolor[HTML]{A7DBC2}98.4\    & \cellcolor[HTML]{97D5B7}98.7\    & \cellcolor[HTML]{97D5B7}98.7\    & \cellcolor[HTML]{97D5B7}98.7\   \\
{square}  & \cellcolor[HTML]{FAE9E7}88.4\        & \cellcolor[HTML]{FAE6E4}87.3\   & \cellcolor[HTML]{FCF4F3}92.6\    & \cellcolor[HTML]{FCF4F3}92.4\    & \cellcolor[HTML]{FDF5F4}92.9\    & \cellcolor[HTML]{FEFCFC}95.7\   &  & \cellcolor[HTML]{FDF7F7}93.9\                     & \cellcolor[HTML]{FDF8F8}94.2\                     & \cellcolor[HTML]{FDF8F7}94.1\    & \cellcolor[HTML]{FEFBFB}95.2\    & \cellcolor[HTML]{FEFCFC}95.7\    & \cellcolor[HTML]{FEFDFD}96.1\   \\
{arm}     & \cellcolor[HTML]{FDF5F5}93.1\        & \cellcolor[HTML]{9DD7BB}98.6\   & \cellcolor[HTML]{75C89F}99.4\    & \cellcolor[HTML]{75C89F}99.4\    & \cellcolor[HTML]{75C89F}99.4\    & \cellcolor[HTML]{75C89F}99.4\   &  & \cellcolor[HTML]{75C89F}99.4\                     & \cellcolor[HTML]{75C89F}99.4\                     & \cellcolor[HTML]{75C89F}99.4\    & \cellcolor[HTML]{75C89F}99.4\    & \cellcolor[HTML]{75C89F}99.4\    & \cellcolor[HTML]{75C89F}99.4\   \\
{digit}   & \cellcolor[HTML]{FEFBFB}95.2\        & \cellcolor[HTML]{FBECEB}89.7\   & \cellcolor[HTML]{FEFBFB}95.2\    & \cellcolor[HTML]{7ECBA5}99.2\    & \cellcolor[HTML]{57BB8A}100\   & \cellcolor[HTML]{57BB8A}100\  &  & \cellcolor[HTML]{6BC398}99.6\                     & \cellcolor[HTML]{57BB8A}100\                    & \cellcolor[HTML]{57BB8A}100\   & \cellcolor[HTML]{57BB8A}100\   & \cellcolor[HTML]{57BB8A}100\   & \cellcolor[HTML]{57BB8A}100\  \\
{bass}    & \cellcolor[HTML]{FCF2F2}92.0\        & \cellcolor[HTML]{FDF6F6}93.4\   & \cellcolor[HTML]{FDF9F9}94.4\    & \cellcolor[HTML]{FEFBFA}95.1\    & \cellcolor[HTML]{FEFCFC}95.6\    & \cellcolor[HTML]{FEFDFC}95.8\   &  & \cellcolor[HTML]{F9E4E2}86.6\                     & \cellcolor[HTML]{F9E3E1}86.1\                     & \cellcolor[HTML]{F9E2E0}85.8\    & \cellcolor[HTML]{F9E1DF}85.5\    & \cellcolor[HTML]{F9E0DE}85.2\    & \cellcolor[HTML]{F8DEDC}84.5\   \\
{yard}    & \cellcolor[HTML]{FCEFEE}90.7\        & \cellcolor[HTML]{FDF8F7}94.0\   & \cellcolor[HTML]{FEFBFB}95.4\    & \cellcolor[HTML]{DFF2E9}97.2\    & \cellcolor[HTML]{6EC59A}99.5\    & \cellcolor[HTML]{6EC59A}99.5\   &  & \cellcolor[HTML]{FBEDEB}89.8\                     & \cellcolor[HTML]{FBEAE9}88.9\                     & \cellcolor[HTML]{FAE7E6}87.8\    & \cellcolor[HTML]{FAE6E5}87.5\    & \cellcolor[HTML]{FAE5E3}86.8\    & \cellcolor[HTML]{FBEAE9}88.9\   \\
{pound}   & \cellcolor[HTML]{FAE9E7}88.3\        & \cellcolor[HTML]{FBEDEC}90.0\   & \cellcolor[HTML]{FBECEB}89.7\    & \cellcolor[HTML]{FBEAE9}89.0\    & \cellcolor[HTML]{FEFAFA}94.9\    & \cellcolor[HTML]{FEFAFA}94.9\   &  & \cellcolor[HTML]{FDF4F3}92.6\                     & \cellcolor[HTML]{FDF4F4}92.8\                     & \cellcolor[HTML]{FCF2F2}92.0\    & \cellcolor[HTML]{FBEEED}90.4\    & \cellcolor[HTML]{FBECEB}89.7\    & \cellcolor[HTML]{FBECEB}89.7\   \\
{deck}    & \cellcolor[HTML]{FDF7F6}93.6\        & \cellcolor[HTML]{FDF5F4}92.9\   & \cellcolor[HTML]{FDF5F4}92.9\    & \cellcolor[HTML]{FDF8F7}93.9\    & \cellcolor[HTML]{FEFAFA}95.0\    & \cellcolor[HTML]{FEFBFB}95.3\   &  & \cellcolor[HTML]{FCF1F0}91.4\                     & \cellcolor[HTML]{FCF2F1}91.9\                     & \cellcolor[HTML]{FCF2F1}91.7\    & \cellcolor[HTML]{FCF1F0}91.6\    & \cellcolor[HTML]{FCF1F0}91.4\    & \cellcolor[HTML]{FCF2F1}91.9\   \\
{bank}    & \cellcolor[HTML]{FEFBFB}95.2\        & \cellcolor[HTML]{FEFCFC}95.5\   & \cellcolor[HTML]{E6F5EE}97.1\    & \cellcolor[HTML]{70C69C}99.5\    & \cellcolor[HTML]{7BCAA3}99.3\    & \cellcolor[HTML]{7BCAA3}99.3\   &  & \cellcolor[HTML]{66C194}99.7\                     & \cellcolor[HTML]{5EBE8F}99.9\                     & \cellcolor[HTML]{5DBE8E}99.9\    & \cellcolor[HTML]{5FBE8F}99.9\    & \cellcolor[HTML]{62C092}99.8\    & \cellcolor[HTML]{62C092}99.8\   \\
{pitcher} & \cellcolor[HTML]{6EC59A}99.5\        & \cellcolor[HTML]{6DC49A}99.6\   & \cellcolor[HTML]{6AC397}99.6\    & \cellcolor[HTML]{5ABD8D}99.9\    & \cellcolor[HTML]{58BC8B}100\   & \cellcolor[HTML]{5ABC8C}100\  &  & \cellcolor[HTML]{5ABC8C}100\                    & \cellcolor[HTML]{5ABC8C}100\                    & \cellcolor[HTML]{5ABD8C}99.9\    & \cellcolor[HTML]{5ABC8C}100\   & \cellcolor[HTML]{5ABD8C}99.9\    & \cellcolor[HTML]{5BBD8D}99.9\   \\
\midrule
{\it Average}          & \cellcolor[HTML]{FBEAE9}89.0\        & \cellcolor[HTML]{FDF6F6}93.5\   & \cellcolor[HTML]{FEFBFB}95.3\    & \cellcolor[HTML]{FEFEFE}96.3\    & \cellcolor[HTML]{E8F6EF}97.0\    & \cellcolor[HTML]{D0ECDF}97.5\   &  & \multicolumn{1}{c}{\cellcolor[HTML]{FEFCFB}95.5} & \multicolumn{1}{c}{\cellcolor[HTML]{FEFCFC}95.8} & \cellcolor[HTML]{FEFCFC}95.7\    & \cellcolor[HTML]{FEFCFC}95.7\    & \cellcolor[HTML]{FEFCFC}95.6\    & \cellcolor[HTML]{FEFCFC}95.8 \\
\bottomrule
\end{tabular}
}
\caption{Micro-F1 results on the CoarseWSD-20 test set using training sets of different sizes sampled from the original training set.\label{table:micro_results}}
\end{table}

\item Table \ref{table:mfs_results} shows the micro-F1 performance for fine-tuning and 1NN and for varying sizes of the training data (with similar skewed distributions) for both Most Frequent Sense (MFS) and Least Frequent Sense (LFS) classes (cf. Section \ref{trainingsize} of the paper).

\begin{table}[]
\setlength{\tabcolsep}{5.0pt}
\resizebox{\textwidth}{!}{
\begin{tabular}{lrrrrrrlrrrrrr}
\toprule
& \multicolumn{12}{c}{\bf Most Frequent Sense (MFS)}\\
\midrule

                 & \multicolumn{6}{c}{\textbf{Fine-tuning (BRT-L)}}            &  & \multicolumn{6}{c}{\textbf{1NN (BRT-L)}}      \\
\cmidrule(lr){2-7}
\cmidrule(lr){8-14}
                 & \multicolumn{1}{c}{\textbf{1\%}}      & \multicolumn{1}{c}{\textbf{5\%}} & \multicolumn{1}{c}{\textbf{10\%}} & \multicolumn{1}{c}{\textbf{25\%}} & \multicolumn{1}{c}{\textbf{50\%}} & \multicolumn{1}{c}{\textbf{ALL}} &  & \multicolumn{1}{c}{\textbf{1\%}}                   & \multicolumn{1}{c}{\textbf{5\%}}                   & \multicolumn{1}{c}{\textbf{10\%}} & \multicolumn{1}{c}{\textbf{25\%}} & \multicolumn{1}{c}{\textbf{50\%}} & \multicolumn{1}{c}{\textbf{ALL}} \\
\cmidrule(lr){2-7}
\cmidrule(lr){8-14}
crane    &   86.9   &   96.1   &   96.0   &   97.0   &   95.9   &   98.2   &      &   100   &   100   &   100   &   100   &   100   &   100\\ 
java   &   99.2   &   99.3   &   99.7   &   99.6   &   99.7   &   99.8   &      &   99.4   &   99.4   &   99.4   &   99.4   &   99.4   &   99.4\\ 
apple   &   99.5   &   99.5   &   99.6   &   99.6   &   99.6   &   99.7   &      &   99.5   &   99.5   &   99.5   &   99.5   &   99.5   &   99.5\\ 
mole   &   75.4   &   96.2   &   97.1   &   98.7   &   98.7   &   98.5   &      &   95.2   &   97.7   &   97.4   &   97.4   &   97.4   &   97.4\\ 
spring   &   96.0   &   97.7   &   97.2   &   97.0   &   98.0   &   98.8   &      &   97.7   &   97.8   &   97.8   &   97.7   &   97.7   &   97.5\\ 
chair   &   88.8   &   95.5   &   97.0   &   97.6   &   97.8   &   97.2   &      &   96.6   &   98.2   &   98.9   &   98.9   &   98.9   &   98.9\\ 
hood   &   91.6   &   93.4   &   95.6   &   98.3   &   97.9   &   99.7   &      &   100   &   100   &   100   &   100   &   100   &   100\\ 
seal   &   90.9   &   97.0   &   97.5   &   98.7   &   98.4   &   98.9   &      &   98.6   &   98.5   &   98.5   &   98.2   &   98.5   &   98.5\\ 
bow   &   85.5   &   97.7   &   97.2   &   98.2   &   98.2   &   98.7   &      &   97.6   &   98.1   &   98.3   &   98.3   &   98.3   &   98.3\\ 
club   &   74.6   &   80.4   &   80.6   &   81.9   &   84.1   &   85.2   &      &   79.5   &   78.5   &   78.5   &   78.4   &   78.2   &   77.8\\ 
trunk   &   90.6   &   91.4   &   98.2   &   98.9   &   98.9   &   98.6   &      &   97.9   &   97.9   &   97.9   &   97.9   &   97.9   &   97.9\\ 
square   &   89.6   &   88.5   &   93.0   &   92.8   &   93.2   &   95.7   &      &   93.7   &   93.6   &   93.4   &   95.8   &   95.1   &   94.2\\ 
arm   &   95.5   &   99.1   &   99.6   &   99.6   &   99.6   &   99.6   &      &   99.2   &   99.2   &   99.2   &   99.2   &   99.2   &   99.2\\ 
digit   &   97.0   &   93.9   &   97.1   &   99.5   &   100   &   100   &      &   100   &   100   &   100   &   100   &   100   &   100\\ 
bass   &   95.2   &   96.0   &   96.8   &   96.7   &   97.1   &   97.2   &      &   86.0   &   85.2   &   84.6   &   84.1   &   83.7   &   82.9\\ 
yard   &   94.4   &   96.6   &   97.4   &   98.4   &   99.7   &   99.7   &      &   88.3   &   86.9   &   85.7   &   85.2   &   84.4   &   86.9\\ 
pound   &   93.7   &   94.7   &   94.6   &   94.1   &   97.2   &   97.2   &      &   94.1   &   92.9   &   91.7   &   89.7   &   88.5   &   88.5\\ 
deck   &   96.7   &   96.3   &   96.3   &   96.8   &   97.3   &   97.5   &      &   92.4   &   93.0   &   92.1   &   91.7   &   91.3   &   91.3\\ 
bank   &   97.6   &   97.7   &   98.5   &   99.7   &   99.6   &   99.6   &      &   100   &   100   &   100   &   100   &   100   &   100\\ 
pitcher   &   99.8   &   99.8   &   99.8   &   100   &   100   &   100   &      &   100   &   100   &   99.9   &   100   &   99.9   &   99.9\\ 
\midrule
\it Average  &   91.9   &   95.3   &   96.4   &   97.2   &   97.5   &   98.0   &      &   95.8   &   95.8   &   95.6   &   95.6   &   95.4   &   95.4 \\
\\
\toprule
& \multicolumn{12}{c}{\bf Least Frequent Sense (LFS)}\\
\midrule
                 & \multicolumn{6}{c}{\textbf{Fine-tuning (BRT-L)}}            &  & \multicolumn{6}{c}{\textbf{1NN (BRT-L)}}      \\
\cmidrule(lr){2-7}
\cmidrule(lr){8-14}
                 & \multicolumn{1}{c}{\textbf{1\%}}      & \multicolumn{1}{c}{\textbf{5\%}} & \multicolumn{1}{c}{\textbf{10\%}} & \multicolumn{1}{c}{\textbf{25\%}} & \multicolumn{1}{c}{\textbf{50\%}} & \multicolumn{1}{c}{\textbf{ALL}} &  & \multicolumn{1}{c}{\textbf{1\%}}                   & \multicolumn{1}{c}{\textbf{5\%}}                   & \multicolumn{1}{c}{\textbf{10\%}} & \multicolumn{1}{c}{\textbf{25\%}} & \multicolumn{1}{c}{\textbf{50\%}} & \multicolumn{1}{c}{\textbf{ALL}} \\
\cmidrule(lr){2-7}
\cmidrule(lr){8-14}

crane   &  79.7  &  95.4  &  95.4  &  96.6  &  95.2  &  98.0  &&  92.8  &  93.2  &  93.4  &  93.4  &  93.4  &  93.4\\ 
 java  &  98.8  &  98.8  &  99.5  &  99.4  &  99.5  &  99.6  &&  99.9  &  99.9  &  99.9  &  99.9  &  99.9  &  99.9\\ 
 apple  &  99.2  &  99.2  &  99.3  &  99.3  &  99.4  &  99.5  &&  98.7  &  98.7  &  98.7  &  98.7  &  98.7  &  98.7\\ 
 mole  &  72.5  &  86.7  &  97.4  &  100  &  100  &  100  &&  100  &  100  &  100  &  100  &  100  &  100\\ 
 spring  &  95.0  &  98.2  &  97.0  &  97.9  &  97.9  &  97.5  &&  97.3  &  97.3  &  97.3  &  97.3  &  97.3  &  97.3\\ 
 chair  &  63.7  &  88.9  &  93.4  &  94.6  &  95.0  &  93.8  &&  92.1  &  91.0  &  90.5  &  90.5  &  90.5  &  90.5\\ 
 hood  &  65.7  &  82.5  &  89.2  &  95.2  &  95.2  &  99.2  &&  97.0  &  97.3  &  97.7  &  98.5  &  100  &  100\\ 
 seal  &  39.1  &  89.3  &  91.0  &  96.0  &  96.0  &  97.3  &&  100  &  100  &  100  &  100  &  100  &  100\\ 
 bow  &  0.0  &  73.2  &  95.3  &  94.6  &  98.0  &  99.4  &&  91.0  &  98.5  &  100  &  100  &  100  &  100\\ 
 club  &  63.5  &  74.4  &  73.3  &  80.0  &  81.6  &  82.5  &&  95.2  &  95.2  &  95.2  &  95.2  &  95.2  &  95.2\\ 
 trunk  &  49.7  &  66.6  &  95.2  &  100  &  100  &  96.5  &&  95.2  &  97.1  &  98.2  &  100  &  100  &  100\\ 
 square  &  9.5  &  21.4  &  4.8  &  20.5  &  30.3  &  76.0  &&  53.8  &  58.5  &  57.7  &  56.4  &  69.2  &  84.6\\ 
 arm  &  84.7  &  97.2  &  98.9  &  98.9  &  98.9  &  98.9  &&  100  &  100  &  100  &  100  &  100  &  100\\ 
 digit  &  87.7  &  65.5  &  87.2  &  98.0  &  100  &  100  &&  98.1  &  100  &  100  &  100  &  100  &  100\\ 
 bass  &  29.5  &  48.2  &  61.8  &  65.6  &  68.5  &  67.3  &&  69.7  &  73.2  &  75.6  &  77.7  &  78.4  &  77.3\\ 
 yard  &  70.9  &  74.8  &  79.2  &  90.3  &  98.4  &  98.4  &&  98.5  &  100  &  100  &  100  &  100  &  100\\ 
 pound  &  13.3  &  6.1  &  0.0  &  11.1  &  69.3  &  70.6  &&  80.0  &  92.0  &  95.0  &  96.7  &  100  &  100\\ 
 deck  &  16.7  &  0.0  &  0.0  &  43.6  &  57.1  &  58.6  &&  78.6  &  77.1  &  85.7  &  90.5  &  92.9  &  100\\ 
 bank  &  2.9  &  14.0  &  51.4  &  94.4  &  91.8  &  91.6  &&  93.9  &  97.3  &  97.7  &  97.0  &  95.5  &  95.5\\ 
 pitcher  &  0.0  &  4.8  &  28.0  &  93.0  &  98.7  &  94.6  &&  100  &  100  &  100  &  100  &  100  &  100\\ 
 \midrule
 \it Average &  52.1  &  64.3  &  71.9  &  83.4  &  88.5  &  91.0  &&  91.6  &  93.3  &  94.1  &  94.6  &  95.5  &  96.6\\
 \bottomrule
\end{tabular}
}
\caption{Micro-F1 performance for the two WSD strategies and for varying sizes of the training data (with similar skewed distributions) for the MFS (top) and LFS (bottom). \label{table:mfs_results}}
\end{table}

\item Table \ref{tab:appendixfewshot} includes the complete results for the $n$-shot experiment, including the FastText baselines (cf. Section \ref{an:nshot} of the paper).

\end{enumerate}

\begin{landscape}
\small
\begin{longtabu} to 1.0\textwidth {cll | cccc | cccc}
\toprule
\multicolumn{1}{l}{}     &                                 &            & \multicolumn{4}{c}{\textbf{Micro F1}}                & \multicolumn{4}{|c}{\textbf{Macro F1}}                \\
  \cmidrule(lr){4-7}
  \cmidrule(lr){8-11}

\multicolumn{1}{l}{}     &                                 &            & 1           & 3           & 10          & 30         & 1           & 3           & 10          & 30         \\
\midrule
\endhead
\multirow{6}{*}{crane}   & \multirow{2}{*}{Static emb.} & Fasttext-B & 48.6 (5.3)  & 57.5 (5.3)  & 57.7 (3.5)  & 70.9 (5.2) & 48.1 (4.5)  & 57.9 (4.8)  & 58.5 (3.3)  & 71.1 (5.3) \\
                         &                                 & Fasttext-C & 52.7 (1.2)  & 69.4 (6.3)  & 82.0 (3.0)  & 83.4 (6.6) & 51.4 (1.1)  & 69.6 (6.3)  & 81.9 (3.0)  & 83.5 (6.8) \\
                         \cmidrule(lr){2-11}
                         & \multirow{2}{*}{1NN}            & BERT-Base  & 84.5 (9.8)  & 93.8 (1.8)  & 93.6 (1.4)  & 94.5 (0.3) & 84.3 (10.1) & 93.7 (1.9)  & 93.4 (1.4)  & 94.4 (0.3) \\
                         &                                 & BERT-Large & 86.4 (3.8)  & 94.7 (3.5)  & 95.5 (1.4)  & 96.8 (0.9) & 86.4 (3.9)  & 94.5 (3.7)  & 95.4 (1.4)  & 96.7 (0.9) \\
                         \cmidrule(lr){2-11}
                         & \multirow{2}{*}{Fine-Tuning}    & BERT-Base  & 65.6 (1.9)  & 88.7 (7.8)  & 94.1 (3.2)  & 95.8 (0.6) & 63.4 (1.3)  & 88.7 (7.8)  & 94.0 (3.3)  & 95.7 (0.6) \\
                         &                                 & BERT-Large & 74.7 (10.5) & 92.6 (2.2)  & 95.3 (1.7)  & 96.4 (1.3) & 73.2 (12.3) & 92.5 (2.2)  & 95.3 (1.7)  & 96.4 (1.3) \\
\midrule
\multirow{6}{*}{java}    & \multirow{2}{*}{Static emb.} & Fasttext-B & 63.1 (1.3)  & 66.8 (2.4)  & 68.5 (2.8)  & 80.6 (6.8) & 55.8 (3.5)  & 60.5 (3.9)  & 66.9 (1.6)  & 80.9 (6.4) \\
                         &                                 & Fasttext-C & 78.4 (5.8)  & 90.1 (3.3)  & 90.9 (1.9)  & 94.9 (2.3) & 78.1 (7.8)  & 90.3 (2.5)  & 90.5 (2.3)  & 95.1 (2.1) \\
                         \cmidrule(lr){2-11}
                         & \multirow{2}{*}{1NN}            & BERT-Base  & 99.6 (0.0)  & 99.6 (0.0)  & 99.6 (0.0)  & 99.6 (0.0) & 99.6 (0.0)  & 99.6 (0.0)  & 99.7 (0.0)  & 99.7 (0.0) \\
                         &                                 & BERT-Large & 99.6 (0.1)  & 99.6 (0.1)  & 99.6 (0.0)  & 99.6 (0.0) & 99.7 (0.0)  & 99.7 (0.1)  & 99.7 (0.0)  & 99.7 (0.0) \\
                         \cmidrule(lr){2-11}
                         & \multirow{2}{*}{Fine-Tuning}    & BERT-Base  & 99.2 (0.4)  & 98.8 (0.7)  & 99.4 (0.2)  & 99.3 (0.1) & 99.1 (0.5)  & 98.8 (0.7)  & 99.4 (0.2)  & 99.3 (0.1) \\
                         &                                 & BERT-Large & 99.2 (0.6)  & 99.4 (0.1)  & 99.5 (0.1)  & 99.6 (0.1) & 99.1 (0.6)  & 99.4 (0.1)  & 99.5 (0.1)  & 99.5 (0.1) \\
\midrule
\multirow{6}{*}{apple}   & \multirow{2}{*}{Static emb.} & Fasttext-B & 43.1 (2.1)  & 52.0 (4.9)  & 55.2 (8.4)  & 74.7 (1.2) & 45.4 (3.8)  & 55.3 (5.1)  & 61.3 (5.4)  & 73.9 (2.4) \\
                         &                                 & Fasttext-C & 71.2 (9.9)  & 81.2 (2.9)  & 87.3 (2.1)  & 93.2 (0.2) & 63.7 (12.8) & 80.7 (1.8)  & 86.7 (3.0)  & 92.4 (0.3) \\
                         \cmidrule(lr){2-11}
                         & \multirow{2}{*}{1NN}            & BERT-Base  & 95.5 (3.9)  & 99.0 (0.1)  & 99.0 (0.1)  & 99.0 (0.0) & 96.1 (3.2)  & 99.0 (0.1)  & 99.0 (0.1)  & 99.0 (0.0) \\
                         &                                 & BERT-Large & 98.1 (1.2)  & 99.2 (0.0)  & 99.3 (0.1)  & 99.3 (0.1) & 98.3 (0.9)  & 99.2 (0.1)  & 99.2 (0.1)  & 99.3 (0.1) \\
                         \cmidrule(lr){2-11}
                         & \multirow{2}{*}{Fine-Tuning}    & BERT-Base  & 90.7 (9.0)  & 98.3 (0.6)  & 99.0 (0.1)  & 99.0 (0.1) & 89.6 (10.3) & 98.2 (0.7)  & 98.9 (0.1)  & 98.9 (0.1) \\
                         &                                 & BERT-Large & 91.5 (5.5)  & 96.4 (2.5)  & 99.3 (0.1)  & 99.1 (0.5) & 90.7 (6.2)  & 96.2 (2.6)  & 99.2 (0.1)  & 99.0 (0.5) \\
\midrule
\multirow{6}{*}{mole}    & \multirow{2}{*}{Static emb.} & Fasttext-B & 21.2 (9.9)  & 16.3 (7.3)  & 38.2 (1.1)  & 65.9 (1.6) & 17.9 (2.2)  & 22.8 (3.9)  & 41.5 (9.2)  & 68.8 (2.0) \\
                         &                                 & Fasttext-C & 48.7 (2.6)  & 63.3 (4.3)  & 75.9 (6.3)  & 88.0 (0.9) & 57.3 (1.3)  & 68.6 (2.5)  & 79.4 (4.3)  & 89.4 (1.7) \\
                         \cmidrule(lr){2-11}
                         & \multirow{2}{*}{1NN}            & BERT-Base  & 75.9 (5.5)  & 91.1 (4.0)  & 95.1 (2.2)  & 97.4 (0.6) & 84.9 (4.6)  & 93.9 (1.8)  & 96.6 (1.2)  & 97.7 (0.3) \\
                         &                                 & BERT-Large & 89.3 (1.1)  & 95.6 (0.8)  & 98.1 (0.8)  & 98.5 (0.0) & 93.4 (0.6)  & 97.1 (0.7)  & 98.8 (0.4)  & 99.0 (0.0) \\
                         \cmidrule(lr){2-11}
                         & \multirow{2}{*}{Fine-Tuning}    & BERT-Base  & 71.2 (4.1)  & 86.2 (5.2)  & 95.8 (1.3)  & 97.6 (0.4) & 70.7 (4.8)  & 87.8 (3.3)  & 95.8 (1.4)  & 97.5 (0.6) \\
                         &                                 & BERT-Large & 77.3 (2.2)  & 88.7 (4.3)  & 96.3 (0.9)  & 98.5 (0.7) & 76.4 
                         (2.0)  & 90.2 (2.9)  & 96.3 (1.0)  & 98.8 (0.7) \\
                         \\
\midrule
\multirow{6}{*}{spring}  & \multirow{2}{*}{Static emb.} & Fasttext-B & 33.0 (6.8)  & 43.8 (8.2)  & 46.4 (7.4)  & 67.0 (2.2) & 35.4 (0.5)  & 32.8 (0.7)  & 35.8 (3.6)  & 66.7 (3.1) \\
                         &                                 & Fasttext-C & 46.0 (14.6) & 57.6 (4.0)  & 73.5 (3.7)  & 83.7 (2.8) & 45.0 (9.0)  & 64.2 (3.3)  & 76.5 (3.5)  & 86.1 (2.8) \\
                         \cmidrule(lr){2-11}
                         & \multirow{2}{*}{1NN}            & BERT-Base  & 94.4 (2.0)  & 97.2 (0.5)  & 97.1 (0.3)  & 97.3 (0.1) & 94.6 (2.2)  & 97.3 (0.9)  & 97.0 (0.4)  & 97.3 (0.1) \\
                         &                                 & BERT-Large & 97.1 (1.5)  & 97.9 (0.5)  & 97.6 (0.4)  & 97.7 (0.2) & 96.8 (1.6)  & 97.8 (0.2)  & 97.5 (0.3)  & 97.8 (0.1) \\
                         \cmidrule(lr){2-11}
                         & \multirow{2}{*}{Fine-Tuning}    & BERT-Base  & 75.2 (4.7)  & 92.9 (0.3)  & 96.0 (0.5)  & 95.3 (0.5) & 73.9 (4.3)  & 92.9 (0.2)  & 96.1 (0.5)  & 95.2 (0.6) \\
                         &                                 & BERT-Large & 80.3 (10.1) & 94.2 (2.7)  & 97.0 (0.7)  & 97.2 (0.2) & 77.1 (12.6) & 94.4 (2.4)  & 97.1 (0.6)  & 97.1 (0.4) \\
\midrule
\multirow{6}{*}{chair}   & \multirow{2}{*}{Static emb.} & Fasttext-B & 62.8 (9.0)  & 73.8 (6.6)  & 74.4 (4.5)  & 74.4 (5.2) & 58.4 (6.9)  & 68.4 (5.1)  & 68.4 (4.3)  & 72.1 (4.2) \\
                         &                                 & Fasttext-C & 76.2 (8.0)  & 75.4 (4.5)  & 81.3 (2.0)  & 83.6 (2.4) & 64.1 (12.8) & 72.9 (0.2)  & 75.8 (0.9)  & 81.7 (2.2) \\
                         \cmidrule(lr){2-11}
                         & \multirow{2}{*}{1NN}            & BERT-Base  & 88.7 (7.1)  & 95.9 (0.7)  & 95.9 (0.4)  & 95.6 (0.4) & 84.2 (11.6) & 94.5 (0.5)  & 94.5 (0.3)  & 94.3 (0.3) \\
                         &                                 & BERT-Large & 82.3 (19.0) & 94.6 (1.3)  & 96.2 (0.0)  & 95.9 (0.4) & 84.4 (14.1) & 93.5 (0.9)  & 94.7 (0.0)  & 94.5 (0.3) \\
                         \cmidrule(lr){2-11}
                         & \multirow{2}{*}{Fine-Tuning}    & BERT-Base  & 84.1 (11.3) & 91.0 (5.7)  & 95.6 (0.7)  & 96.7 (0.4) & 75.6 (21.8) & 90.1 (5.8)  & 94.9 (0.8)  & 96.1 (0.4) \\
                         &                                 & BERT-Large & 72.1 (11.9) & 93.3 (1.5)  & 94.4 (2.0)  & 95.1 (1.5) & 65.4 (12.9) & 92.4 (1.7)  & 93.6 (2.1)  & 94.4 (1.6) \\
\midrule
\multirow{6}{*}{hood}    & \multirow{2}{*}{Static emb.} & Fasttext-B & 56.1 (11.3) & 33.3 (16.2) & 47.6 (18.7) & -          & 48.8 (4.1)  & 42.1 (6.3)  & 51.4 (5.1)  & -          \\
                         &                                 & Fasttext-C & 66.3 (5.0)  & 77.6 (2.1)  & 86.6 (5.0)  & -          & 61.5 (6.2)  & 73.7 (4.7)  & 82.1 (6.8)  & -          \\
                         \cmidrule(lr){2-11}
                         & \multirow{2}{*}{1NN}            & BERT-Base  & 96.8 (3.0)  & 98.4 (0.6)  & 98.8 (0.0)  & -          & 94.8 (5.9)  & 98.0 (0.7)  & 98.5 (0.0)  & -          \\
                         &                                 & BERT-Large & 96.3 (4.3)  & 99.2 (0.6)  & 99.6 (0.6)  & -          & 92.7 (9.3)  & 99.0 (0.7)  & 99.5 (0.7)  & -          \\
                         \cmidrule(lr){2-11}
                         & \multirow{2}{*}{Fine-Tuning}    & BERT-Base  & 86.6 (3.6)  & 95.5 (2.5)  & 97.6 (1.7)  & -          & 79.0 (6.4)  & 94.4 (3.0)  & 96.7 (2.4)  & -          \\
                         &                                 & BERT-Large & 89.8 (5.8)  & 96.3 (1.0)  & 98.8 (1.0)  & -          & 80.1 (16.2) & 95.1 (1.5)  & 98.0 (1.7)  & -          \\
\midrule
\multirow{6}{*}{seal}    & \multirow{2}{*}{Static emb.} & Fasttext-B & 29.9 (1.0)  & 31.6 (3.2)  & 39.9 (10.0) & 60.2 (4.2) & 25.0 (0.0)  & 25.7 (1.0)  & 39.8 (5.6)  & 57.4 (1.1) \\
                         &                                 & Fasttext-C & 46.4 (8.1)  & 64.6 (4.0)  & 73.7 (2.3)  & 82.1 (1.6) & 43.6 (11.2) & 67.7 (5.1)  & 79.0 (2.4)  & 85.4 (2.8) \\
                         \cmidrule(lr){2-11}
                         & \multirow{2}{*}{1NN}            & BERT-Base  & 91.5 (6.8)  & 96.1 (0.7)  & 96.4 (0.4)  & 96.6 (0.3) & 89.4 (11.0) & 97.0 (0.6)  & 97.3 (0.3)  & 97.5 (0.3) \\
                         &                                 & BERT-Large & 96.1 (1.8)  & 97.0 (0.7)  & 98.0 (0.6)  & 98.2 (0.1) & 96.5 (1.5)  & 97.7 (0.6)  & 98.4 (0.5)  & 98.6 (0.1) \\
                         \cmidrule(lr){2-11}
                         & \multirow{2}{*}{Fine-Tuning}    & BERT-Base  & 79.6 (7.9)  & 95.0 (1.0)  & 94.4 (1.5)  & 96.6 (0.5) & 72.3 (12.5) & 90.0 (1.5)  & 88.7 (3.4)  & 92.3 (0.8) \\
                         &                                 & BERT-Large & 76.2 (12.4) & 94.0 (0.9)  & 94.6 (2.4)  & 97.1 (0.6) & 68.0 (16.4) & 89.0 (3.5)  & 88.3 (4.4)  & 94.0 (1.4) \\
\midrule
\multirow{6}{*}{bow}     & \multirow{2}{*}{Static emb.} & Fasttext-B & 29.8 (16.7) & 41.2 (20.6) & 40.2 (9.1)  & 63.3 (4.3) & 39.1 (7.6)  & 35.2 (2.3)  & 39.5 (7.9)  & 64.8 (1.5) \\
                         &                                 & Fasttext-C & 52.3 (8.1)  & 62.6 (4.6)  & 79.4 (0.8)  & 87.6 (1.2) & 49.3 (7.2)  & 60.5 (6.1)  & 76.4 (1.8)  & 87.1 (1.5) \\
                         \cmidrule(lr){2-11}
                         & \multirow{2}{*}{1NN}            & BERT-Base  & 86.5 (1.3)  & 91.8 (2.7)  & 95.5 (0.2)  & 95.3 (0.4) & 80.7 (4.2)  & 88.6 (2.0)  & 93.7 (1.5)  & 95.0 (0.5) \\
                         &                                 & BERT-Large & 87.4 (1.7)  & 94.9 (2.3)  & 97.4 (0.6)  & 96.4 (1.0) & 84.3 (4.2)  & 93.3 (4.5)  & 97.8 (0.5)  & 96.7 (1.0) \\
                         \cmidrule(lr){2-11}
                         & \multirow{2}{*}{Fine-Tuning}    & BERT-Base  & 83.1 (3.1)  & 89.5 (4.1)  & 94.0 (0.8)  & 96.1 (0.2) & 73.2 (6.4)  & 85.6 (4.9)  & 93.1 (1.5)  & 95.3 (0.4) \\
                         &                                 & BERT-Large & 78.8 (13.6) & 91.0 (3.9)  & 96.7 (0.7)  & 97.5 (0.6) & 73.1 (12.0) & 91.0 (2.2)  & 96.9 (0.9)  & 97.5 (1.1) \\
\midrule
\multirow{6}{*}{club}    & \multirow{2}{*}{Static emb.} & Fasttext-B & 35.0 (11.8) & 26.2 (19.3) & 23.8 (5.5)  & 56.1 (4.1) & 31.5 (1.0)  & 32.6 (1.0)  & 37.1 (1.5)  & 54.4 (1.2) \\
                         &                                 & Fasttext-C & 35.2 (8.1)  & 47.7 (5.4)  & 60.2 (3.8)  & 74.6 (2.8) & 37.4 (1.9)  & 52.3 (3.6)  & 61.7 (2.5)  & 79.0 (2.3) \\
                         \cmidrule(lr){2-11}
                         & \multirow{2}{*}{1NN}            & BERT-Base  & 58.3 (5.5)  & 80.2 (1.9)  & 81.2 (1.1)  & 81.2 (0.8) & 70.5 (2.8)  & 84.9 (2.3)  & 84.9 (1.7)  & 84.7 (1.0) \\
                         &                                 & BERT-Large & 54.1 (10.5) & 82.8 (3.5)  & 83.8 (1.8)  & 84.5 (0.2) & 69.0 (7.7)  & 87.4 (3.2)  & 88.9 (1.7)  & 88.5 (0.3) \\
                         \cmidrule(lr){2-11}
                         & \multirow{2}{*}{Fine-Tuning}    & BERT-Base  & 52.5 (6.1)  & 68.5 (8.9)  & 80.5 (1.5)  & 84.2 (0.8) & 50.0 (6.2)  & 68.3 (10.0) & 79.9 (1.9)  & 83.4 (0.8) \\
                         &                                 & BERT-Large & 51.3 (7.3)  & 71.8 (1.9)  & 81.2 (3.9)  & 83.7 (1.9) & 49.8 (6.5)  & 69.5 (2.7)  & 80.8 (4.1)  & 83.4 (1.5) \\
\midrule
\multirow{6}{*}{trunk}   & \multirow{2}{*}{Static emb.} & Fasttext-B & 32.9 (16.4) & 21.2 (0.6)  & 45.9 (17.7) & 66.7 (3.4) & 34.0 (2.1)  & 35.4 (2.0)  & 43.5 (7.4)  & 67.2 (2.8) \\
                         &                                 & Fasttext-C & 65.4 (6.8)  & 63.2 (7.4)  & 76.6 (2.1)  & 82.7 (3.7) & 65.3 (8.8)  & 67.0 (7.9)  & 78.3 (0.7)  & 87.4 (2.8) \\
                         \cmidrule(lr){2-11}
                         & \multirow{2}{*}{1NN}            & BERT-Base  & 77.9 (18.4) & 84.8 (7.1)  & 94.8 (1.8)  & 95.2 (2.2) & 84.1 (12.1) & 91.2 (4.6)  & 97.2 (1.0)  & 97.4 (1.2) \\
                         &                                 & BERT-Large & 83.1 (20.2) & 96.1 (1.1)  & 96.5 (1.2)  & 98.3 (0.6) & 89.7 (11.5) & 97.9 (0.6)  & 98.1 (0.7)  & 99.1 (0.3) \\
                         \cmidrule(lr){2-11}
                         & \multirow{2}{*}{Fine-Tuning}    & BERT-Base  & 72.7 (16.7) & 89.2 (5.4)  & 93.9 (1.6)  & 97.0 (1.6) & 72.5 (11.2) & 89.1 (4.6)  & 93.5 (1.6)  & 96.7 (1.8) \\
                         &                                 & BERT-Large & 79.2 (14.7) & 91.3 (2.4)  & 95.2 (1.2)  & 96.1 (1.8) & 79.9 (11.7) & 90.9 (2.4)  & 94.9 (1.4)  & 95.7 (2.0) \\
\midrule
\multirow{6}{*}{square}  & \multirow{2}{*}{Static emb.} & Fasttext-B & 20.5 (7.2)  & 8.9 (3.6)   & 38.5 (10.3) & -          & 25.9 (0.8)  & 25.0 (0.0)  & 38.8 (1.9)  & -          \\
                         &                                 & Fasttext-C & 40.1 (19.0) & 60.4 (10.8) & 71.8 (3.2)  & -          & 39.4 (7.1)  & 61.5 (5.5)  & 74.6 (2.9)  & -          \\
                         \cmidrule(lr){2-11}
                         & \multirow{2}{*}{1NN}            & BERT-Base  & 70.2 (7.7)  & 83.3 (7.8)  & 90.7 (3.0)  & -          & 74.1 (10.3) & 83.7 (5.1)  & 88.6 (2.2)  & -          \\
                         &                                 & BERT-Large & 74.9 (13.0) & 84.7 (8.3)  & 93.7 (1.2)  & -          & 79.4 (4.6)  & 84.4 (5.6)  & 89.0 (4.0)  & -          \\
                         \cmidrule(lr){2-11}
                         & \multirow{2}{*}{Fine-Tuning}    & BERT-Base  & 64.9 (11.3) & 75.4 (11.3) & 85.2 (3.6)  & -          & 56.7 (4.3)  & 71.0 (8.1)  & 81.3 (4.0)  & -          \\
                         &                                 & BERT-Large & 63.9 (17.0) & 81.0 (10.2) & 87.3 (2.0)  & -          & 57.7 (7.3)  & 76.9 (8.2)  & 82.9 (1.8)  & -          \\
\midrule
\multirow{6}{*}{arm}     & \multirow{2}{*}{Static emb.} & Fasttext-B & 53.7 (11.1) & 60.0 (3.9)  & 53.3 (8.9)  & 80.3 (2.0) & 58.6 (1.9)  & 61.4 (3.5)  & 62.3 (3.3)  & 79.9 (2.5) \\
                         &                                 & Fasttext-C & 79.1 (7.1)  & 85.4 (7.0)  & 90.9 (4.7)  & 95.7 (0.5) & 85.1 (5.1)  & 86.6 (5.5)  & 91.6 (3.0)  & 95.1 (0.8) \\
                         \cmidrule(lr){2-11}
                         & \multirow{2}{*}{1NN}            & BERT-Base  & 99.4 (0.00) & 99.4 (0.0)  & 99.4 (0.0)  & 99.4 (0.0) & 99.6 (0.0)  & 99.6 (0.0)  & 99.6 (0.0)  & 99.6 (0.0) \\
                         &                                 & BERT-Large & 99.4 (0.00) & 99.4 (0.0)  & 99.4 (0.0)  & 99.4 (0.0) & 99.6 (0.0)  & 99.6 (0.0)  & 99.6 (0.0)  & 99.6 (0.0) \\
                         \cmidrule(lr){2-11}
                         & \multirow{2}{*}{Fine-Tuning}    & BERT-Base  & 96.3 (2.8)  & 99.4 (0.0)  & 99.4 (0.0)  & 99.4 (0.0) & 95.0 (3.9)  & 99.2 (0.0)  & 99.2 (0.0)  & 99.2 (0.0) \\
                         &                                 & BERT-Large & 97.4 (2.1)  & 98.8 (0.9)  & 99.4 (0.0)  & 99.4 (0.0) & 96.4 (2.9)  & 98.4 (1.1)  & 99.2 (0.0)  & 99.2 (0.0) \\
\midrule
\multirow{6}{*}{digit}   & \multirow{2}{*}{Static emb.} & Fasttext-B & 45.2 (5.1)  & 54.0 (16.8) & 43.7 (9.0)  & -          & 49.0 (5.1)  & 69.4 (11.1) & 62.8 (5.0)  & -          \\
                         &                                 & Fasttext-C & 69.8 (9.2)  & 84.9 (8.1)  & 87.3 (5.9)  & -          & 63.3 (3.8)  & 85.0 (5.5)  & 89.2 (2.7)  & -          \\
                         \cmidrule(lr){2-11}
                         & \multirow{2}{*}{1NN}            & BERT-Base  & 96.8 (2.2)  & 99.2 (1.1)  & 100 (0.0) & -          & 92.6 (5.2)  & 98.1 (2.6)  & 100 (0.0) & -          \\
                         &                                 & BERT-Large & 94.4 (6.3)  & 100 (0.0) & 100 (0.0) & -          & 87.0 (14.6) & 100 (0.0) & 100 (0.0) & -          \\
                         \cmidrule(lr){2-11}
                         & \multirow{2}{*}{Fine-Tuning}    & BERT-Base  & 96.0 (2.2)  & 100 (0.0) & 100 (0.0) & -          & 93.6 (4.1)  & 100 (0.0) & 100 (0.0) & -          \\
                         &                                 & BERT-Large & 95.2 (5.1)  & 98.4 (1.1)  & 100 (0.0) & -          & 91.2 (10.0) & 97.5 (1.7)  & 100 (0.0) & -          \\
\midrule
\multirow{6}{*}{bass}    & \multirow{2}{*}{Static emb.} & Fasttext-B & 27.8 (8.9)  & 22.2 (0.5)  & 33.5 (15.3) & 60.7 (3.6) & 39.2 (3.6)  & 36.8 (2.6)  & 39.2 (6.7)  & 71.4 (1.1) \\
                         &                                 & Fasttext-C & 37.1 (8.2)  & 49.8 (5.3)  & 65.1 (6.2)  & 78.9 (3.2) & 49.4 (3.9)  & 57.5 (3.3)  & 67.1 (1.8)  & 78.3 (2.2) \\
                         \cmidrule(lr){2-11}
                         & \multirow{2}{*}{1NN}            & BERT-Base  & 65.6 (17.9) & 75.2 (7.1)  & 70.4 (5.7)  & 77.2 (2.1) & 60.4 (3.2)  & 71.8 (5.8)  & 71.2 (1.6)  & 77.2 (0.5) \\
                         &                                 & BERT-Large & 70.2 (17.8) & 76.9 (6.5)  & 77.1 (4.5)  & 83.4 (4.2) & 70.3 (4.6)  & 78.6 (4.5)  & 80.3 (1.9)  & 83.4 (0.6) \\
                         \cmidrule(lr){2-11}
                         & \multirow{2}{*}{Fine-Tuning}    & BERT-Base  & 63.0 (12.3) & 67.8 (5.5)  & 82.5 (1.8)  & 86.5 (1.2) & 54.2 (2.6)  & 62.2 (3.9)  & 71.1 (1.4)  & 76.2 (0.8) \\
                         &                                 & BERT-Large & 79.4 (15.5) & 70.4 (10.0) & 76.4 (7.5)  & 88.1 (2.8) & 67.6 (7.7)  & 65.0 (5.3)  & 68.7 (5.2)  & 78.5 (2.7) \\
\midrule
\multirow{6}{*}{yard}    & \multirow{2}{*}{Static emb.} & Fasttext-B & 48.6 (10.8) & 35.6 (2.6)  & 40.3 (12.3) & -          & 56.0 (7.0)  & 54.6 (7.5)  & 61.0 (5.6)  & -          \\
                         &                                 & Fasttext-C & 63.4 (26.2) & 74.1 (9.8)  & 83.8 (9.2)  & -          & 62.3 (13.9) & 78.5 (7.2)  & 86.7 (6.2)  & -          \\
                         \cmidrule(lr){2-11}
                         & \multirow{2}{*}{1NN}            & BERT-Base  & 52.8 (15.8) & 70.8 (11.9) & 77.3 (2.9)  & -          & 68.4 (13.0) & 82.8 (7.1)  & 86.6 (1.7)  & -          \\
                         &                                 & BERT-Large & 65.3 (20.5) & 81.0 (14.3) & 85.6 (7.6)  & -          & 79.5 (12.1) & 88.8 (8.4)  & 91.5 (4.5)  & -          \\
                         \cmidrule(lr){2-11}
                         & \multirow{2}{*}{Fine-Tuning}    & BERT-Base  & 56.0 (13.6) & 67.1 (15.9) & 92.1 (4.6)  & -          & 48.3 (10.5) & 62.1 (13.1) & 87.8 (6.2)  & -          \\
                         &                                 & BERT-Large & 48.6 (8.6)  & 82.9 (9.6)  & 90.7 (5.7)  & -          & 46.7 (7.2)  & 77.4 (9.7)  & 85.9 (7.8)  & -          \\
\midrule
\multirow{6}{*}{pound}   & \multirow{2}{*}{Static emb.} & Fasttext-B & 57.7 (20.7) & 39.9 (18.3) & 40.5 (3.4)  & -          & 57.3 (7.0)  & 51.7 (2.6)  & 55.1 (2.0)  & -          \\
                         &                                 & Fasttext-C & 61.2 (20.5) & 58.8 (5.8)  & 68.7 (5.1)  & -          & 57.7 (7.5)  & 62.3 (2.9)  & 73.7 (3.8)  & -          \\
                         \cmidrule(lr){2-11}
                         & \multirow{2}{*}{1NN}            & BERT-Base  & 61.9 (19.0) & 66.3 (14.5) & 77.7 (9.1)  & -          & 55.1 (6.6)  & 81.2 (8.1)  & 86.1 (4.1)  & -          \\
                         &                                 & BERT-Large & 69.4 (26.0) & 69.4 (8.5)  & 82.1 (5.1)  & -          & 59.4 (3.0)  & 81.5 (6.1)  & 90.0 (2.8)  & -          \\
                         \cmidrule(lr){2-11}
                         & \multirow{2}{*}{Fine-Tuning}    & BERT-Base  & 61.9 (10.2) & 63.6 (16.6) & 74.2 (9.7)  & -          & 45.1 (4.7)  & 52.8 (10.7) & 64.5 (8.2)  & -          \\
                         &                                 & BERT-Large & 64.6 (17.2) & 59.1 (2.6)  & 81.1 (9.4)  & -          & 47.9 (5.9)  & 51.4 (1.5)  & 70.0 (8.0)  & -          \\
\midrule
\multirow{6}{*}{deck}    & \multirow{2}{*}{Static emb.} & Fasttext-B & 54.6 (17.2) & 73.1 (10.8) & 71.0 (10.5) & -          & 51.4 (7.9)  & 54.7 (4.3)  & 62.4 (2.6)  & -          \\
                         &                                 & Fasttext-C & 68.4 (32.0) & 66.0 (13.0) & 77.8 (3.3)  & -          & 61.0 (2.1)  & 61.9 (4.1)  & 72.6 (10.0) & -          \\
                         \cmidrule(lr){2-11}
                         & \multirow{2}{*}{1NN}            & BERT-Base  & 81.8 (12.0) & 85.2 (8.1)  & 86.9 (1.4)  & -          & 81.4 (9.0)  & 81.0 (4.6)  & 90.7 (2.8)  & -          \\
                         &                                 & BERT-Large & 76.4 (15.9) & 87.5 (2.5)  & 90.9 (0.8)  & -          & 78.5 (2.9)  & 84.5 (5.3)  & 95.1 (0.4)  & -          \\
                         \cmidrule(lr){2-11}
                         & \multirow{2}{*}{Fine-Tuning}    & BERT-Base  & 86.9 (7.3)  & 87.9 (1.4)  & 86.9 (2.2)  & -          & 70.8 (11.9) & 69.9 (1.3)  & 70.3 (3.6)  & -          \\
                         &                                 & BERT-Large & 77.4 (17.8) & 88.6 (2.4)  & 90.6 (2.7)  & -          & 55.2 (10.4) & 63.8 (12.2) & 77.7 (4.0)  & -          \\
\midrule
\multirow{6}{*}{bank}    & \multirow{2}{*}{Static emb.} & Fasttext-B & 46.5 (12.1) & 46.9 (4.6)  & 51.0 (8.8)  & 77.8 (5.5) & 56.8 (2.0)  & 64.9 (1.2)  & 62.7 (2.7)  & 72.5 (3.7) \\
                         &                                 & Fasttext-C & 38.4 (6.8)  & 70.1 (11.9) & 80.7 (2.8)  & 88.2 (4.6) & 61.9 (0.3)  & 72.8 (4.6)  & 79.1 (3.4)  & 85.9 (2.4) \\
                         \cmidrule(lr){2-11}
                         & \multirow{2}{*}{1NN}            & BERT-Base  & 98.8 (0.7)  & 99.4 (0.5)  & 99.7 (0.1)  & 99.8 (0.0) & 94.3 (3.3)  & 95.4 (4.9)  & 97.7 (0.1)  & 97.7 (0.0) \\
                         &                                 & BERT-Large & 99.0 (0.7)  & 99.5 (0.1)  & 99.9 (0.1)  & 99.9 (0.1) & 95.9 (3.5)  & 97.6 (1.8)  & 99.2 (1.1)  & 99.2 (1.1) \\
                         \cmidrule(lr){2-11}
                         & \multirow{2}{*}{Fine-Tuning}    & BERT-Base  & 91.9 (5.7)  & 97.8 (1.4)  & 97.9 (0.8)  & 99.0 (0.1) & 76.1 (9.4)  & 89.4 (5.2)  & 90.5 (3.3)  & 95.3 (0.5) \\
                         &                                 & BERT-Large & 58.0 (28.9) & 98.6 (0.7)  & 99.5 (0.1)  & 98.6 (0.6) & 52.3 (28.3) & 92.3 (4.0)  & 97.3 (0.5)  & 93.5 (2.4) \\
\midrule
\multirow{6}{*}{pitcher} & \multirow{2}{*}{Static emb.} & Fasttext-B & 92.1 (2.2)  & 82.7 (9.3)  & 82.8 (1.3)  & -          & 69.2 (8.0)  & 73.4 (10.1) & 82.4 (4.3)  & -          \\
                         &                                 & Fasttext-C & 96.5 (4.2)  & 95.9 (1.8)  & 91.2 (3.0)  & -          & 84.2 (13.5) & 94.1 (0.9)  & 94.3 (3.0)  & -          \\
                         \cmidrule(lr){2-11}
                         & \multirow{2}{*}{1NN}            & BERT-Base  & 100 (0.0)   & 99.9 (0.1)  & 99.8 (0.0)  & -          & 100 (0.0) & 99.9 (0.0)  & 99.9 (0.0)  & -          \\
                         &                                 & BERT-Large & 100 (0.0)   & 100 (0.0) & 99.9 (0.0)  & -          & 100 (0.0) & 100 (0.0) & 100 (0.0) & -          \\
                         \cmidrule(lr){2-11}
                         & \multirow{2}{*}{Fine-Tuning}    & BERT-Base  & 98.6 (0.5)  & 98.9 (0.6)  & 97.2 (1.1)  & -          & 70.6 (5.3)  & 76.2 (10.2) & 62.8 (4.0)  & -          \\
                         &                                 & BERT-Large & 97.1 (2.3)  & 98.7 (1.2)  & 98.5 (1.1)  & -          & 70.5 (17.3) & 77.3 (13.8) & 74.8 (13.6) & -         \\
\bottomrule
\caption{Micro- and macro-F1 results in the $n$-shot setting for all the two BERT-based WSD strategies (as well as for the static embedding baseline) in our experiments and for all the words in the dataset. Results are the average of three runs (standard deviation is shown in parentheses).\label{tab:appendixfewshot}}
\end{longtabu}

\small
\begin{longtabu} to 0.9\textwidth{ll p{0.15\textwidth} p{0.65\textwidth} p{0.5\textwidth}}
\toprule
\bf Word &
\bf Sense \# &
\bf Sense ID &
{\bf Definition} (1st sentence from Wikipedia) &
{\bf Example usage} (tokenized) \\
    \midrule
    \endhead
    \multirow{2}{*}{\begin{sideways}\textbf{Crane}\end{sideways}}
      &   crane$_1$  &   crane (machine)  &   A crane is a type of machine, generally equipped with a hoist rope, wire ropes or chains, and sheaves, that can be used both to lift and lower materials and to move them horizontally.  &   launching and recovery is accomplished with the assistance of a shipboard crane . \\
      \cmidrule(lr){2-5}
  &   crane$_2$  &   crane (bird)  &   Cranes are a family, the Gruidae, of large, long-legged, and long-necked birds in the group Gruiformes.  &   tibet hosts species of wolf , wild donkey , crane , vulture , hawk , geese , snake , and buffalo . \\
  \midrule
  \multirow{2}{*}{\begin{sideways}\textbf{Java}\end{sideways}}
  &   java$_1$  &   java  &   Java is an island of Indonesia, bordered by the Indian Ocean on the south and the Java Sea on the north.  &   in indonesia , only sumatra , borneo , and papua are larger in territory , and only java and sumatra have larger populations . \\
      \cmidrule(lr){2-5}
  &   java$_2$  &   java (programming language)  &   Java is a general-purpose programming language that is class-based, object-oriented, and designed to have as few implementation dependencies as possible.  &   examples include the programming languages perl , java and lua . \\
 
  \midrule
  \multirow{2}{*}{\begin{sideways}\textbf{Apple}\end{sideways}}
  &   apple$_1$  &   apple inc.  &   Apple Inc. is an American multinational technology company headquartered in Cupertino, California, that designs, develops, and sells consumer electronics, computer software, and online services.  &   shopify released a free mobile app on the apple app store on may 13 , 2010 . \\
  \cmidrule(lr){2-5}
  &   apple$_2$  &   apple  &   An apple is an edible fruit produced by an apple tree.  &   cherry , apple , pear , peach and apricot trees are available . \\
  
  \midrule
  \multirow{5}{*}{\begin{sideways}\textbf{Mole}\end{sideways}}
  &   mole$_1$  &   mole (animal)  &   Moles are small mammals adapted to a subterranean lifestyle (i.e., fossorial).  &   its primary prey consists of mice , rat , squirrel , chipmunk , shrew , mole and rabbits . \\
  \cmidrule(lr){2-5}
  &   mole$_2$  &   mole (espionage)  &   In espionage jargon, a mole is a long-term spy who is recruited before having access to secret intelligence, subsequently managing to get into the target organization.  &   philip meets claudia where she tells him that there is a mole working for the fbi . \\
  \cmidrule(lr){2-5}
  &   mole$_3$  &   mole (unit)  &   The mole (symbol: mol) is the unit of measurement for amount of substance in the International System of Units (SI).  &   so the specific heat of a classical solid is always 3k per atom , or in chemistry units , 3r per mole of atoms . \\
  \cmidrule(lr){2-5}
  &   mole$_4$  &   mole sauce  &   Mole is a traditional marinade and sauce originally used in Mexican cuisine.  &   food such as cake , chicken with mole , hot chocolate , coffee , and atole are served . \\
  \cmidrule(lr){2-5}
  &   mole$_5$  &   mole (architecture)  &   A mole is a massive structure, usually of stone, used as a pier, breakwater, or a causeway between places separated by water.  &   the islands of pomègues and ratonneau are connected by a mole built in 1822 . \\
 
  \midrule
  \multirow{3}{*}{\begin{sideways}\textbf{Spring}\end{sideways}}
  &   spring$_1$  &   spring (hydrology)  &   A spring  is a point at which water flows from an aquifer to the Earth's surface. It is a component of the hydrosphere.  &   the village was famous for its mineral water spring used for healing in sanatorium , including the hawthorne and lithia springs . \\
  \cmidrule(lr){2-5}
  &   spring$_2$  &   spring (season)  &   Spring, also known as springtime, is one of the four temperate seasons, succeeding winter and preceding summer.  &   the species is most active during the spring and early summer although it may be seen into late june . \\
  \cmidrule(lr){2-5}
  &   spring$_3$  &   spring (device)  &   A spring is an elastic object that stores mechanical energy.  &   often spring are used to reduce backlash of the mechanism . \\

\midrule
  \multirow{2}{*}{\begin{sideways}\textbf{Chair}\end{sideways}}
  &   chair$_1$  &   chairman  &   The chairperson (also chair, chairman, or chairwoman) is the presiding officer of an organized group such as a board, committee, or deliberative assembly.  &   gan is current chair of the department of environmental sciences at university of california , riverside . \\
  \cmidrule(lr){2-5}
  &   chair$_2$  &   chair  &   One of the basic pieces of furniture, a chair is a type of seat.  &   a typical western living room may contain furnishings such as a sofa , chair , occasional table , and bookshelves , electric lamp , rugs , or other furniture . \\

\midrule
\multirow{3}{*}{\begin{sideways}\textbf{Hood}\end{sideways}}
  &   hood$_1$  &   hood (comics)  &   Hood (real name Parker Robbins) is a fictional character, a supervillain, and a crime boss appearing in American comic books published by Marvel Comics.  &   the hood has hired him as part of his criminal organization to take advantage of the split in the superhero community caused by the super-human registration act . \\
  \cmidrule(lr){2-5}
  &   hood$_2$  &   hood (vehicle)  &   The hood (North American English) or bonnet (Commonwealth English excluding Canada) is the hinged cover over the engine of motor vehicles that allows access to the engine compartment, or trunk (boot in Commonwealth English) on rear-engine and some mid-engine vehicles) for
  maintenance and repair.  &   european versions of the car also had an air intake on the hood . \\
  \cmidrule(lr){2-5}
  &   hood$_3$  &   hood (headgear)  &   A hood is a kind of headgear that covers most of the head and neck, and sometimes the face.  &   in some sauna suits , the jacket also includes a hood to provide additional retention of body heat . \\
  
  \midrule
  \multirow{4}{*}{\begin{sideways}\textbf{Seal}\end{sideways}}
  &   seal$_1$  &   pinniped  &   Pinnipeds, commonly known as seals, are a widely distributed and diverse clade of carnivorous, fin-footed, semiaquatic marine mammals.  &   animals such as shark , stingray , weever fish , seal and jellyfish can sometimes present a danger . \\
  \cmidrule(lr){2-5}
  &   seal$_2$  &   seal (musician)  &   Henry Olusegun Adeola Samue (born 19 February 1963), known professionally as Seal, is a British singer-songwriter.  &   she was married to english singer seal from 2005 until 2012 . \\
  \cmidrule(lr){2-5}
  &   seal$_3$  &   seal (emblem)  &   A seal is a device for making an impression in wax, clay, paper, or some other medium, including an embossment on paper, and is also the impression thus made.  &   each level must review , add information as necessary , and stamp or seal that the submittal was examined and approved by that party . \\
  \cmidrule(lr){2-5}
  &   seal$_4$  &   seal (mechanical)  &   A mechanical seal is a device that helps join systems or mechanisms together by preventing leakage (e.g. in a pumping system), containing pressure, or excluding contamination.  &   generally speaking , standard ball joints will outlive sealed ones because eventually the seal will break , causing the joint to dry out and rust . \\

\midrule
\multirow{3}{*}{\begin{sideways}\textbf{Bow}\end{sideways}}
  &   bow$_1$  &   bow (ship)  &   The bow is the forward part of the hull of a ship or boat,  &   the stem is the most forward part of a boat or ship 's bow and is an extension of the keel itself . \\
  \cmidrule(lr){2-5}
  &   bow$_2$  &   bow and arrow  &   The bow and arrow is a ranged weapon system consisting of an elastic launching device (bow) and long-shafted projectiles (arrows).  &   bow and arrow used in warfare . \\
  \cmidrule(lr){2-5}
  &   bow$_3$  &   bow (music)  &   In music, a bow is a tensioned stick which has hair (usually horse-tail hair) coated in rosin (to facilitate friction) affixed to it.  &   horsehair is used for brush , the bow of musical instruments and many other things . \\

\midrule
\multirow{3}{*}{\begin{sideways}\textbf{Club}\end{sideways}}
  &   club$_1$  &   club  &   A club is an association of people united by a common interest or goal.  &   this is a partial list of women 's association football club teams from all over the world sorted by confederation . \\
  \cmidrule(lr){2-5}
  &   club$_2$  &   nightclub  &   A nightclub, music club, or club, is an entertainment venue and bar that usually operates late into the night.  &   although several of his tracks were club hits , he had limited chart success . \\
  \cmidrule(lr){2-5}
  &   club$_3$  &   club (weapon)  &   A club (also known as a cudgel, baton, bludgeon, truncheon, cosh, nightstick or impact weapon) is among the simplest of all weapons: a short staff or stick, usually made of wood, wielded as a weapon since prehistoric times.  &   before their adoption of guns , the plains indians hunted with spear , bows and arrows , and various forms of club . \\

\midrule
\multirow{3}{*}{\begin{sideways}\textbf{Trunk}\end{sideways}}
  &   trunk$_1$  &   trunk (botany)  &   In botany, the trunk (or bole) is the stem and main wooden axis of a tree.  &   its leaves are different from the leaves of true palms , and unlike true palms it does not develop a woody trunk . \\
  \cmidrule(lr){2-5}
  &   trunk$_2$  &   trunk (automobile)  &   The trunk (North American English), boot (British English), dickey (Indian English) (also spelled dicky or diggy) or compartment (South-East Asia) of a car is the vehicle's main storage or cargo compartment.  &   unlike the bmw x5 , the x-coupe had an aluminium body , a trunk opening downwards and two doors that swing outward . \\
  \cmidrule(lr){2-5}
  &   trunk$_3$  &   trnuk (anatomy)  &   The torso or trunk is an anatomical term for the central part or core of many animal bodies (including humans) from which extend the neck and limbs.  &   surface projections of the major organs of the trunk , using the vertebral column and rib cage as main reference points of superficial anatomy . \\
  
  \midrule
  \multirow{4}{*}{\begin{sideways}\textbf{Square}\end{sideways}}
  &   square$_1$  &   square  &   In geometry, a square is a regular quadrilateral, which means that it has four equal sides and four equal angles (90-degree angles, or 100-gradian angles or right angles).  &   similarly , a square with all sides of length has the perimeter and the same area as the rectangle . \\
  &   square$_2$  &   square (company)  &   Square Co., Ltd. was a Japanese video game company founded in September 1986 by Masafumi Miyamoto. It merged with Enix in 2003 to form Square Enix.  &   video game by square , features the orbital elevator '' a.t.l.a.s. '' . \\
  \cmidrule(lr){2-5}
  &   square$_3$  &   town square  &   A town square is an open public space commonly found in the heart of a traditional town used for community gatherings.  &   here is a partial list of notable expressways , tunnel , bridge , road , avenues , street , crescent , square and bazaar in hong kong . \\
  \cmidrule(lr){2-5}
  &   square$_4$  &   square number  &   In mathematics, a square number or perfect square is an integer that is the square of an integer.  &   in mathematics eighty-one is the square of 9 and the fourth power of 3 . \\
  
  \midrule
  \multirow{2}{*}{\begin{sideways}\textbf{Arm}\end{sideways}}
  &   arm$_1$  &   arm architecture  &   Arm (previously officially written all caps as ARM and usually written as such today), previously Advanced RISC Machine, originally Acorn RISC Machine, is a family of reduced instruction set computing (RISC) architectures for computer processors, configured for various environments.  &   windows embedded compact is available for arm , mips , superh and x86 processor architectures . \\
  \cmidrule(lr){2-5}
  &   arm$_2$  &   arm  &   In human anatomy, the arm is the part of the upper limb between the glenohumeral joint (shoulder joint) and the elbow joint.  &   on the human body , the limb can be divided into segments , such as the arm and the forearm of the upper limb , and the thigh and the leg of the lower limb . \\
  
  \midrule
  \multirow{2}{*}{\begin{sideways}\textbf{Digit}\end{sideways}}
  &   digit$_1$  &   numerical digit  &   A numerical digit is a single symbol (such as "2" or "5") used alone, or in combinations (such as "25"), to represent numbers (such as the number 25) according to some positional numeral systems.  &   it uses the digit 0 , 1 , 2 and 3 to represent any real number . \\
  \cmidrule(lr){2-5}
  &   digit$_2$  &   digit (anatomy)  &   A digit is one of several most distal parts of a limb, such as fingers or toes, present in many vertebrates.  &   a finger is a limb of the human body and a type of digit , an organ of and found in the hand of human and other primate . \\

\midrule
\multirow{3}{*}{\begin{sideways}\textbf{Bass}\end{sideways}}
  &   bass$_1$  &   bass (guitar)  &   The bass guitar, electric bass, or simply bass, is the lowest-pitched member of the guitar family.  &   the band decided to continue making music after thirsk 's death , and brought in bass guitarist randy bradbury from one hit wonder . \\
\cmidrule(lr){2-5}
  &   bass$_2$  &   bass (voice type)  &   A bass is a type of classical male singing voice and has the lowest vocal range of all voice types.  &   he is known for his distinctive and untrained bass voice . \\
  \cmidrule(lr){2-5}
  &   bass$_3$  &   double bass  &   The double bass, also known simply as the bass (or by other names), is the largest and lowest-pitched bowed (or plucked) string instrument in the modern symphony orchestra.  &   his instruments were the bass and the tuba . \\
  
  \midrule
  \multirow{2}{*}{\begin{sideways}\textbf{Yard}\end{sideways}}
  &   yard$_1$  &   yard  &   The yard (abbreviation: yd) is an English unit of length, in both the British imperial and US customary systems of measurement, that comprises 3 feet or 36 inches.  &   accuracy is sufficient for hunting small game at ranges to 50 yard . \\
  \cmidrule(lr){2-5}
  &   yard$_2$  &   yard (sailing)  &   A yard is a spar on a mast from which sails are set.  &   aubrey improves sophie s sailing qualities by adding a longer yard which allows him to spread a larger mainsail . \\
  
  \midrule
  \multirow{2}{*}{\begin{sideways}\textbf{Pound}\end{sideways}}
  &   pound$_1$  &   pound (mass)  &   The pound or pound-mass is a unit of mass used in the imperial, United States customary and other systems of measurement.  &   it is approximately 16.38 kilogram ( 36.11 pound ) . \\
  \cmidrule(lr){2-5}
  &   pound$_2$  &   pound (currency)  &   A pound is any of various units of currency in some nations.  &   in english , the maltese currency was referred to as the pound originally and for many locals this usage continued . \\
  
  \midrule
  \multirow{2}{*}{\begin{sideways}\textbf{Deck}\end{sideways}}
  &   deck$_1$  &   deck (ship)  &   A deck is a permanent covering over a compartment or a hull of a ship.  &   the protective deck was thick and ran the full length of the ship . \\
  \cmidrule(lr){2-5}
  &   deck$_2$  &   deck (building)  &   In architecture, a deck is a flat surface capable of supporting weight, similar to a floor, but typically constructed outdoors, often elevated from the ground, and usually connected to a building.  &   typically , it is a wooden deck near a hiking trail that provides the hikers a clean and even place to sleep . \\
  
  \midrule
    \multirow{2}{*}{\begin{sideways}\textbf{Bank}\end{sideways}}
  &   bank$_1$  &   bank  &   A bank is a financial institution that accepts deposits from the public and creates a demand deposit, while simultaneously making loans.  &   the bank , which loans money to the player after they have a house for collateral . \\
  \cmidrule(lr){2-5}
  &   bank$_2$  &   bank (geography)  &   In geography, a bank is the land alongside a body of water.  &   singapore 's first market was located at the south bank of the singapore river . \\
  
  \midrule
  \multirow{2}{*}{\begin{sideways}\textbf{Pitcher}\end{sideways}}
  &   pitcher$_1$  &   pitcher  &   In baseball, the pitcher is the player who throws the baseball from the pitcher's mound toward the catcher to begin each play, with the goal of retiring a batter, who attempts to either make contact with the pitched ball or draw a walk.  &   kasey garret olemberger ( born march 18 , 1978 ) is an italian american professional baseball pitcher . \\
  \cmidrule(lr){2-5}
  &   pitcher$_2$  &   pitcher (container)  &   In American English, a pitcher is a container with a spout used for storing and pouring liquids.  &   pottery was found as grave goods , including combinations of pitcher and cup . \\
      \bottomrule    
  \caption{Sense definitions in the CoarseWSD-20 dataset. Each sense is accompanied with an example usage from the dataset. Sense IDs correspond to the current Wikipedia page of each sense by the date of the submission.}
  \label{tab:sense-definitions}
\end{longtabu} 

\end{landscape}

\end{document}